\documentclass[iicol]{sn-jnl} 

\usepackage{graphicx}%
\usepackage{multirow}%
\usepackage{amsmath,amssymb,amsfonts}%
\usepackage{amsthm}%
\usepackage{mathrsfs}%
\usepackage[title]{appendix}%
\usepackage{xcolor}%
\usepackage{textcomp}%
\usepackage{manyfoot}%
\usepackage{booktabs}%
\usepackage{algorithm}%
\usepackage{algorithmicx}%
\usepackage{algpseudocode}%
\usepackage{listings}%
\usepackage{tabu}
\usepackage{natbib}

\usepackage{array}
\usepackage{multirow}
\usepackage{textcomp}
\usepackage{stfloats}
\usepackage{url}
\usepackage{verbatim}
\usepackage{graphicx}
\usepackage{times}
\usepackage{epsfig}
\usepackage{amssymb}

\usepackage[capitalize]{cleveref}
\crefname{section}{Sec.}{Secs.}
\Crefname{section}{Section}{Sections}
\Crefname{table}{Table}{Tables}
\crefname{table}{Tab.}{Tabs.}

\usepackage{graphicx}
\usepackage{subcaption}
\usepackage{float}
\usepackage[justification=raggedright]{caption}	
\usepackage{lscape}                                         
\usepackage{wrapfig}
\usepackage{pgfplots}

\usepackage{animate} 

\usepackage{booktabs}                   
\usepackage{multirow}

\usepackage{makecell}

\usepackage{paralist}
\usepackage{enumitem}
\setlist[enumerate]{itemsep=0mm}

\usepackage{bm}                          
\usepackage{epsfig}                      
\usepackage{graphicx}                  
\usepackage{mathtools}
\usepackage{amssymb,amsmath}   

\usepackage{units}
\usepackage{color}

\usepackage{url}
\usepackage{xspace}
\usepackage{grfext}
\PrependGraphicsExtensions*{.jpg,.png,.PNG}
\definecolor{firebrick}{HTML}{d48265}
\definecolor{lightfirebrick}{HTML}{c23531}
\definecolor{grayblue}{HTML}{61a0a8}
\definecolor{lightgrayblue}{HTML}{2f4554}
\definecolor{turquoise}{rgb}{0., 0.4, 0.7}
\definecolor{lightturquoise}{rgb}{0.4, 0.8, 0.96}
\definecolor{roseblue}{rgb}{0.27, 0.458, 0.988}
\definecolor{lightorange}{HTML}{C27000}
\definecolor{darkorange}{HTML}{D3A171}
\definecolor{lightgreen}{HTML}{61a0a8}
\definecolor{darkgreen}{HTML}{749f83}
\definecolor{goldyellow}{rgb}{0.964, 0.886, 0.4}

\usepackage{textcomp,gensymb}
\usepackage{tikz}
\usepackage{cuted}

\pgfplotsset{compat=1.18} 

\usepackage{soul}
\usepackage{svg}

\DeclareMathAlphabet{\altmathcal}{OMS}{cmsy}{m}{n}
\DeclareMathAlphabet{\mathbfit}{OT1}{ptm}{bx}{it}

\newlength\paramargin
\newlength\figmargin

\newlength\secmargin
\newlength\figcapmargin
\newlength\tabcapmargin

\newcommand{\topic}[1]
{
\vspace{2mm}\noindent\textbf{#1}
}

\long\def\ignorethis#1{}
\newcommand {\junyan}[1]{}
\newcommand {\new}[1]{{#1}\normalfont}

\newbox\jsavebox%

\makeatletter
\newcommand{\providelength}[1]{%
  \@ifundefined{\expandafter\@gobble\string#1}
   {
    \typeout{\string\providelength: making new length \string#1}%
    \newlength{#1}%
   }
   {
    \sdaau@checkforlength{#1}%
   }%
}

\newcommand{\sdaau@checkforlength}[1]{%
  \edef\sdaau@temp{\expandafter\sdaau@getfive\meaning#1TTTTT$}%
  \ifx\sdaau@temp\sdaau@skipstring
    \typeout{\string\providelength: \string#1 already a length}%
  \else
    \@latex@error
      {\string#1 illegal in \string\providelength}
      {\string#1 is defined, but not with \string\newlength}%
  \fi
}
\def\sdaau@getfive#1#2#3#4#5#6${#1#2#3#4#5}
\edef\sdaau@skipstring{\string\skip}
\makeatother

\def\xi{\mathbf{x}_i}

\graphicspath{{figures}, {examples}}

\begin{document}

\title[Rich Text to Image]{Expressive Image Generation and Editing with Rich Text}


\author{\fnm{Songwei} \sur{Ge}}

\author{\fnm{Taesung} \sur{Park}}

\author{\fnm{Jun-Yan} \sur{Zhu}}

\author{\fnm{Jia-Bin} \sur{Huang}}











\abstract{\new{Plain text has become a prevalent interface for text-based image synthesis and editing}. 
Its limited customization options, however, hinder users from accurately describing desired outputs. 
For example, plain text makes it hard to specify continuous quantities, such as the precise RGB color value or importance of each word. 
Creating detailed text prompts for complex scenes is tedious for humans to write and challenging for text encoders to interpret.
Furthermore, describing a reference concept or texture in plain text is non-trivial.
To address these challenges, we propose using a rich-text editor supporting formats such as font style, size, color, texture fill, footnote, and embedded image.
We extract each word's attributes from rich text to enable local style control, explicit token reweighting, precise color rendering, and detailed region synthesis with reference concepts or texture.
We achieve these capabilities through a region-based diffusion process.
We first obtain each word's mask that characterizes the region guided by the word. 
\new{For each region, we enforce its text attributes by creating customized prompts, applying guidance within the region, and maintaining its fidelity against plain-text generations or input images through region-based injections.
We present various examples of image generation and editing from rich text and demonstrate that our method outperforms strong baselines with quantitative evaluations.}}




\maketitle
\begin{figure*}
\centering
\includegraphics[width=\linewidth,trim=0 0 0 0, clip]{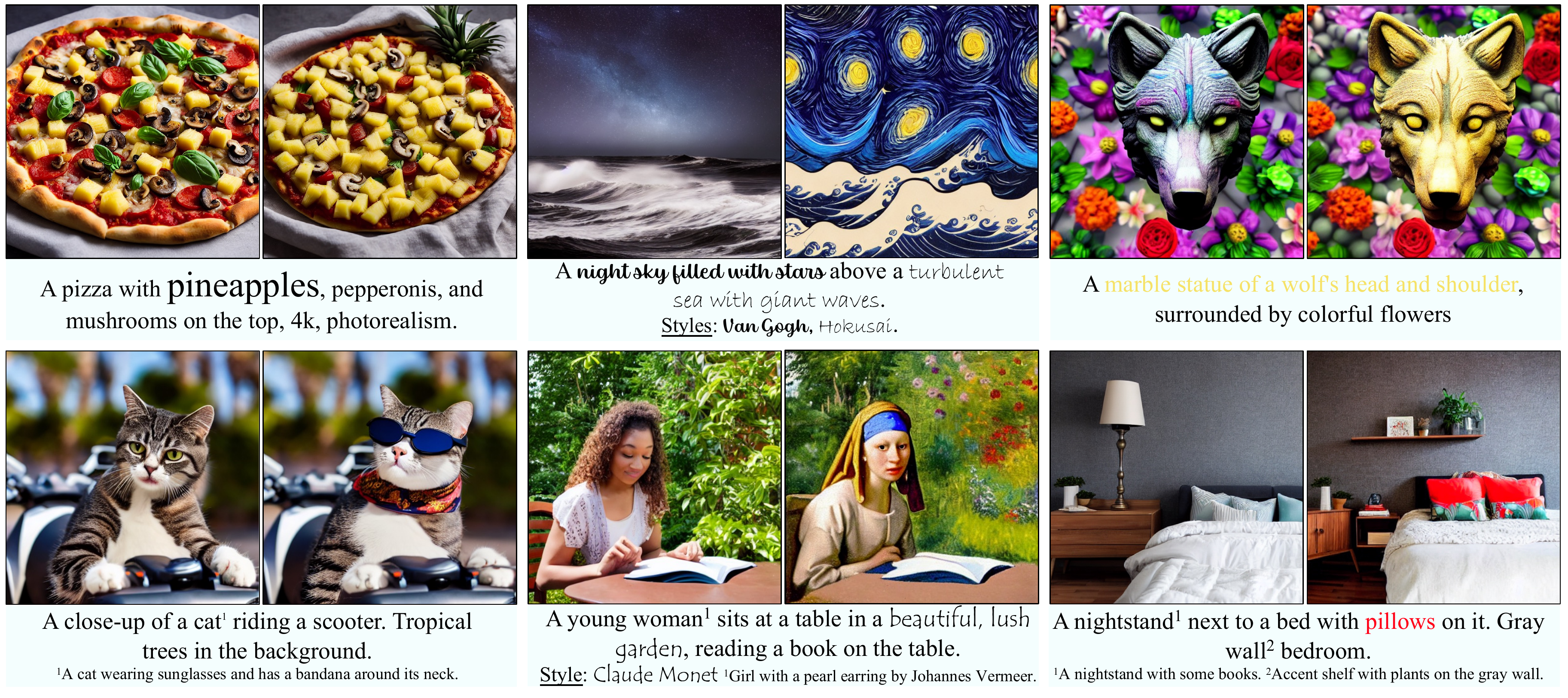}
\captionof{figure}{{\bf Plain text (left image) vs. Rich text (right image)} Our method allows a user to describe an image using a rich text editor that supports various text attributes like font family, size, color, and footnote. Given these text attributes extracted from rich-text prompts, our method enables precise control of text-to-image synthesis regarding colors, styles, and object details compared to plain text. }
\label{fig:teaser}
\end{figure*}

\section{Introduction}
\label{sec:intro}
The development of large-scale text-to-image generative models~\citep{DALLE, Imagen, rombach2022high,kang2023gigagan} has propelled image generation to an unprecedented era.
The great flexibility of these large-scale models further offers users powerful control of the generation through visual cues~\citep{balaji2022ediffi,Make-A-Scene,zhang2023adding} and textual inputs~\citep{brooks2022instructpix2pix,hertz2022prompt}.
Without exception, existing studies use \emph{plain text} encoded by a pretrained language model to guide the generation. 
However, in our daily lives, it is rare to use only plain text when working on text-based tasks such as writing blogs or editing essays.
Instead, a \emph{rich text} editor~\citep{rtf,witten2009build} is the more popular choice, providing versatile formatting options for writing and editing text. 
In this paper, we seek to introduce accessible and precise textual control from rich text editors to text-to-image synthesis.

Rich text editors offer unique solutions for incorporating conditional information separate from the text. 
For example, using the font color, one can indicate an \textit{arbitrary} color. 
In contrast, describing the precise color with plain text proves more challenging as general text encoders do not understand RGB or Hex triplets, and many color names, such as `olive' and `orange', have ambiguous meanings. 
This font color information can be used to define the color of generated objects. 
For example, in Figure~\ref{fig:teaser}, a specific \textcolor{goldyellow}{yellow} can be selected to instruct the generation of a marble statue with that exact color.

Beyond providing precise color information, various font formats make it simple to augment the word-level information. 
For example, reweighting token influence~\citep{hertz2022prompt} can be implemented using the font size, which is challenging to achieve with existing visual or textual interfaces.  
Nevertheless, rich text editors offer more options than font size -- similar to how font style distinguishes the styles of individual text elements, we propose using it to capture the artistic style of specific regions. 
\new{Furthermore, embedded images provide extra information in \emph{visual} format. 
We utilize such images as reference concepts to guide the generated objects. 
This intuitive design enables personalized/customized generation.
}

But how can we use rich text? A straightforward implementation is to convert a rich-text prompt with detailed attributes into lengthy plain text and feed it directly into existing methods~\citep{rombach2022high,hertz2022prompt,brooks2022instructpix2pix}. Unfortunately, these methods struggle to synthesize images corresponding to lengthy text prompts involving multiple objects with distinct visual attributes, as noted in a recent study~\citep{chefer2023attend}. They often mix styles and colors, applying a uniform style to the entire image. Furthermore, the lengthy prompt introduces extra difficulty for text encoders to interpret accurate information, making generating intricate details more demanding. 

\begin{figure*}[t]
    \centering
    \includegraphics[width=\linewidth, trim=0 0 0 0, clip]{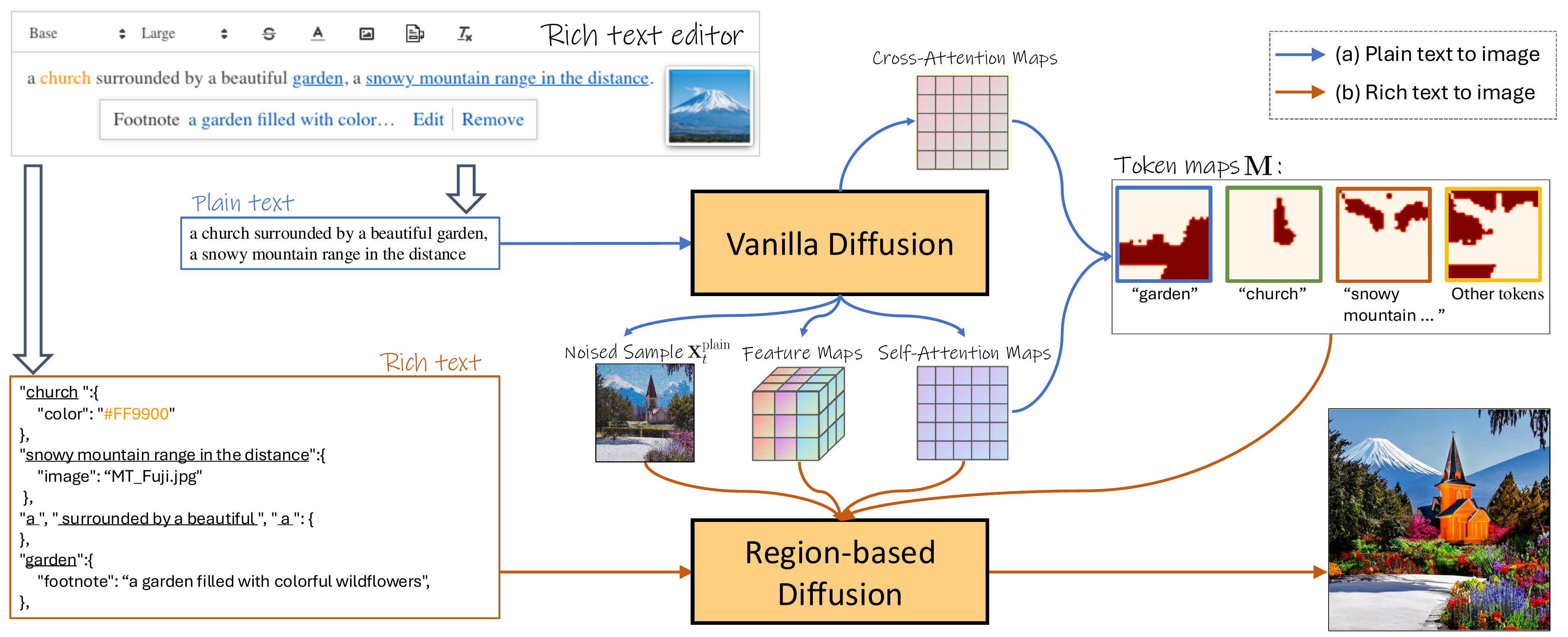}
    \caption{\textbf{Rich-text-to-image framework.} First, the plain-text prompt is processed by a diffusion model to collect self- and cross-attention maps, noised generation, and residual feature maps at certain steps. The token maps of the input prompt are constructed by first creating a segmentation using the self-attention maps and then labeling each segment using the cross-attention maps.
    Then the rich texts are processed as JSON to provide attributes for each token span. The resulting token maps and attributes are used to guide our region-based control. 
    We inject the self-attention maps, noised generation, and feature maps to improve fidelity to the plain-text generation.
    }
    \label{fig:framework}
\end{figure*}

To address these challenges, our insight is to decompose a rich-text prompt into two components (1) a short plain-text prompt (without formatting) and (2)  multiple region-specific prompts that include text attributes, as shown in Figure~\ref{fig:framework}.  
First, we obtain the self- and cross-attention maps using a vanilla denoising process with the short \emph{plain-text} prompt to associate each word with a specific region.
Second, we create a prompt for each region using the attributes derived from \emph{rich-text} prompt. For example, we use ``mountain in the style of Ukiyo-e'' as the prompt for the region corresponding to the word ``mountain'' with the attribute ``font style: Ukiyo-e''. 
For RGB font colors that cannot be converted to the prompts, we iteratively update the region with region-based guidance to match the target color. We apply a separate denoising process for each region and fuse the predicted noises to get the final update.  
During this process, regions associated with the tokens that do not have any formats are supposed to look the same as the plain-text results. Also, the overall shape of the objects should stay unchanged in cases such as only the color is changed. To this end, we propose to use region-based injection approaches.

We demonstrate qualitatively and quantitatively that our method generates more precise color, distinct styles, and accurate details compared to plain text-based methods. 
\new{
We conduct a thorough quantitative evaluation by building a rich-text benchmark by collecting a diverse set of rich-text prompts with font color, style, and footnotes.
}

\new{
A preliminary version of this work was published earlier in \citep{ge2023expressive}.
In this paper, we extend our work and summarize the core differences below. 
\begin{itemize}
\item We build a benchmark for evaluating the task of rich text-to-image generation. 
This includes a quantitative evaluation of image generation with complex prompts. 
The benchmark presents new challenges for future research.
\item We develop two novel applications, leveraging additional rich-text font attributes: 1) using embedded images to guide the concept being generated in the image and 2) enabling texture fill to guide the texture being rendered in the object.
\item We apply rich texts for editing \emph{real} images. 
By leveraging diffusion inversion techniques and segmentation methods, we show rich text-based editing allows precise control of the editing results.
\end{itemize}
}

\section{Related Work}
\label{sec:related}

\topic{Text-to-image models.} 
Text-to-image systems aim to synthesize realistic images according to descriptions~\citep{zhu2007text,mansimov2015generating}.
Fueled by the large-scale text-image datasets~\citep{schuhmann2022laion,kakaobrain2022coyo-700m}, various training and inference techniques~\citep{ho2020denoising, song2021denoising,ho2022cascaded,Classifier}, and scalibility~\citep{DALLE2}, significant progress has been made in text-to-image generation using diffusion models~\citep{balaji2022ediffi,DALLE2,GLIDE,Imagen,Make-A-Scene}, autoregressive models~\citep{DALLE,Parti,chang2023muse,ding2022cogview2}, GANs~\citep{sauer2023stylegan,kang2023gigagan}, and their hybrids~\citep{rombach2022high}. 
Our work focuses on making these models more accessible and providing precise controls. 
In contrast to existing work that uses \emph{plain text}, we use a \emph{rich text} editor with various formatting options.

\topic{Controllable image synthesis with diffusion models.} 
A wide range of image generation and editing applications are achieved through either fine-tuning pre-trained diffusion models~\citep{ruiz2022dreambooth,kumari2022multi,zhang2023adding,avrahami2022spatext,wu2022tune,kawar2023imagic,ma2023follow,li2023gligen} or modifying the denoising process~\citep{meng2021sdedit,choi2021ilvr,hertz2022prompt,parmar2023zero,bansal2023universal,chefer2023attend,avrahami2022blended,balaji2022ediffi,jimenez2023mixture,bar2023multidiffusion,Sarukkai2023CollageD,zhang2023diffcollage,cao2023masactrl,phung2023grounded,xiao2023fastcomposer,feng2023trainingfree}. For example, 
Prompt-to-prompt~\citep{hertz2022prompt} uses attention maps from the original prompt to guide the spatial structure of the target prompt. 
Although these methods can be applied to some rich-text-to-image applications, the results often fall short, as shown in Section~\ref{sec:result}. 
Concurrent with our work, Mixture-of-diffusion~\citep{jimenez2023mixture} and MultiDiffusion~\citep{bar2023multidiffusion} propose merging multiple diffusion-denoising processes in different image regions through linear blending. 
Instead of relying on user-provided regions, we automatically compute regions of selected tokens using attention maps. Gradient~\citep{ho2022video} and Universal~\citep{bansal2023universal} guidance control the generation by optimizing the denoised generation at each time step. We apply them to precise color generation by designing an objective on the target region to be optimized.

\topic{\new{Layout-controlled image generation.}} 
\new{Spatial layouts serve as a meaningful intermediate representation, bridging the gap between abstract scene descriptions and pixel-level image synthesis. 
Generating images according to layout has been a long-standing and compelling problem~\citep{park2019semantic,johnson2018image,isola2017image,couairon2023zero,park2023learning,feng2024layoutgpt,qu2023layoutllm}. \cite{park2023learning} proposed Gaussian-categorical diffusion process to extend these frameworks by jointly generating images and semantic layouts, further enriching the generative model landscape. \cite{couairon2023zero} proposed ZestGuide, a zero-shot method that uses cross-attention-derived segmentation maps for layout-guided image generation. Similarly, \cite{qu2023layoutllm} introduced LayoutLLM-T2I to use Large Language Models (LLMs) for layout planning. They propose feedback-based sampler and relation-aware object interaction modules to enhance the quality of complex-scene image generation. \cite{feng2024layoutgpt} presented LayoutGPT, leveraging in-context learning with structured CSS-style prompts to generate layouts for both 2D images and 3D scenes. A key component in our rich-text-to-image generation framework is the \emph{region-based diffusion process}. 
Unlike these prior works, where the goal is to produce layout-guided image generation, we aim to \emph{extract the layout} from the diffusion model generation process and synthesize images with rich-text. 
}

\topic{Attention in diffusion models.}
The attention mechanism has been used in various diffusion-based applications such as view synthesis~\citep{liu2023zero,tseng2023consistent,watson2022novel}, image editing~\citep{hertz2022prompt,chefer2023attend,patashnik2023localizing,parmar2023zero,kumari2022multi}, and video editing~\citep{liu2023videop2p,qi2023fatezero,ceylan2023pix2video,ma2023directed}. 
We also leverage the spatial structure in self-attention maps and alignment information between texts and regions in cross-attention maps for rich-text-to-image generation.

\topic{Rich text modeling and application.} 
Exploiting information beyond the intrinsic meanings of the texts has been previously studied~\citep{meng2019glyce,sun2021chinesebert,xu2020layoutlm,li2022dit}. 
For example, visual information, such as underlining and bold type, have also been extracted for various document understanding tasks~\citep{xu2020layoutlm,li2022dit}. 
To our knowledge, we are the first to leverage rich text information for text-to-image synthesis.

\topic{Image stylization and colorization.}
Style transfer~\citep{gatys2016image,Zhu_2017_ICCV,Luan2017DeepPS} and Colorization~\citep{946629,1467343,Xu2013ASC,Levin2004ColorizationUO,zhang2016colorful,zhang2017real} for \emph{editing real images} have also been extensively studied.
In contrast, our work focuses on local style and precise color control for \emph{generating images} from text-to-image models.

\topic{\new{Image generation with complex prompts.}} 
\new{To accurately generate the image the users expect, one option is to provide more detailed prompts. Several studies have thus focused on generating images based on complex prompts~\citep{betker2023improving,wang2024instancediffusion,wu2023paragraph}. DALL-E 3~\citep{betker2023improving} finds that training on highly descriptive synthetic captions reliably improves the alignment between text prompts and generation results. ParaDiffusion~\citep{wu2023paragraph} utilizes pretrained LLM with a larger context window to process complex prompts. 
Instead, we decouple the complex prompts into multiple detailed prompts that describe local regions. 
Similar to ours, InstanceDiffusion~\citep{wang2024instancediffusion} also studies detailed prompts for individual regions, while ours does \emph{not} require layout as part of the user input.
}

\topic{\new{Text-to-image generation benchmark.}}\new{As text-to-image models develop rapidly, many works have paid attention to the evaluation of these models~\citep{bakr2023hrs,hu2023tifa,huang2024t2i,patel2024conceptbed,zhao2024flasheval}. While these benchmarks focus on different aspects of text-to-image generation such as the text-image alignment~\citep{hu2023tifa,bakr2023hrs} and concept learning~\citep{kumari2022multi,patel2024conceptbed}, we aim at building a benchmark for evaluating rich text to image generation on several applications, including precise color rendering, local style control, and complex prompt alignment.
}
\def\D{\altmathcal{D}}
\def\I{\altmathcal{I}}
\def\O{\altmathcal{O}}
\def\res{\altmathcal{R}}

\def\b{\boldsymbolit{b}}
\def\c{\boldsymbolit{c}}
\def\d{\boldsymbolit{d}}
\def\o{\boldsymbolit{o}}
\def\p{\boldsymbolit{p}}
\def\t{\boldsymbolit{t}}
\def\x{\boldsymbolit{x}}
\def\z{\boldsymbolit{z}}

\def\K{\boldsymbolit{K}}
\def\R{\boldsymbolit{R}}

\def\ang{\phi}
\def\dehom{\mu}
\def\proj{\pi}
\def\sigmoid{S}
\def\vis{\nu}
\def\r{\boldsymbolit{r}}

\def\bp{(\p\!)} 
\def\bt{(t\!)} 
\def\bx{(\x\neg)} 

\def\ok{\o_{\neg k}}
\def\tk{\t_{\neg k}}
\def\wk{w_{\neg k}}
\def\xi{\x_{\neg i}}
\def\zk{\z_{\neg k}}
\def\Kk{\K_{\neg k}}
\def\Rk{\R_{\neg k}}

\def\ng{\hspace{-0.1mm}}
\def\neg{\hspace{-0.2mm}}
\def\pos{\hspace{0.2mm}}

\makeatletter
\newcommand*\MY@rightharpoonupfill@{%
    \arrowfill@\relbar\relbar\rightharpoonup
}
\newcommand*\overrightharpoon{%
    \mathpalette{\overarrow@\MY@rightharpoonupfill@}%
}
\makeatother

\newlength{\depthofsumsign}
\setlength{\depthofsumsign}{\depthof{$\sum$}}
\newcommand{\nsum}[1][1.4]{
    \mathop{%
        \raisebox
            {-#1\depthofsumsign+1\depthofsumsign}
            {\scalebox
                {#1}
                {$\displaystyle\sum$}%
            }
    }
}


\begin{figure*}[t]
    \centering
    \includegraphics[width=\linewidth, trim=0 0 0 0, clip]{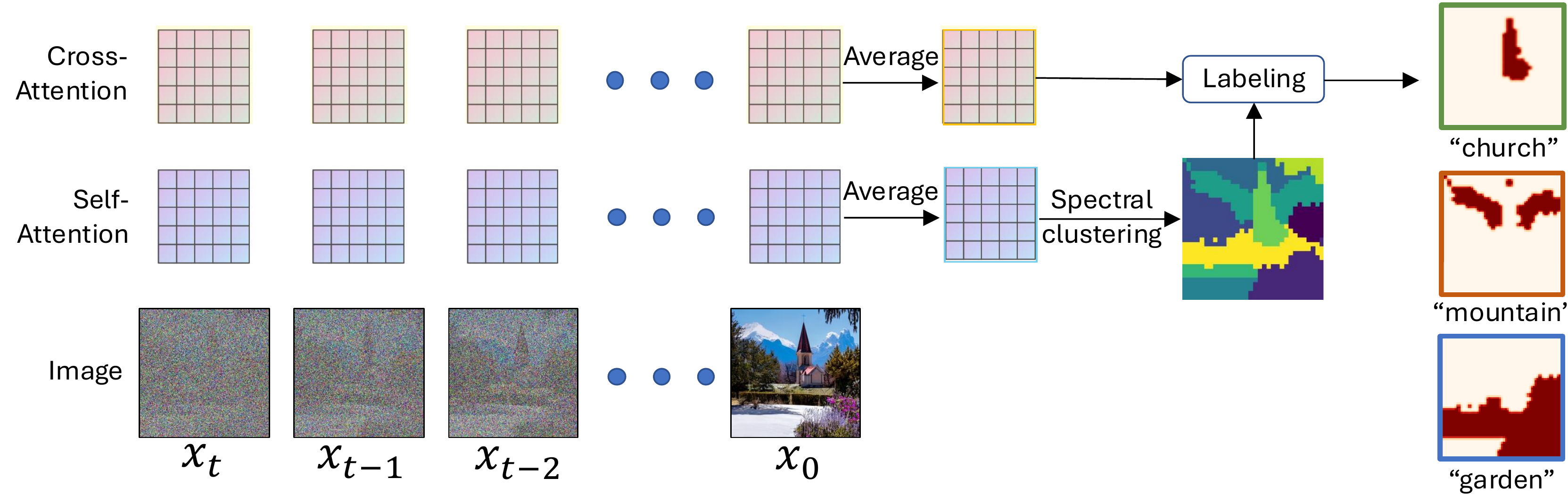}
    \caption{\new{\textbf{Token map creation.} We average the collected self- and cross-attention maps to create token maps that indicate the layout of the input prompt. The segmentation is first constructed by spectral clustering using the self-attention maps. Then, the averaged cross-attention maps are adopted to label each segment using annotated tokens.
    }}
    \label{fig:tokenmap}
\end{figure*}

\section{Rich Text to Image Generation}
\label{sec:method}

From writing messages on communication apps, designing websites~\citep{sahuguet1999wysiwyg}, to collaboratively editing a document~\citep{litt2022peritext,ignat2021enhancing}, a rich text editor is often the primary interface to edit texts on digital devices. 
\new{Such rich-text formats allow complementary information to be added to the plain texts.}
Nonetheless, only plain text has been used in text-to-image generation. 
To use formatting options in rich-text editors for more precise control over the black-box generation process~\citep{magrawala2023unpredictable}, we first introduce a problem setting called \emph{rich-text-to-image generation and editing}.
We then discuss our approach to this task. 

\subsection{Problem Setting}

As shown in Figure~\ref{fig:framework}, a rich text editor supports various formatting options, such as font styles, font size, color, and more.
We leverage these text attributes as extra information to increase control of text-to-image generation.
We interpret the rich-text prompt as JSON, where each text element consists of a span of tokens $\boldsymbol{e}_i$ (e.g., `church') and attributes $\boldsymbol{a}_i$ describing the span (e.g., `color:\#FF9900'). \new{Note that some tokens $\boldsymbol{e}_U$ may not have any attributes, and some other tokens may have multiple attributes.
Using these annotated prompts, we explore several applications: 
1) local style control using \emph{font style}, 
2) precise color control using \emph{font color}, 
3) texture rendering with \emph{texture fill},
4) detailed region description using \emph{footnotes},
5) concept guidance through \emph{embedded images}, and 
6) explicit token reweighting with \emph{font sizes}.} 


\emph{Font style} is used to apply a specific artistic style $\boldsymbol{a}^s_i$, e.g., $\boldsymbol{a}^s_i=$ `Ukiyo-e', to the synthesis of the span of tokens $\boldsymbol{e}_i$. 
For instance, in Figure~\ref{fig:teaser}, we apply the Ukiyo-e painting style to the ocean waves and the style of Van Gogh to the sky, enabling the application of localized artistic styles. 
This task presents a unique challenge for existing text-to-image models, as there are limited training images featuring multiple artistic styles. 
Consequently, existing models tend to generate a \emph{uniform} mixed style across the entire image rather than distinct local styles.


\emph{Font color} indicates a specific color of the modified text span. 
Given the prompt ``a red toy'', the existing text-to-image models generate toys in various shades of red, such as light red, crimson, or maroon. 
The color attribute provides a way for specifying a \emph{precise color} in the RGB color space, denoted as $\boldsymbol{a}^c_i$. 
For example, to generate a toy in fire brick red, one can change the font color to ``a \textcolor{lightfirebrick}{toy}'', where the word ``toy'' is associated with the attribute $\boldsymbol{a}^c_i = [178, 34, 34]$. 
However, as shown in the experiment section, the pretrained text encoder cannot interpret the RGB values and have difficulty understanding obscure color names, such as lime and orange. 

\new{\emph{Texture fill} is another way to stylize text, similar to font color, which is often known as solid fill in the rich-text editor. 
Such flexible text-fill formatting further supports finer-grained control of the object's appearance using more concise information beyond text prompts. Specifically, we use $\boldsymbol{a}^t_i$ to denote the target texture that one wants to render for certain objects. In practice, we use a texture image as the reference.}

\begin{figure*}[t]
    \centering
    \includegraphics[width=\linewidth, trim=0 0 0 0, clip]{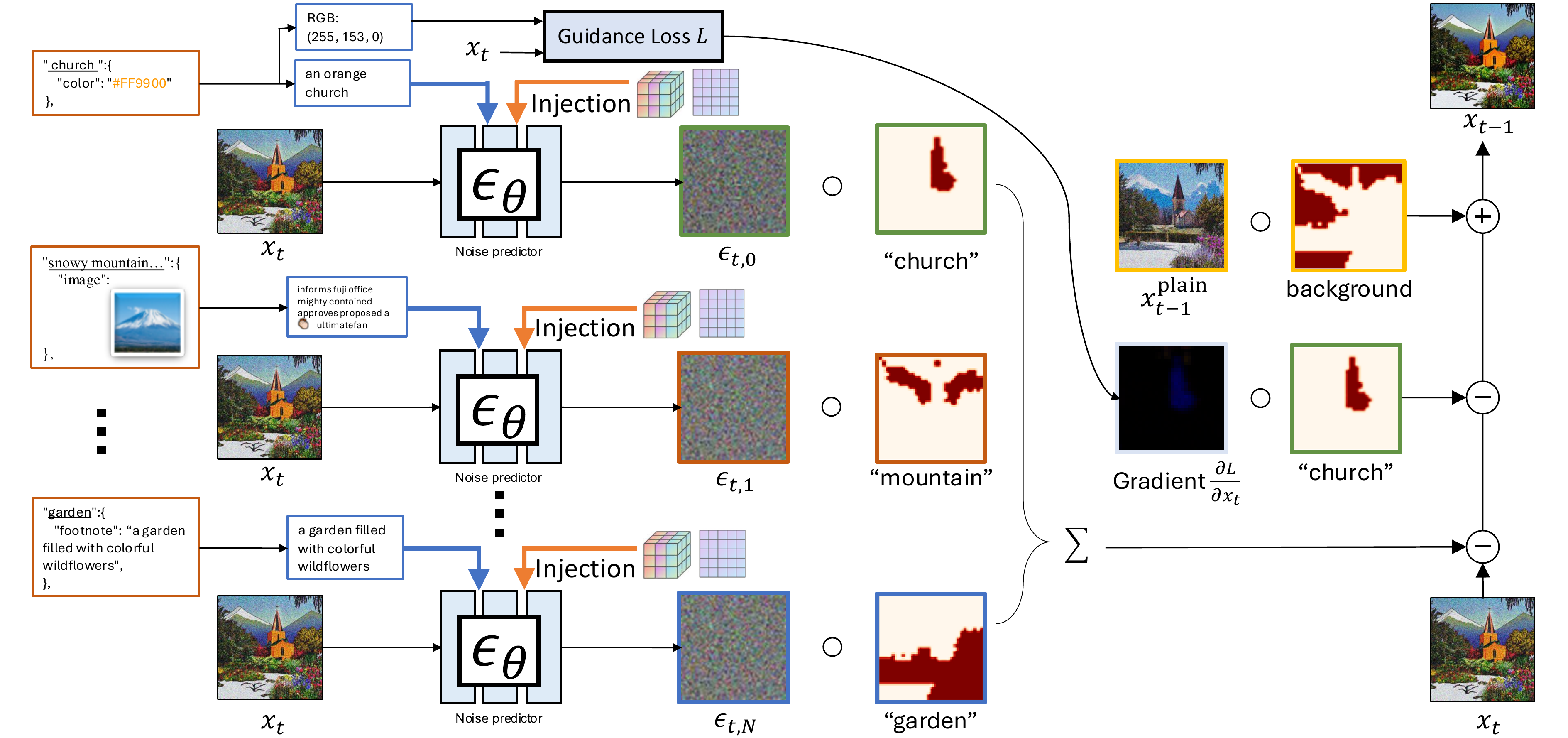}
    \caption{\new{\textbf{Region-based diffusion.} 
    We fulfill the guidance specified by the rich-text attributes through separate diffusion processes. 
    Depending on the functionality, the attributes are either interpreted as a region-based guidance target (e.g. re-coloring the church), or as a textual input to the diffusion UNet (e.g. handling the embedded image describing the snowy mountain). The self-attention maps and feature maps extracted from the plain-text generation process are injected to help preserve the structure.} The predicted noise $\epsilon_{t,\boldsymbol{e}_i}$, weighted by the token map, and the guidance gradient $\frac{\partial{L}}{\partial{\mathbf{x}_t}}$ are used to denoise and update the previous generation $\mathbf{x}_t$ to $\mathbf{x}_{t-1}$. The noised plain text generation $\mathbf{x}^\text{plain}_t$ is blended with the current generation to preserve the exact content in those regions of the unformatted tokens.
    }
    \label{fig:method}
\end{figure*}

\emph{Footnote} provides supplementary explanations of the target span without hindering readability with lengthy sentences. 
Writing detailed descriptions of complex scenes is tedious work, and it inevitably creates lengthy prompts~\citep{karpathy2015deep,johnson2016densecap}. 
Additionally, existing text-to-image models are prone to ignoring some objects when multiple objects are present~\citep{chefer2023attend}, especially with long prompts.
Moreover, excess tokens are discarded when the prompt's length surpasses the text encoder's maximum length, e.g., 77 tokens for CLIP models~\citep{radford2021learning}. 
We aim to mitigate these issues using a footnote string $\boldsymbol{a}^f_i$. 



\emph{Font size} can indicate an object's importance, quantity, or size.
We use a scalar $\boldsymbol{a}^w_i$ to denote the weight of each token. 

\new{\emph{Embedded} images provide visual cues complementary to the textual information.
Such visual information is more accurate than texts when describing the objects on the identity and low-level details. For example, by connecting the word ``cat'' in ``a cat chasing a butterfly'' to a specific cat image, one can generate an image of their cat chasing a butterfly. 
Generating customized images of the reference concepts has been an active research question~\citep{kumari2022multi,gal2023an,chen2023anydoor,li2023blip,jia2023taming,chen2023subject,gal2023encoder,shi2023instantbooth,chen2023anydoor,xiao2023fastcomposer}. 
In contrast to free-form generation in prior work, we aim to preserve the original image structure in plain-text generation while synthesizing customized concepts.}

\subsection{Method}

To utilize rich text annotations, our method consists of two steps, as shown in Figure~\ref{fig:framework}.
First, we compute the spatial layouts of individual token spans. 
Second, we use a new region-based diffusion to render each region's attributes into a globally coherent image.

\vspace{-2mm}
\topic{Step 1.~Token maps for spatial layout.} 
Several works \citep{Tang2022WhatTD,ma2023directed,balaji2022ediffi,hertz2022prompt,chefer2023attend,patashnik2023localizing,tumanyan2022plug} have discovered that the attention maps in the self- and cross-attention layers of the diffusion UNet characterize the spatial layout of the generation. 
\new{As shown in Figure~\ref{fig:tokenmap}, we develop a method to precisely extract the layout of the diffusion model generation. Our intuition is that the overall structure of image is determined by the plain text prompt while the rich-text formatting offers more informative descriptions of the local regions.}

We use the plain text as the input to the diffusion model and collect self-attention maps of size $32\times 32\times 32\times 32$ across different heads, layers, and time steps. We take the average across all the extracted maps and reshape the result into $1024 \times 1024$. Note that the value at $i^{th}$ row and $j^{th}$ column of the map indicates the probability of pixel $i$ attending to pixel $j$. We average the map with its transpose to convert it to a symmetric matrix. It is used as a similarity map to perform spectral clustering~\citep{868688,von2007tutorial} and obtain the binary segmentation maps $\mathbf{\widehat{M}}$ of size $K\times 32\times32$, where $K$ is the number of segments.

To associate each segment with a textual span, we also extract cross-attention maps for each token $w_j$:
\begin{equation}
\mathbf{m}_j = \frac{\exp(\mathbf{s}_j)}{\sum_k \exp(\mathbf{s}_k)},
\end{equation}
where $\mathbf{s}_j$ is the attention score.
We first interpolate each cross-attention map $\mathbf{m}_j$ 
to the same resolution as $\mathbf{\widehat{M}}$ of $32\times32$. 
Similar to the processing steps of the self-attention maps, we compute the mean across heads, layers, and time steps to get the averaged map $\mathbf{\widehat{m}}_j$.
We associate each segment with a texture span $\boldsymbol{e}_i$ following ~\cite{patashnik2023localizing}:
\begin{align}
\small
\mathbb{M}_{\boldsymbol{e}_i} = \{\mathbf{\widehat{M}}_k \: \left.\right\vert & \left\vert \: \mathbf{\widehat{M}}_k\cdot \frac{\mathbf{\widehat{m}}_j-\min(\mathbf{\widehat{m}}_j)}{\max(\mathbf{\widehat{m}}_j)-\min(\mathbf{\widehat{m}}_j)}\right\vert_1 > \epsilon,  
\\ &\forall j \; s.t. \; w_j \in \boldsymbol{e}_i\},
\end{align}
where $\epsilon$ is a hyperparameter that controls the labeling threshold, that is, the segment $\mathbf{\widehat{M}}_k$ is assigned to the span $\boldsymbol{e}_i$ if the normalized attention score of any tokens in this span is higher than $\epsilon$. We associate the segments unassigned to any formatted spans with the unformatted tokens $\boldsymbol{e}_U$. Finally, we obtain the \textit{token map} in Figure~\ref{fig:framework} as below:
\begin{equation}
\mathbf{M}_{\boldsymbol{e}_i} = \frac{\sum_{\mathbf{\widehat{M}}_j\in \mathbb{M}_{\boldsymbol{e}_i}}\mathbf{\widehat{M}}_j}{\sum_i \sum_{\mathbf{\widehat{M}}_j\in \mathbb{M}_{\boldsymbol{e}_i}} \mathbf{\widehat{M}}_j}
\label{eqn:map}
\end{equation}

\begin{figure*}[t]
    \centering
    \includegraphics[width=\linewidth, trim=0 0 0 0, clip]{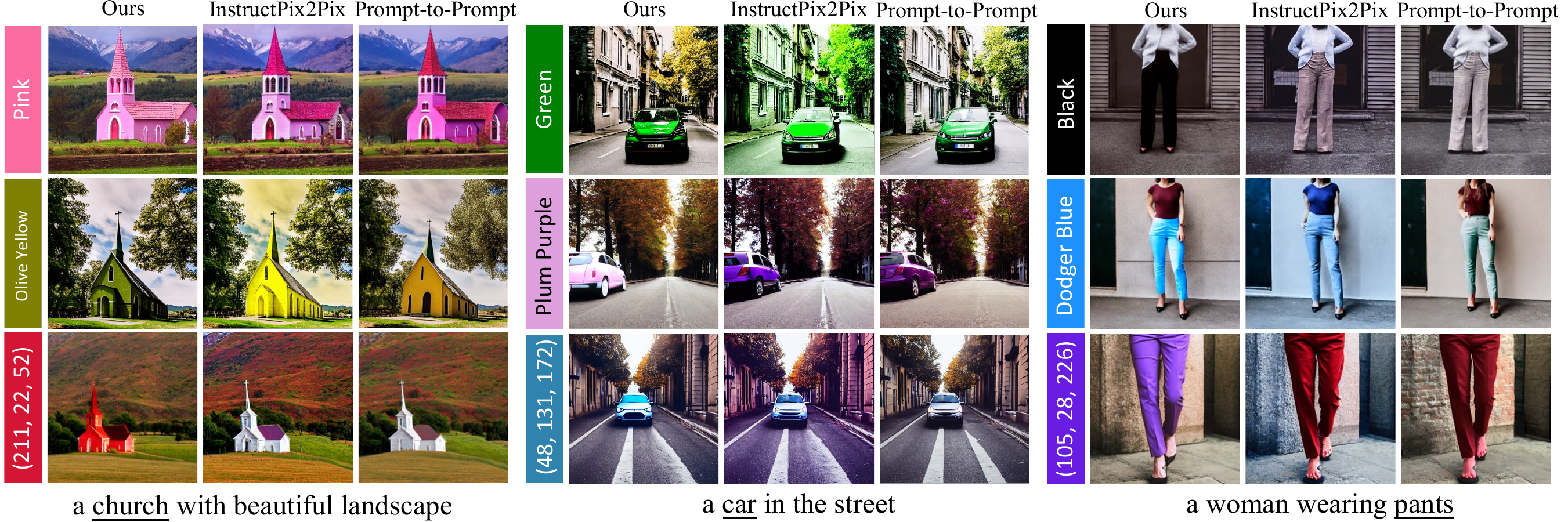}
    \caption{\textbf{Qualitative comparison on precise color generation.} We show images generated by Prompt-to-Prompt \citep{hertz2022prompt}, InstructPix2Pix~\citep{brooks2022instructpix2pix}, and our method using prompts with font colors. Our method generates precise colors according to either color names or RGB values. Both baselines generate plausible but inaccurate colors given color names, while neither understands the color defined by RGB values. InstructPix2Pix tends to apply the color globally, even outside the target object. 
    }
    \label{fig:qualitative_color}
\end{figure*}

\vspace{-2mm}
\topic{Step 2.~Region-based denoising and guidance.} As shown in Figure~\ref{fig:framework}, given the text attributes and \textit{token maps}, we divide the overall image synthesis into several region-based denoising and guidance processes to incorporate each attribute, {similar to}  an ensemble of diffusion models~\citep{kumari2022multi,bar2023multidiffusion}. 
More specificially, given the span $\boldsymbol{e}_i$, the region defined by its \textit{token map} $\mathbf{M}_{\boldsymbol{e}_i}$, {and the attribute $\boldsymbol{a}_i$}, the predicted noise $\boldsymbol{\epsilon}_t$ for noised generation $\mathbf{x}_t$ at time step $t$ is 
\begin{equation}
\boldsymbol{\epsilon}_t = \sum_i \mathbf{M}_{\boldsymbol{e}_i} \cdot \epsilon_{t,\boldsymbol{e}_i} = \sum_i \mathbf{M}_{\boldsymbol{e}_i} \cdot D(\mathbf{x}_t, f(\boldsymbol{e}_i, \boldsymbol{a}_i), t),
\label{eqn:noise}
\end{equation}
where $D$ is the pretrained diffusion model, and $f(\boldsymbol{e}_i, \boldsymbol{a}_i)$ is a plain text representation derived from text span $\boldsymbol{e}_i$ and attributes $a_i$ using the following process: 
\begin{enumerate}
    \item Initially, we set $f(\boldsymbol{e}_i, \boldsymbol{a}_i)=\boldsymbol{e}_i$.
    \item \new{If an embedded image is available, we convert it into a footnote $\boldsymbol{a}^f_i$ using the gradient-based discrete optimization~\citep{wen2023hard}.}
    \item If footnote $\boldsymbol{a}^f_i$ is available, we set $f(\boldsymbol{e}_i, \boldsymbol{a}_i) = \boldsymbol{a}^f_i$. 
    \item The style $\boldsymbol{a}^s_i$ is appended if it exists. $f(\boldsymbol{e}_i, \boldsymbol{a}_i) \mathrel= f(\boldsymbol{e}_i, \boldsymbol{a}_i) +\text{`in the style of'} + \boldsymbol{a}^s_i$.
    \item The closest color name (string) of font color $\hat{\boldsymbol{a}}^c_i$ from a predefined set $\mathcal{C}$ is prepended.  
    $f(\boldsymbol{e}_i, \boldsymbol{a}_i) =\hat{\boldsymbol{a}}^c_i + f(\boldsymbol{e}_i, \boldsymbol{a}_i)$. For example, $\hat{\boldsymbol{a}}^c_i$ = `brown' for RGB color $\boldsymbol{a}^c_i$ = [136,68,20]. 
    \item \new{If a texture description $\hat{\boldsymbol{a}}^t_i$ is provided together with the texture image $\boldsymbol{a}^c_i$, we prepend the texture description: $f(\boldsymbol{e}_i, \boldsymbol{a}_i) =\hat{\boldsymbol{a}}^t_i + f(\boldsymbol{e}_i, \boldsymbol{a}_i)$}.
\end{enumerate}
We use $f(\boldsymbol{e}_i, \boldsymbol{a}_i)$ as the original plain text prompt of Step 1 for the unformatted tokens $\boldsymbol{e}_U$.
This helps us generate a coherent image, especially around region boundaries. 

\topic{Guidance.} By default, we use classifier-free guidance~\citep{ho2022classifier} for each region to better match the prompt $f(\boldsymbol{e}_i, \boldsymbol{a}_i)$. \new{In addition, if the font color or texture fill is specified, to exploit the appearance information further, we apply gradient guidance~\citep{ho2022video,NEURIPS2021_49ad23d1,bansal2023universal} on the current clean image prediction}: 
\begin{equation}
\widehat{\mathbf{x}}_0 = \frac{\mathbf{x}_t-\sqrt{1-\bar{\alpha}_t}\boldsymbol{\epsilon}_t}{\sqrt{\bar{\alpha}_t}}, 
\end{equation}
where $\mathbf{x}_t$ is the noisy image at time step $t$, and $\bar{\alpha}_t$ is the coefficient defined by noise scheduling strategy~\citep{ho2020denoising}.  
\new{When font color is specified, we compute an MSE loss $\mathcal{L}_\text{color}$ between the average color of $\widehat{\mathbf{x}}$ weighted by the \textit{token map} $\mathbf{M}_{\boldsymbol{e}_i}$ and the RGB triplet $\boldsymbol{a}^c_i$. }
The gradient is calculated below,
\begin{equation}
\label{eq:color_grad}
\frac{d{\mathcal{L}_\text{color}}}{d{\mathbf{x}_t}} = \frac{d \|\sum_p{(\mathbf{M}_{\boldsymbol{e}_i} \cdot \widehat{\mathbf{x}}_0)}/\sum_p{\mathbf{M}_{\boldsymbol{e}_i}}-\boldsymbol{a}^c_i\|_2^2}{\sqrt{\bar{\alpha}_t}d{\widehat{\mathbf{x}}_0}},
\end{equation}
where the summation is over all pixels $p$. 

\new{When texture fill is selected, we adopt a region-based perceptual loss~\citep{johnson2016perceptual}, $\mathcal{L}_\text{perceptual}$, to optimize the region specified by \textit{token map} $\mathbf{M}_{\boldsymbol{e}_i}$ to align with the guided texture:}

\new{\begin{equation}
\label{eq:perceptual}
\mathcal{L}_\text{perceptual}(\mathbf{x}, \mathbf{y})=\sum_j{\|G_j^\text{VGG}(\mathbf{x})-G_j^\text{VGG}(\mathbf{y})}\|_2^2,
\end{equation}}

\new{where $G_j^\text{VGG}(\cdot)$ computes the Gram matrix of neural features extracted from a VGG model at layer $j$. Specifically, we mask the image with $\mathbf{M}_{\boldsymbol{e}_i}$ and fill the hole with the texture image to ensure that the loss is only computed on the token region. The guidance gradient is then computed as:}

\new{\begin{equation}
\label{eq:texture_grad}
\frac{d{\mathcal{L}_\text{texture}}}{d{\mathbf{x}_t}} = \frac{d \mathcal{L}_\text{perceptual}(\mathbf{M}_{\boldsymbol{e}_i} \cdot \mathbf{x}_t+(1-\mathbf{M}_{\boldsymbol{e}_i}) \cdot  \mathbf{a}^t_i, \mathbf{a}^t_i)}{\sqrt{\bar{\alpha}_t}d{\widehat{\mathbf{x}}_0}},
\end{equation}}

\new{We denote the total guidance loss as $\mathcal{L}=\lambda_\text{texture}\mathcal{L}_\text{texture}+\lambda_\text{color}\mathcal{L}_\text{texture}$, where $\lambda_\text{texture}$ and $\lambda_\text{color}$ are hyperparameters to control the strength of the guidance.  We use $\lambda_\text{color}=1$ and $\lambda_\text{texture}=0.2$ unless denoted otherwise.}We then update $\mathbf{x}_t$ with the following equation:
\begin{equation}
    \mathbf{x}_t \leftarrow \mathbf{x}_t - \lambda \cdot \mathbf{M}_{\boldsymbol{e}_i} \cdot \frac{d{\mathcal{L}}}{d{\mathbf{x}_t}},
\end{equation}

\topic{Token reweighting with font size.} 
Last, to re-weight the impact of the token $w_j$ according to the font size $\boldsymbol{a}^w_j$, we modify its cross-attention maps $\mathbf{m}_j$. 
However, instead of applying direct multiplication as in Prompt-to-Prompt~\citep{hertz2022prompt} where $\sum_j\boldsymbol{a}^w_j\mathbf{m}_j\ne 1$, we find that it is critical to preserve the probability property of $\mathbf{m}_j$.
We thus propose the following reweighting approach: 
\begin{equation}
\widehat{\mathbf{m}}_j = \frac{\boldsymbol{a}^w_j\exp(\mathbf{s}_j)}{\sum_k \boldsymbol{a}^w_k\exp(\mathbf{s}_k)}.
\end{equation}

We can compute the token map (Equation~\ref{eqn:map}) and predict the noise (Equation~\ref{eqn:noise}) with the reweighted attention map. 

\begin{figure*}[t!]
    \centering
    \includegraphics[width=\linewidth, trim=0 0 0 0, clip]{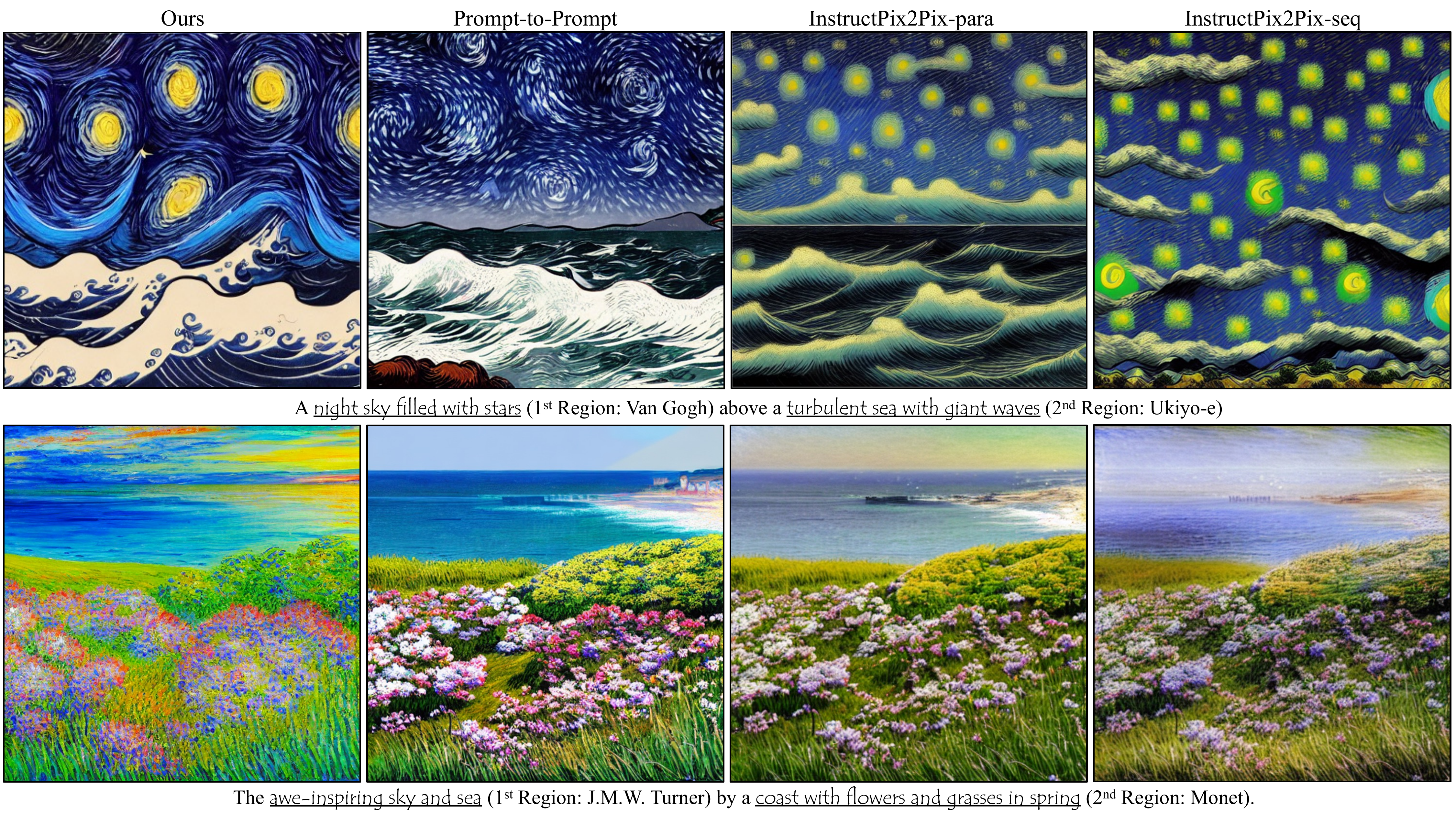}
    \caption{\textbf{Qualitative comparison on style control.} We show images generated by Prompt-to-Prompt, InstructPix2Pix, and our method using prompts with multiple styles. Only our method can generate distinct styles for both regions.}
    \label{fig:qualitative_style}
\end{figure*}

\begin{figure}[t]
\scriptsize
\centering
\begin{tikzpicture}
\begin{axis}[
    ybar,
    bar width=9pt,
    ylabel={CLIP Similarity ($\uparrow$)},
    ymin=0.22,
    ymax=0.286,
    ytick={0.24,0.26,0.28},
    ylabel near ticks,
    ymajorgrids=true,
    legend style={at={(0.45,1.1)},anchor=south},
    legend columns=2, 
    legend image code/.code={
        \draw [#1] (0cm,-0.1cm) rectangle (0.1cm,0.11cm); },
    symbolic x coords={$1^{\text{st}}$ Region, $2^{\text{nd}}$ Region, Both},
    nodes near coords,
    nodes near coords style={
        font=\tiny,/pgf/number format/.cd,fixed,precision=2
    },
    xtick=data,
    major x tick style = transparent,
    enlarge x limits=0.25,
    width=75mm,
    height=40mm,
    ]
    
\addplot [color=lightfirebrick,fill=firebrick] coordinates {($1^{\text{st}}$ Region, 0.271) ($2^{\text{nd}}$ Region, 0.280) (Both, 0.276)};
\addplot [color=lightgrayblue,fill=grayblue] coordinates {($1^{\text{st}}$ Region, 0.268) ($2^{\text{nd}}$ Region, 0.249) (Both, 0.259)};
\addplot [color=lightorange,fill=darkorange] coordinates {($1^{\text{st}}$ Region, 0.263) ($2^{\text{nd}}$ Region, 0.237) (Both, 0.249)};
\addplot [color=lightgreen,fill=darkgreen] coordinates {($1^{\text{st}}$ Region, 0.264) ($2^{\text{nd}}$ Region, 0.243) (Both, 0.254)};

\legend{Ours, Prompt-to-Prompt, InstructPix2Pix-seq, InstructPix2Pix-para}
\end{axis}
\end{tikzpicture}
\centering
\caption{\textbf{Quantitative evaluation of local style control.} We report the CLIP similarity between each stylized region and its region prompt. Our method achieves the best stylization.}
\vspace{-1pt}
\label{tab:style_quantitative}
\end{figure}
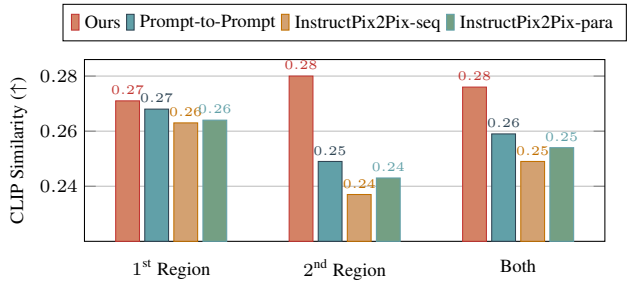

\topic{Preserve the fidelity against plain-text generation.} Although our region-based method naturally maintains the layout, there is no guarantee that the details and shape of the objects are retained when no rich-text attributes or only the color is specified, as shown in Figure~\ref{fig:ablation5}. To this end, we follow Plug-and-Play~\citep{tumanyan2022plug} to inject the self-attention maps and the residual features extracted from the plain-text generation process when $t>T_\text{pnp}$ to improve the structure fidelity. In addition, for the regions associated with the unformatted tokens $\boldsymbol{e}_U$, stronger content preservation is desired. Therefore, at certain $t=T_\text{blend}$, we blend the noised sample $\mathbf{x}^\text{plain}_t$ based on the plain text into those regions:
\begin{equation}
    \mathbf{x}_t \leftarrow \mathbf{M}_{\boldsymbol{e}_U}\cdot \mathbf{x}^\text{plain}_t + (1-\mathbf{M}_{\boldsymbol{e}_U})\cdot \mathbf{x}_t
\end{equation}


\section{Experimental Results}
\label{sec:result}

\begin{figure*}[t]
    \centering
    \includegraphics[width=\linewidth, trim=0 0 0 0, clip]{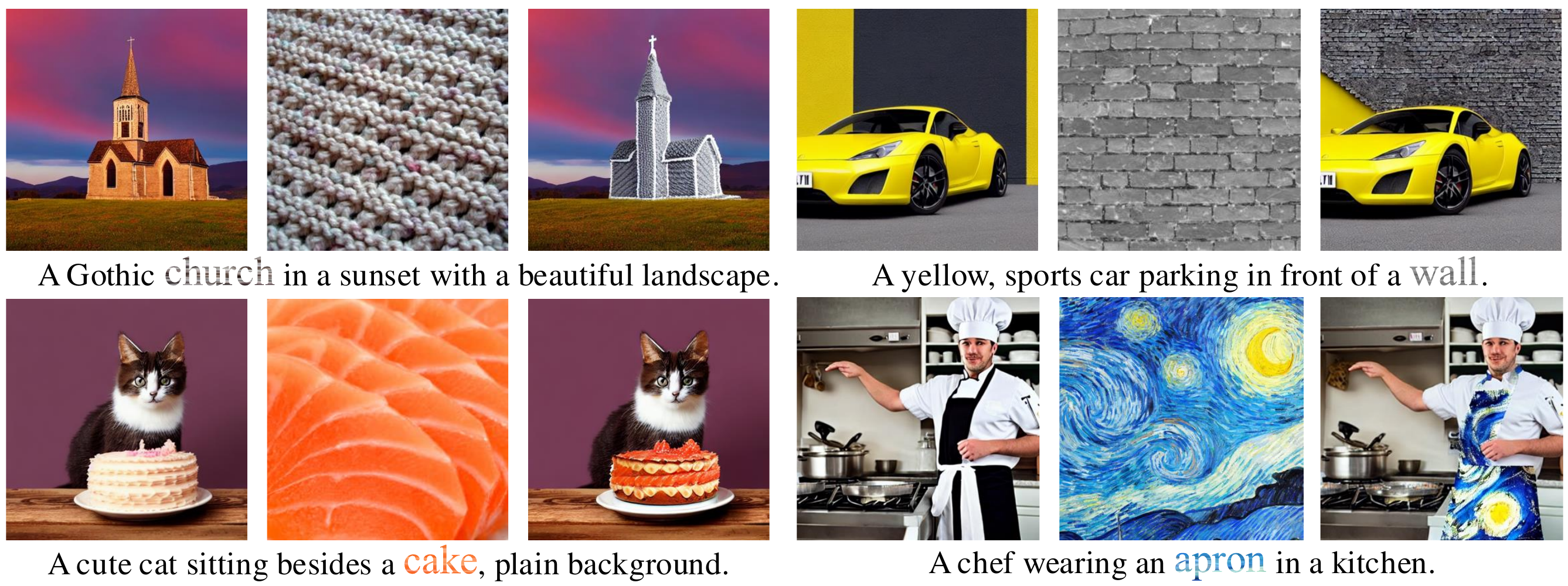}
    \caption{\new{\textbf{Qualitative results on texture transfer.} We show images our method generates with certain textures specified in the rich-text prompt by texture infill. We present the plain-text generation, reference texture image, and rich-text generation for each example. Note that we resize texture images to a similar size for display. Our method can faithfully align with the texture guidance while organically combining it with the rest of the generated image.}}
    \label{fig:qualitative_texture}
    \vspace{-5pt}
\end{figure*}

\begin{figure}[t]
\scriptsize
\centering
\begin{subfigure}{0.25\textwidth}
\centering
\begin{tikzpicture}
\begin{axis}[
    ybar,
    bar width=7pt,
    xlabel={Minimal Distance},
    xlabel near ticks,
    ylabel={Distance to Target Color ($\downarrow$)},
    ymin=0,
    ymax=0.175,
    ytick={0,0.1},
    ylabel near ticks,
    ymajorgrids=true,
    legend style={at={(1.1,1.1)},anchor=south},
    legend columns=3, 
    legend image code/.code={
        \draw [#1] (0cm,-0.1cm) rectangle (0.1cm,0.11cm); },
    symbolic x coords={Common, HTML, RGB},
    nodes near coords,
    nodes near coords style={
        font=\tiny,/pgf/number format/.cd,fixed,precision=2
    },
    xtick=data,
    major x tick style = transparent,
    enlarge x limits=0.25,
    width=46mm,
    height=38mm,
    ]
    
\addplot [color=lightfirebrick,fill=firebrick] coordinates {(Common, 0.021) (HTML, 0.043) (RGB, 0.057)};
\addplot [color=lightgrayblue,fill=grayblue] coordinates {(Common, 0.134) (HTML, 0.109) (RGB, 0.124)};
\addplot [color=lightorange,fill=darkorange] coordinates {(Common, 0.069) (HTML, 0.070) (RGB, 0.142)};
    
\legend{Ours, Prompt-to-Prompt, InstructPix2Pix}
\end{axis}
\end{tikzpicture}
\end{subfigure}%
\begin{subfigure}{0.247\textwidth}
\centering
\begin{tikzpicture}
\begin{axis}[
    ybar,
    bar width=7pt,
    xlabel={Mean Distance},
    xlabel near ticks,
    ymin=0.3,
    ymax=0.8,
    ytick={0.4,0.6, 0.8},
    ylabel near ticks,
    ymajorgrids=true,
    symbolic x coords={Common, HTML, RGB},
    nodes near coords,
    nodes near coords style={
        font=\tiny,/pgf/number format/.cd,fixed,precision=2
    },
    xtick=data,
    major x tick style = transparent,
    enlarge x limits=0.25,
    width=46mm,
    height=38mm,
    ]
    
\addplot [color=lightfirebrick,fill=firebrick] coordinates {(Common, 0.377) (HTML, 0.418) (RGB, 0.459)};
\addplot [color=lightgrayblue,fill=grayblue] coordinates {(Common, 0.713) (HTML, 0.675) (RGB, 0.730)};
\addplot [color=lightorange,fill=darkorange] coordinates {(Common, 0.697) (HTML, 0.652) (RGB, 0.712)};
\end{axis}
\end{tikzpicture}
\end{subfigure}
\centering
\caption{\textbf{Quantitaive evaluation on precise color generation.} Distance against target color is reported (lower is better). Our method consistently outperforms baselines. 
}
\label{fig:color_quantitative}
\end{figure}
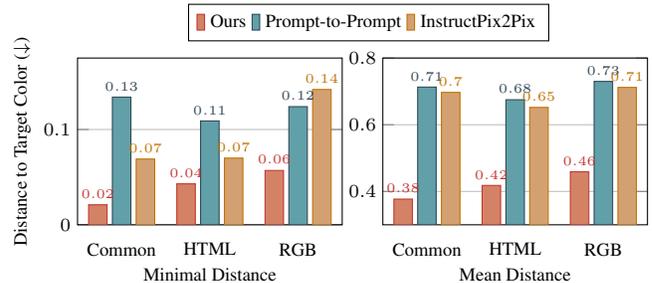


\begin{figure*}[t]
    \centering
    \includegraphics[width=\linewidth, trim=0 0 0 0, clip]{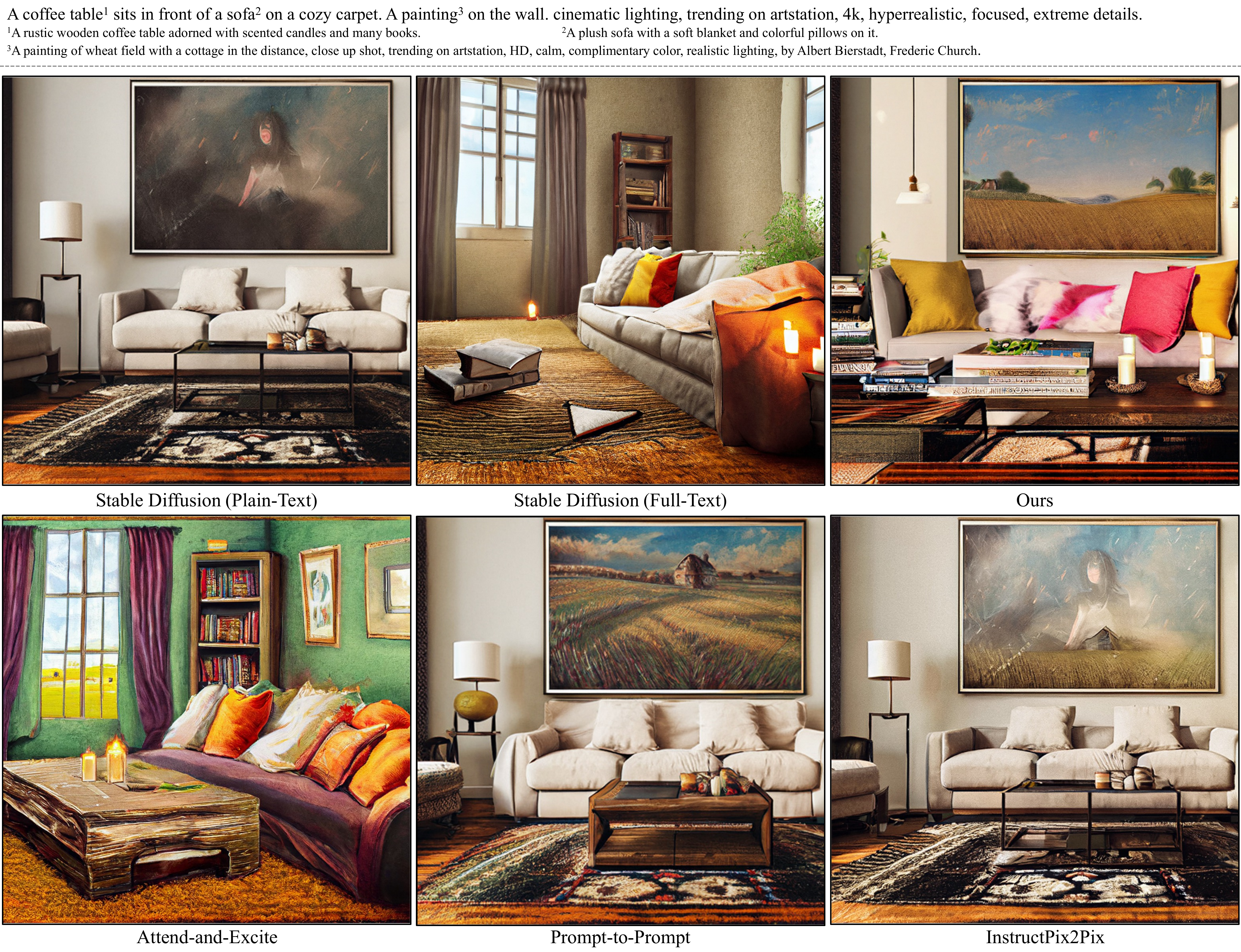}
    \caption{\textbf{Qualitative comparison on detailed description generation.} We show images generated by Attend-and-Excite, Prompt-to-Prompt, InstructPix2Pix, and our method using complex prompts. Our method is the only one that can generate all the details faithfully.}
    \label{fig:qualitative_footnote}
\end{figure*}

\subsection{Experimental Setups}
\topic{Implementation details.} 
\label{sec:setup} 
We use Stable Diffusion V1-5~\citep{rombach2022high} for our main comparisons, as most baselines are built on this model. We also demonstrate the generalizability of the method using SDXL~\citep{podell2024sdxl}, which achieves better-quality and higher-resolution generation results over SD.
To create the token maps, we use the cross-attention layers in all blocks, excluding the first encoder and last decoder blocks, as the attention maps in these high-resolution layers are often noisy. 
We discard the maps at the initial denoising steps with $T > 750$. 
We use $K=15$, $\epsilon=0.3, T_\text{pnp}=0.3, T_\text{blend}=0.3$, and report the results averaged from three random seeds for all quantitative experiments. 
More details, such as the running time, can be found in Appendix B.
We notice that most hyperparameters we use in SD work well for SDXL. Notably, SDXL additionally conditions the generation on the pooled CLIP feature~\citep{radford2021learning} by using adaptive normalization. For rich-text generation, we use region-specific prompts to compute these features for region-based diffusion process. In addition, we use the residual features from the first layer of the decoder for the Plug-and-Play method.

\topic{Baselines.} For font color and style, we quantitatively compare our method with two strong baselines, Prompt-to-Prompt \citep{hertz2022prompt} and InstructPix2Pix \citep{brooks2022instructpix2pix}. When two instructions exist for each image in our font style experiments,  we apply them in parallel (InstructPix2Pix-para) and sequential manners (InstructPix2Pix-seq).
More details can be found in Appendix B. We also perform a human evaluation on these two methods in Appendix Table 1.
For re-weighting token importance, we visually compare with Prompt-to-Prompt \citep{hertz2022prompt} and two heuristic methods, repeating and adding parentheses. 
For complex scene generation with footnotes, we also compare with Attend-and-Excite \citep{chefer2023attend}. 

\begin{figure*}[t]
    \centering
    \includegraphics[width=\linewidth, trim=0 0 0 0, clip]{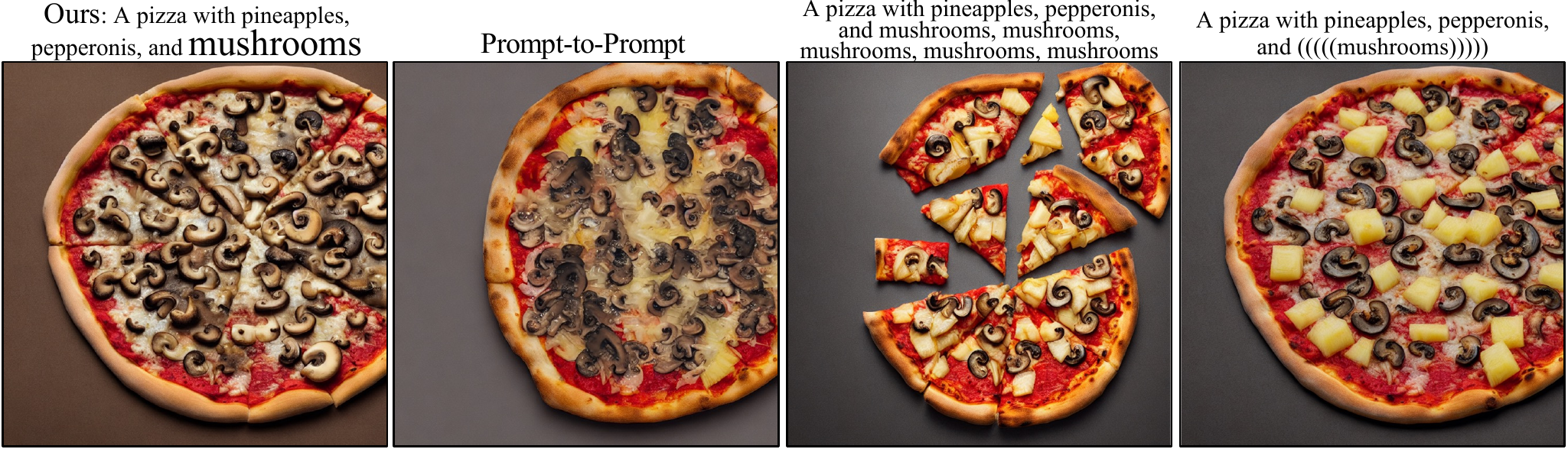}
    \caption{\textbf{Qualitative comparison on token reweighting.} We show images generated by our method and Prompt-to-Prompt using token weight of $13$ for `mushrooms'. Prompt-to-Prompt suffers from artifacts due to the large weight. Heuristic methods like repeating and parenthesis do not work well.}
    \label{fig:qualitative_size}
    \vspace{-5pt}
\end{figure*}
\begin{figure*}[t]
    \centering
    \includegraphics[width=\linewidth, trim=0 0 0 0, clip]{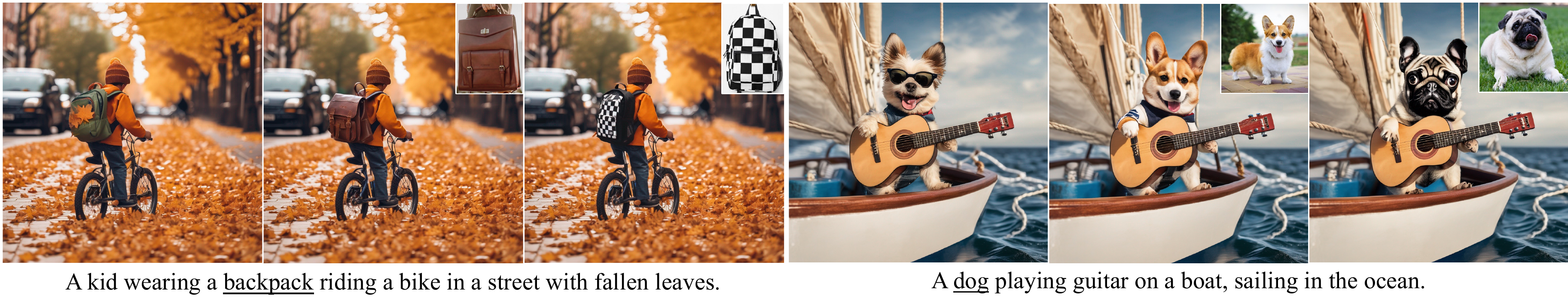}
    \vspace{-10pt}
    \caption{\new{\textbf{Qualitative results on customized concept generation.} We underline the modified concepts in the prompts and display the reference image on the top right of each rich-text generation result. Our method is able to synthesize the image with the object according to the reference image without changing the overall structure.}}
    \label{fig:qualitative_embeds}
    \vspace{-5pt}
\end{figure*}

\subsection{Rich-Text Benchmark}
Most existing studies on text-to-image generation~\citep{DALLE,Imagen,hertz2022prompt} are only evaluated on the relatively short prompts~\citep{bakr2023hrs,hu2023tifa} with a single image style or simple colors. 
To evaluate the model capacity on the challenging rich-text applications, including local style rending, precise color synthesis, and complex prompt alignment, we build a benchmark with pairs of rich-text prompts and their equivalent plain-text prompts focusing on these tasks. For each task, we also design the pipeline to automatically assess the critical aspects of generation quality. More details about the construction process is explained in Appendix B.

\topic{Footnote.} 
To understand how well a model performs when a lengthy text prompt is given, we collect a set of long prompts using the GPT-4 model~\citep{openai2023gpt4}. 
We follow a two-step procedure for prompting the language model. First, we ask the model to generate global descriptions for a scene and then generate detailed captions for the objects that appear in the scene. 
Such a format also makes it easy to convert to rich-text prompts with footnotes. 
Second, we create the full plain and rich text prompts based on these global and local descriptions.

In the first step, to control the quality of the collected prompts, we manually create a few hierarchical descriptions as the in-context examples. In addition, we find that the model can hallucinate in the local object descriptions by describing the objects not shown in the scene. Also, some descriptions can include non-visual features like the flavor of the food. Therefore, after collecting the results from GPT-4, we manually filter and edit the low-quality text prompts.

We collect $100$ prompts in total. 
In the second step, we use GPT-4 to produce single, long, plain-text prompts. We also create scripts to produce rich-text prompts from these hierarchical prompts. We provide the prompt template we used for both stages in Appendix B. We summarize the statistics of the plain-text prompts in Table~\ref{tab:stats}. 
Note that the average length of the collected prompts is around $4\times$ longer than the prior arts and closer to the concurrent work~\citep{wu2023paragraph}.

To evaluate the alignment between the generated images and the lengthy text prompts, we follow the process in TIFA~\citep{hu2023tifa} to sample question-answer pairs according to the scene and object prompts using language models. 
For example, given the scene prompt ``a nightstand next to a bed in the bedroom,'' the question-and-answer pair could be ``Q: Is there a bed? A: Yes.'' 
Through this process, we collect $2921$ such pairs. 
To evaluate a generated image, either a VQA model or human annotators can answer the question based on the generated image and check whether it is consistent with the ground truth answer. 

We manually modify the questions based on the object prompts to reflect that the object is part of the scene. For example, for the object prompt ``a nightstand with some books,'' we modify the question ``Are there books?'' to ``Are there books on the nightstand?'' 
We manually review and drop low-quality, irrelevant, or repeated questions. 
For example, we drop the pair ``Q: What is in the bedroom? A: A bed.'' since ``a nightstand'' or other reasonable objects could be the correct answer. Instead, we prefer to ask the more concrete question, ``Is there a bed in the bedroom?'' 
After the manual processing, we obtain $1974$ question-answer pairs in total.

\topic{Font color.} 
We divide colors into three categories to evaluate a method's capacity to understand and generate a specific color. 
The \emph{Common color} category contains 17 standard names, such as ``red'', ``yellow'', and ``pink''. 
The \emph{HTML color} names are selected from the web color names\footnote{\url{https://simple.wikipedia.org/wiki/Web_color}} used for website design, such as ``sky blue'', ``lime green'', and ``violet purple''. 
The \emph{RGB color} category contains 50 randomly sampled RGB triplets
to be used as ``color of RGB values $[128, 128, 128]$''. 
To create a complete prompt, we use 12 objects exhibiting different colors, such as ``flower'', ``gem'', and ``house''. 
This gives us a total of $1,200$ prompts. 
We evaluate color accuracy by computing the mean L2 distance between the region and target RGB values. 
We also compute the minimal L2 distance as sometimes the object should contain other colors for fidelity, e.g., the ``black tires'' of a ``yellow car''.

\topic{Font style.} 
We come up with prompts featuring two objects and styles to evaluate the ability to generate accurate local styles. 
We create combinations using $7$ popular styles and $10$ objects, resulting in $420$ prompts. 
To automatically evaluate the performance, we compute CLIP scores~\citep{radford2021learning} for each local region to evaluate the stylization quality. 
Specifically, for each generated image, we mask it by the token maps of each object and attach the masked output to a black background. 
Then, we compute the CLIP score using the region-specific prompt. 
For example, for the prompt ``a \underline{lighthouse} (Cyberpunk) among the turbulent \underline{waves} (Ukiyo-e)'', the local CLIP score of the lighthouse region is measured by comparing its similarity with the prompt ``lighthouse in the style of cyberpunk.'' 
In this example, we refer to ``lighthouse'' as the first region and ``waves'' as the second region.

\begin{table*}[th]
\caption{\new{The statistics of text prompts collected in the previous benchmarks or adopted in the previous studies.}}
\centering
\label{tab:stats}
\begin{tabular}{lllllll}
\toprule
\multirow{2}{*}{Benchmark or paper} & \multicolumn{3}{c}{Number of words}                                         & \multicolumn{3}{c}{Number of tokens}                                        \\
                                    & \multicolumn{1}{c}{Avg} & \multicolumn{1}{c}{Max} & \multicolumn{1}{c}{Min} & \multicolumn{1}{c}{Avg} & \multicolumn{1}{c}{Max} & \multicolumn{1}{c}{Min} \\
\midrule
DALLE~\citep{DALLE}                               & 15.8                    & 36                      & 6                       & 18.35                   & 42                      & 9                       \\
Prompt-to-Prompt~\citep{hertz2022prompt}                    & 7.43                    & 12                      & 2                       & 8.56                    & 14                      & 3                       \\
DrawBench~\citep{Imagen}                           & 11.37                   & 51                      & 1                       & 14.03                   & 57                      & 3                       \\
HRS Bench~\citep{bakr2023hrs}                           & 12.85                   & 36                      & 1                       & 14.94                   & 42                      & 3                       \\
TIFA~\citep{hu2023tifa}                                & 11.46                   & 67                      & 3                       & 11.63                   & 82                      & 3                       \\ 
ParaImage-Big~\citep{wu2023paragraph} & 132.9 & - & - & - & - & - \\ 
ParaImage-Small~\citep{wu2023paragraph} & 70.6 & - & - & - & - & - \\ \hline
Ours                                & 46.56                   & 68                      & 19                      & 57.68                   & 90                      & 22       \\              
\bottomrule
\end{tabular}
\end{table*}

\begin{table}[t!]
\caption{\new{\textbf{Quantitative evaluation of long prompt generation.} We report the percentage of the time that the VQA output is aligned with the answer to a question regarding the generation. Our method consistently improves Stable Diffusion models.}}
\label{tab:quantitative_long}
\centering
\begin{tabular}{lll}
\toprule
           & \multicolumn{1}{c}{Stable Diffusion} & \multicolumn{1}{c}{Stable Diffusion XL} \\
\midrule
Plain-text generation & $72.86\pm{\small{1.13}}$               & $79.05\pm{\small{0.41}}$                  \\
Rich-text generation & $73.24\pm{\small{1.00}}$               & $81.08\pm{\small{0.66}}$                 \\
\bottomrule
\end{tabular}
\end{table}

\subsection{Quantitative Comparison}
\label{sec:quanitative}

We report the local CLIP scores computed by a ViT-B/32 model in Figure~\ref{tab:style_quantitative}. 
Our method achieves the best overall CLIP score compared to the two baselines. 
This demonstrates the advantage of our region-based diffusion method for localized stylization. 
To further understand the each model's capacity for generating multiple styles, we report the metric on each region. 
Prompt-to-Prompt and InstructPix2Pix-para achieve a decent score on the $1^{\text{st}}$ Region, i.e., the region first occurs in the sentence. 
However, they often fail to fulfill the style in the $2^{\text{nd}}$ Region. 
We conjecture that the Stable Diffusion model tends to generate a uniform style for the entire image, which can be attributed to single-style training images. 
Furthermore, InstructPix2Pix-seq performs the worst in $2^{\text{nd}}$ Region. 
This is because the first instruction contains no information about the second region, and the second region's content could be compromised when we apply the first instruction. 

We show quantitative results of precise color generation in Figure~\ref{fig:color_quantitative}. 
The distance of \emph{HTML color} is generally the lowest for baseline methods, as they provide the most interpretable textual information for text encoders. 
This aligns with our expectation that the diffusion model can handle simple color names, whereas they struggle to handle the RGB triplet. 
Our rich-text-to-image generation method consistently improves on the three categories and two metrics over the baselines.

\new{
In Table~\ref{tab:quantitative_long}, we follow TIFA~\citep{hu2023tifa} to evaluate the long prompt generation using either Stable Diffusion or Stable Diffusion XL as the base models. 
Specifically, we generate $5$ images for each text prompt using different random seeds and use the \emph{mplug-large} model~\citep{li2022mplug} to generate the answers to each benchmark question. 
We report how often the generated answers are consistent with the benchmark answers by averaging across different seeds. 
We also report the standard deviation. 
Our rich-text method consistently improves the generation quality with different base models when a lengthy prompt is given. 
Moreover, we find that the Stable Diffusion XL significantly outperforms the Stable Diffusion model.}

\begin{figure*}[t]
    \centering
    \includegraphics[width=\linewidth, trim=0 0 0 0, clip]{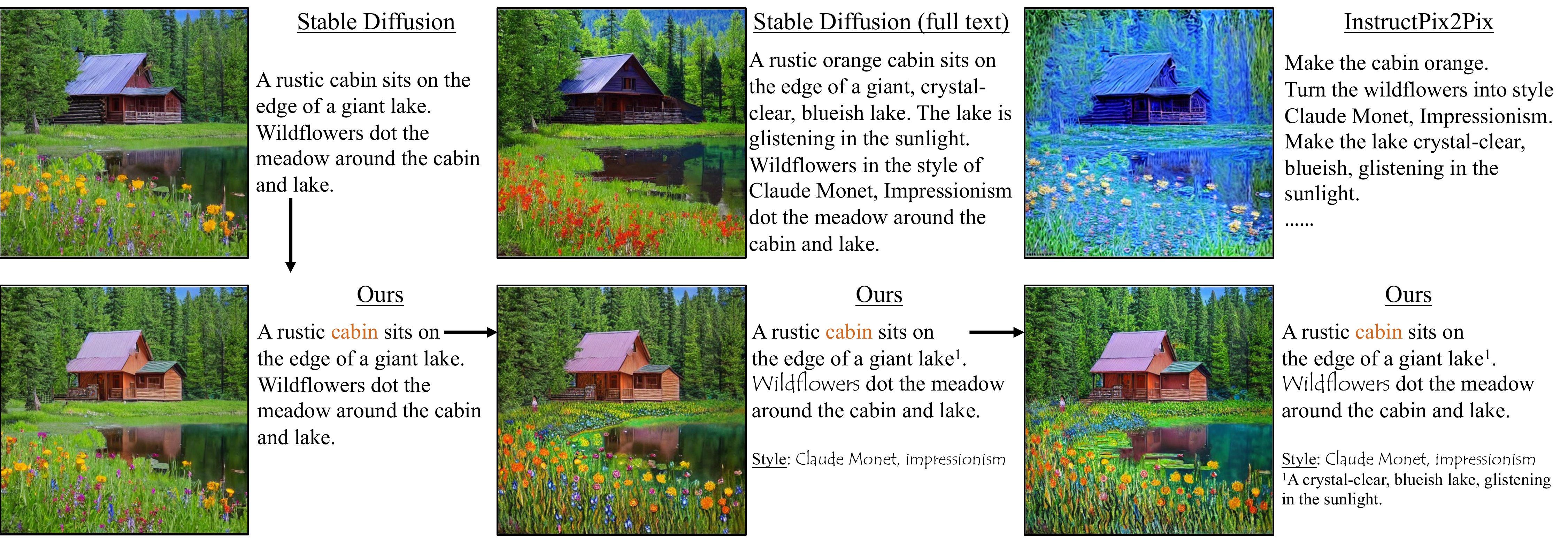}
    \caption{\textbf{Our workflow.} (top left) A user begins with an initial plain-text prompt and wishes to refine the scene by specifying the color, details, and styles. (top center) Naively inputting the whole description in plain text does not work. (top right) InstructPix2Pix~\citep{brooks2022instructpix2pix} fails to make accurate editing. (bottom) Our method supports precise refinement with region-constrained diffusion processes. Moreover, our framework can naturally be integrated into a rich text editor, enabling a tight, streamlined UI. }
    \label{fig:qualitative_workflow}
\end{figure*}

\subsection{Visual Comparison}
\label{sec:visual_comparison}

\topic{Precise color generation.} We show qualitative comparison on precise color generation in Figure~\ref{fig:qualitative_color}. 
InstructPix2Pix~\citep{brooks2022instructpix2pix} is prone to create global color effects rather than accurate local control. 
For example, in the flower results, 
both the vase and background change to the target colors. 
Prompt-to-Prompt~\citep{hertz2022prompt} provides more precise control over the target region. 
However, neither Prompt-to-Prompt nor InstructPix2Pix can generate precise colors. 
In contrast, our method can generate precise colors for all categories and prompts.

\topic{Local style generation.}  
Figure~\ref{fig:qualitative_style} visually compares local style generation.
When applying InstructPix2Pix-seq, the style in the first instruction dominates the entire image and undermines the second region. 
Figure 13 in the Appendix shows this cannot be fully resolved using different hyperparameters of classifier-free guidance. 
Similar to our observation in the quantitative evaluation, our baselines tend to generate the image in a globally uniform style instead of distinct local styles for each region. 
In contrast, our method synthesizes the correct styles for both regions. 
One may suggest independently applying baselines with two stylization processes and composing the results using token maps. 
However, Figure 12 (Appendix) shows that such methods generate artifacts on the region boundaries.

\topic{\new{Texture-guidaded generation.}} \new{The results in Figure~\ref{fig:qualitative_texture} demonstrate our method's ability to seamlessly incorporate texture information specified by a reference image into image generation results while preserving the other information of the textual descriptions, such as the overall shape and geometry. For instance, the generated cake aligns with the visual style of the salmon in the texture guidance while preserving the overall shape and structure of the original cake. The other objects, such as the cat, remain unchanged. Also, the chef's apron exhibits a clear artistic style inspired by the reference while retaining the size and shape of the original apron in plain-text generation. }

\topic{Complex scene generation.} 
Figure~\ref{fig:qualitative_footnote} shows comparisons on complex scene generation. 
Attend-and-Excite~\citep{chefer2023attend} uses the tokens missing in the full-text generation result as input to fix the missing objects, like the coffee table and carpet in the living room example. 
However, it still fails to generate all the details correctly, e.g., the books, the painting, and the blanket. 
Prompt-to-Prompt~\citep{hertz2022prompt} and InstructPix2Pix~\citep{brooks2022instructpix2pix} can edit the painting accordingly, but many objects, like the colorful pillows and stuff on the table, are still missing. 
In contrast, our method faithfully synthesizes all these details described in the target region.

\topic{Token importance control.} 
Figure~\ref{fig:qualitative_size} shows the qualitative comparison on token reweighting. 
When using a large weight for `mushroom,' Prompt-to-Prompt generates clear artifacts as it modifies the attention probabilities to be unbounded and creates out-of-distribution intermediate features. 
Heuristic methods fail when adding more mushrooms, while our method generates more mushrooms and preserves the quality. More results of different font sizes and target tokens are shown in Figures 23 - 25 in the Appendix.

\topic{\new{Customized concept generation.}} \new{We show reference-driven image generation in Figure~\ref{fig:qualitative_embeds}. 
Our method can generate the customized concepts specified by the embedded images without making undesired changes to the plain-text generation. For example, only the backpack design is modified for the example on the left, and the dog's breeds are changed (instead of the pose) on the right.}

\begin{figure}[!t]
    \centering
    \includegraphics[width=1.0\linewidth, trim=0 0 0 0, clip]{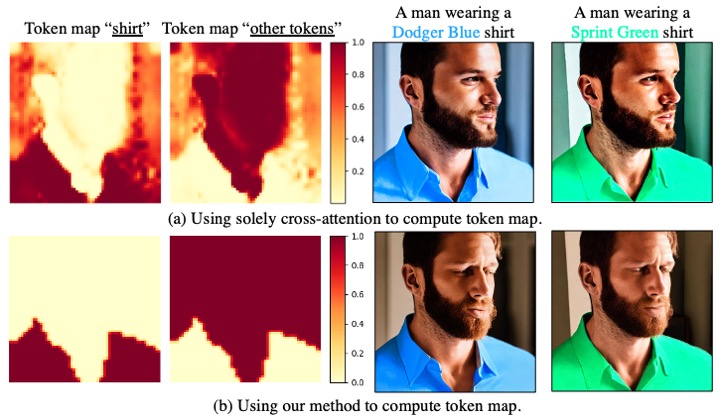}
    \caption{\textbf{Ablation of token maps.} Using solely cross-attention maps to create token maps leads to inaccurate segmentations, causing the background to be colored in an undesired way.}
    \label{fig:ablation4}
    \vspace{-10pt}
\end{figure}

\topic{\new{More visual generation results.}} \new{We show more rich-text generation results using Stable Diffsuion XL model~\citep{podell2024sdxl} with a focus on multiple rich-text attributes in Figure~\ref{fig:qualitative_composition}. We show that these rich-text formatting options provide great flexibility in customizing the text-to-image generation process using our region-based diffusion and guidance. Our region-based injection also helps preserve the fidelity with respect to plain-text results. For example, the font color supports accurate controlling the color of dressing and looking, font style allows freely rendering artistic styles in a certain region, embedded image insert personal reference image into the generation seamlessly, and so on. In addition, combining these options can further boost the controlibitly. For example, with the font color specifying the color of the hat, a footnote can be further used to describe the type of hat. Also, powered by the improved generation capacity of SDXL, we also show the generation of large and complex scenes, such as a snowy forest and a city square, with many visual elements indicated by the rich-text attributes. In conclusion, these results highlight that our method and design choices are agnostic to the base model and can be readily generalized to different pre-trained diffusion models.
}

\topic{Real image editing.} 
We also explore using rich text as an alternative to plain text in real image editing. Given a real image, we manually caption the image with text descriptions and use the off-the-shelf diffusion inversion methods to obtain the noise latent that reconstructs the image~\citep{HubermanSpiegelglas2023,wu2022unifying,ju2024pnp}. We find that our method is generally robust to different inversion methods. In the experiments, we use the edit-friendly DDPM~\citep{HubermanSpiegelglas2023} as our default inversion method. Unlike the image generation setting, the source image is already available here. 
We thus use the state-of-the-art grounded segmentation methods~\citep{kirillov2023segany,liu2023grounding,ren2024grounded} to produce the token maps. 
This is generally more robust than the cross-attention-based methods since the manually created text prompts for inversion may not accurately describe the image as the model expects.

\begin{figure}[!t]
    \centering
    \includegraphics[width=1.0\linewidth, trim=0 0 0 0, clip]{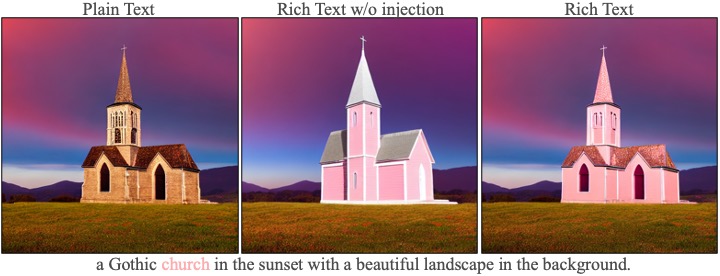}
    \caption{\textbf{Ablation of injection method.} We show images generated based on plain text and rich text with or without injection methods. Injecting features and noised samples help preserve the structure of the church and unformatted token regions.}
    \vspace{-0.5cm}
    \label{fig:ablation5}
\end{figure}

We adopt the same configuration we used in image generation, including the region-based guidance denoising and injection methods. We compare with existing editing methods, including inversion and editing with plain text~\citep{HubermanSpiegelglas2023}, InstructPix2Pix~\citep{brooks2022instructpix2pix}, and Plug-and-Play~\citep{tumanyan2022plug}.

As shown in Figure~\ref{fig:editing}, the rich text allows editing the regions' color, style, and content with accurate controllability. Specifically, all existing methods struggle with preserving the details in the non-edited regions. For example, in the Van Gogh portrait example, all the baselines change the styles of areas other than the hat. Instead, our injection method and the region information in the token masks greatly improve the fidelity in these regions. In addition, as we allow direct usage of the color RGB information, we edit the dress color (in the example of the second row) more accurately.

\begin{figure*}[ht!]
    \centering
    \vspace{-1cm}
    \includegraphics[width=\linewidth, trim=0 0 0 0, clip]{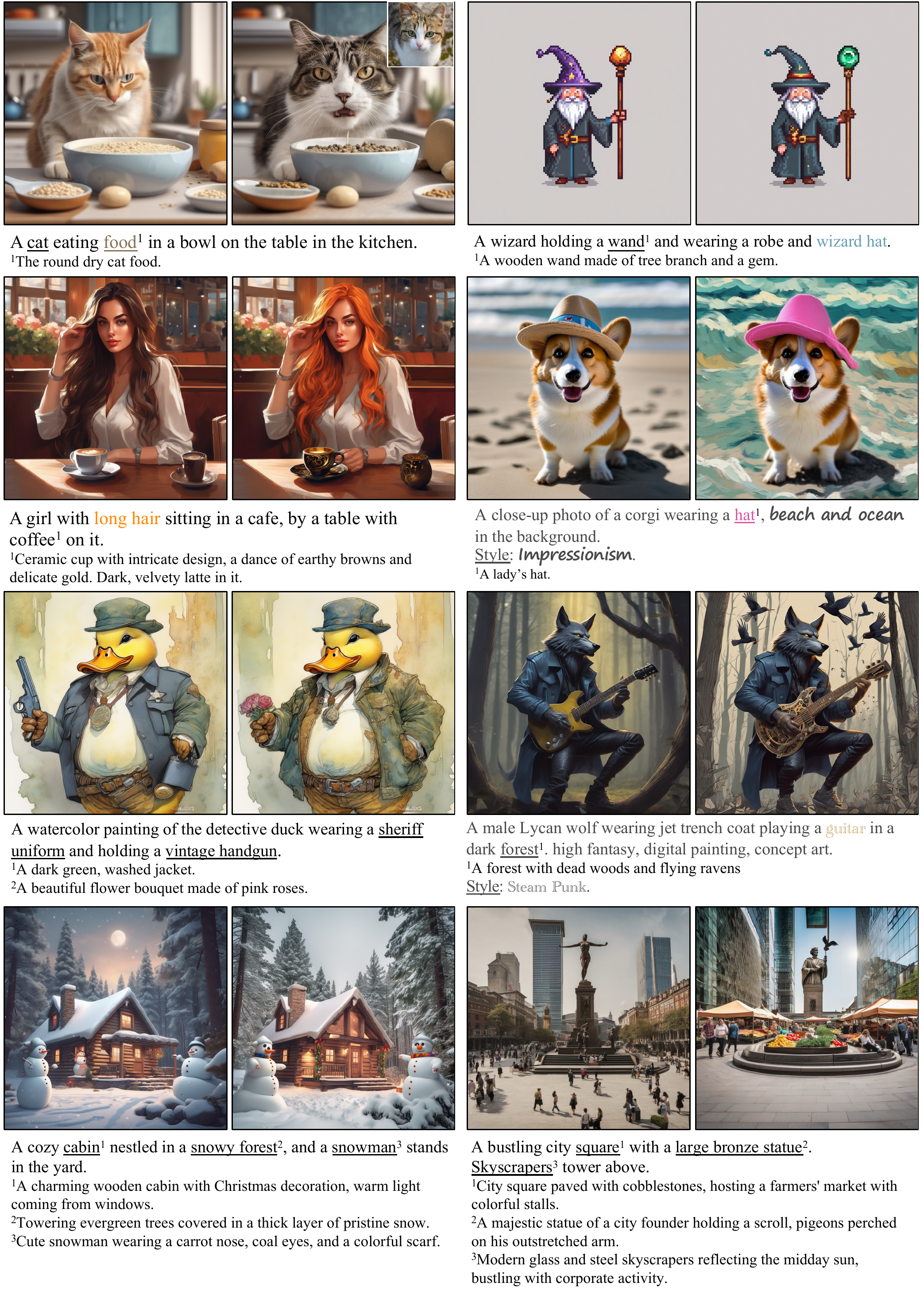}
    \caption{\textbf{More rich-text guided image generation results.} We show more visual generation results by SDXL using rich texts with \emph{multiple} text attributes, with a highlight on generating complex scenes.}
    \label{fig:qualitative_composition}
\end{figure*}
\clearpage

\begin{figure*}[t]
    \centering
    \includegraphics[width=\linewidth, trim=0 0 0 0, clip]{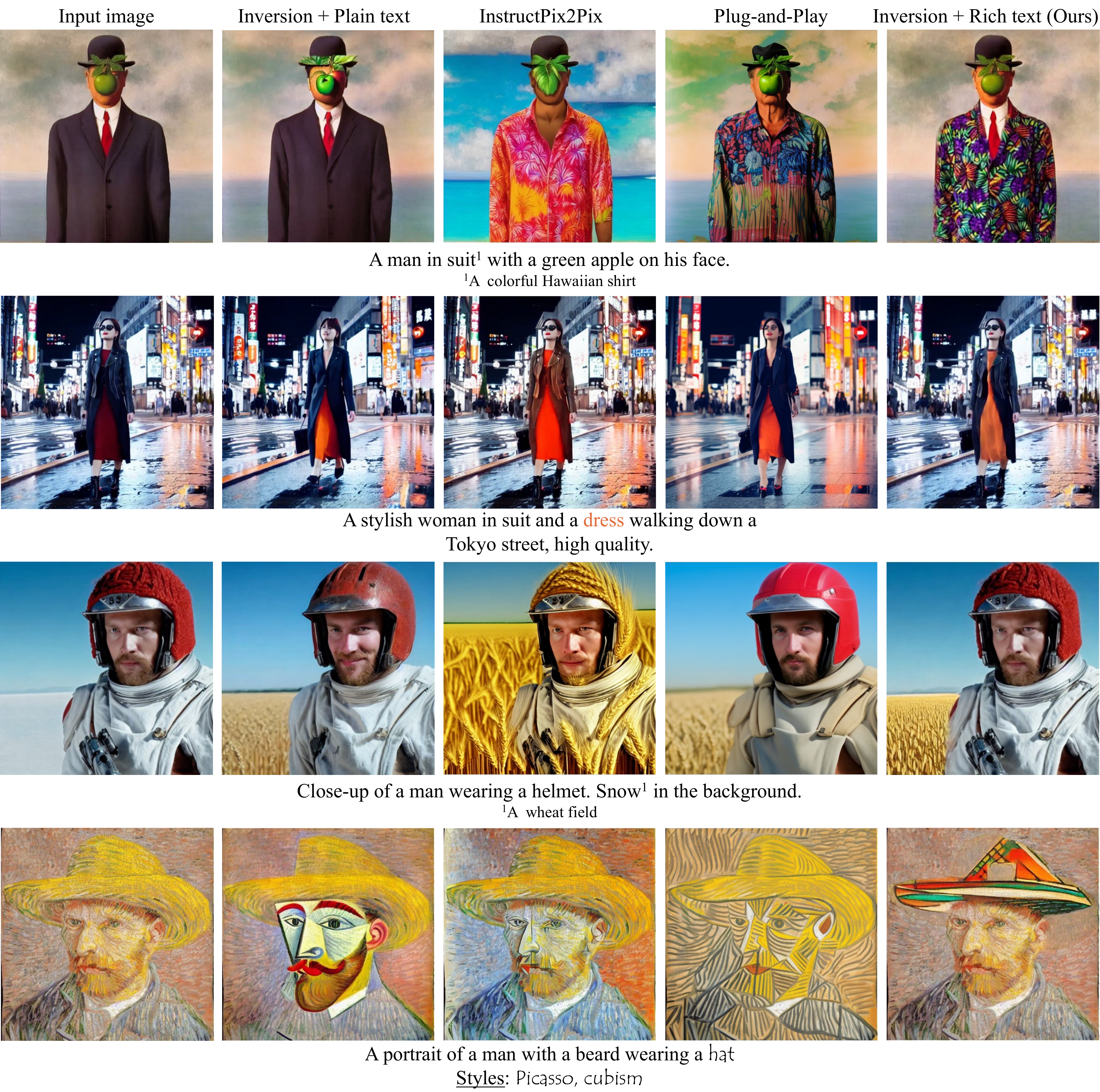}
    \caption{\new{\textbf{Image editing with rich text.} We show that with the off-the-shelf inversion techniques, our methods easily slot in as an alternative to plain-text editing. Using rich text with our region-based guidance and diffusion allows more precise control in editing real images than existing methods.}}
    \label{fig:editing}
    \vspace{-7pt}
\end{figure*}

\topic{Interactive editing.} 
In Figure~\ref{fig:qualitative_workflow}, we showcase a sample workflow to illustrate our method's interactive strength and editing capacity over InstructPix2Pix~\citep{brooks2022instructpix2pix}.

\subsection{Ablation Study}
\label{sec:ablation}


\topic{Generating token maps solely from cross-attention.} 
The other straightforward way to create token maps is to use cross-attention maps directly. To ablate this, we first take the average of cross-attention maps across heads, layers, and time steps and then take the maximum across tokens. Finally, we apply softmax across all the spans to normalize the token maps. However, as shown by the example in Figure~\ref{fig:ablation4}, since the prompt has no correspondence with the background, the token map of ``shirt'' also covers partial background regions. Note that simple thresholding is ineffective as some regions still have high values, e.g., the right shoulder. As a result, the target color \emph{bleeds} into the background. Our methods obtain more accurate token maps and, consequently, more precise colorization.

\topic{Ablation of the injection methods.} 
To demonstrate the effectiveness of our injection method, we compare image generation with and without it in Figure~\ref{fig:ablation5}. In the font color example, we show that applying the injection effectively preserves the shape and details of the target church and the structure of the sunset in the background. In the footnote example, we show that the injection keeps the look of the black door and the color of the floor.
\section{Discussion and Limitations}
\label{sec:conclusions}
In this paper, we have expanded the controllability of text-to-image models by incorporating rich-text attributes as the input. 
We have demonstrated the potential for generating and editing images with local styles, precise colors, texture guidance, different token importance, reference images, and complex descriptions.
Nevertheless, numerous formatting options remain unexplored, such as bold/italic, hyperlinks, spacing, and bullets/numbering. 
Also, there are multiple ways to use the same formatting options. 
For example, one can use font style to characterize the shape of the objects. 
We hope this paper encourages further exploration of integrating accessible user interfaces into text-based content creation tasks, even beyond images.

\bibliographystyle{sn-mathphys-num}
\bibliography{sn-bibliography}


\begin{thebibliography}{115}
\ifx \bisbn   \undefined \def \bisbn  #1{ISBN #1}\fi
\ifx \binits  \undefined \def \binits#1{#1}\fi
\ifx \bauthor  \undefined \def \bauthor#1{#1}\fi
\ifx \batitle  \undefined \def \batitle#1{#1}\fi
\ifx \bjtitle  \undefined \def \bjtitle#1{#1}\fi
\ifx \bvolume  \undefined \def \bvolume#1{\textbf{#1}}\fi
\ifx \byear  \undefined \def \byear#1{#1}\fi
\ifx \bissue  \undefined \def \bissue#1{#1}\fi
\ifx \bfpage  \undefined \def \bfpage#1{#1}\fi
\ifx \blpage  \undefined \def \blpage #1{#1}\fi
\ifx \burl  \undefined \def \burl#1{\textsf{#1}}\fi
\ifx \doiurl  \undefined \def \doiurl#1{\url{https://doi.org/#1}}\fi
\ifx \betal  \undefined \def \betal{\textit{et al.}}\fi
\ifx \binstitute  \undefined \def \binstitute#1{#1}\fi
\ifx \binstitutionaled  \undefined \def \binstitutionaled#1{#1}\fi
\ifx \bctitle  \undefined \def \bctitle#1{#1}\fi
\ifx \beditor  \undefined \def \beditor#1{#1}\fi
\ifx \bpublisher  \undefined \def \bpublisher#1{#1}\fi
\ifx \bbtitle  \undefined \def \bbtitle#1{#1}\fi
\ifx \bedition  \undefined \def \bedition#1{#1}\fi
\ifx \bseriesno  \undefined \def \bseriesno#1{#1}\fi
\ifx \blocation  \undefined \def \blocation#1{#1}\fi
\ifx \bsertitle  \undefined \def \bsertitle#1{#1}\fi
\ifx \bsnm \undefined \def \bsnm#1{#1}\fi
\ifx \bsuffix \undefined \def \bsuffix#1{#1}\fi
\ifx \bparticle \undefined \def \bparticle#1{#1}\fi
\ifx \barticle \undefined \def \barticle#1{#1}\fi
\bibcommenthead
\ifx \bconfdate \undefined \def \bconfdate #1{#1}\fi
\ifx \botherref \undefined \def \botherref #1{#1}\fi
\ifx \url \undefined \def \url#1{\textsf{#1}}\fi
\ifx \bchapter \undefined \def \bchapter#1{#1}\fi
\ifx \bbook \undefined \def \bbook#1{#1}\fi
\ifx \bcomment \undefined \def \bcomment#1{#1}\fi
\ifx \oauthor \undefined \def \oauthor#1{#1}\fi
\ifx \citeauthoryear \undefined \def \citeauthoryear#1{#1}\fi
\ifx \endbibitem  \undefined \def \endbibitem {}\fi
\ifx \bconflocation  \undefined \def \bconflocation#1{#1}\fi
\ifx \arxivurl  \undefined \def \arxivurl#1{\textsf{#1}}\fi
\csname PreBibitemsHook\endcsname

\bibitem[\protect\citeauthoryear{Ramesh et~al.}{2021}]{DALLE}
\begin{bchapter}
\bauthor{\bsnm{Ramesh}, \binits{A.}},
\bauthor{\bsnm{Pavlov}, \binits{M.}},
\bauthor{\bsnm{Goh}, \binits{G.}},
\bauthor{\bsnm{Gray}, \binits{S.}},
\bauthor{\bsnm{Voss}, \binits{C.}},
\bauthor{\bsnm{Radford}, \binits{A.}},
\bauthor{\bsnm{Chen}, \binits{M.}},
\bauthor{\bsnm{Sutskever}, \binits{I.}}:
\bctitle{Zero-shot text-to-image generation}.
In: \bbtitle{ICML},
pp. \bfpage{8821}--\blpage{8831}
(\byear{2021})
\end{bchapter}
\endbibitem

\bibitem[\protect\citeauthoryear{Saharia et~al.}{2022}]{Imagen}
\begin{bchapter}
\bauthor{\bsnm{Saharia}, \binits{C.}},
\bauthor{\bsnm{Chan}, \binits{W.}},
\bauthor{\bsnm{Saxena}, \binits{S.}},
\bauthor{\bsnm{Li}, \binits{L.}},
\bauthor{\bsnm{Whang}, \binits{J.}},
\bauthor{\bsnm{Denton}, \binits{E.}},
\bauthor{\bsnm{Ghasemipour}, \binits{S.K.S.}},
\bauthor{\bsnm{Ayan}, \binits{B.K.}},
\bauthor{\bsnm{Mahdavi}, \binits{S.S.}},
\bauthor{\bsnm{Lopes}, \binits{R.G.}},
\bauthor{\bsnm{Salimans}, \binits{T.}},
\bauthor{\bsnm{Ho}, \binits{J.}},
\bauthor{\bsnm{Fleet}, \binits{D.J.}},
\bauthor{\bsnm{Norouzi}, \binits{M.}}:
\bctitle{Photorealistic text-to-image diffusion models with deep language understanding}.
In: \bbtitle{NeurIPS}
(\byear{2022})
\end{bchapter}
\endbibitem

\bibitem[\protect\citeauthoryear{Rombach et~al.}{2022}]{rombach2022high}
\begin{bchapter}
\bauthor{\bsnm{Rombach}, \binits{R.}},
\bauthor{\bsnm{Blattmann}, \binits{A.}},
\bauthor{\bsnm{Lorenz}, \binits{D.}},
\bauthor{\bsnm{Esser}, \binits{P.}},
\bauthor{\bsnm{Ommer}, \binits{B.}}:
\bctitle{High-resolution image synthesis with latent diffusion models}.
In: \bbtitle{CVPR}
(\byear{2022})
\end{bchapter}
\endbibitem

\bibitem[\protect\citeauthoryear{Kang et~al.}{2023}]{kang2023gigagan}
\begin{bchapter}
\bauthor{\bsnm{Kang}, \binits{M.}},
\bauthor{\bsnm{Zhu}, \binits{J.-Y.}},
\bauthor{\bsnm{Zhang}, \binits{R.}},
\bauthor{\bsnm{Park}, \binits{J.}},
\bauthor{\bsnm{Shechtman}, \binits{E.}},
\bauthor{\bsnm{Paris}, \binits{S.}},
\bauthor{\bsnm{Park}, \binits{T.}}:
\bctitle{Scaling up gans for text-to-image synthesis}.
In: \bbtitle{CVPR}
(\byear{2023})
\end{bchapter}
\endbibitem

\bibitem[\protect\citeauthoryear{Balaji et~al.}{2022}]{balaji2022ediffi}
\begin{botherref}
\oauthor{\bsnm{Balaji}, \binits{Y.}},
\oauthor{\bsnm{Nah}, \binits{S.}},
\oauthor{\bsnm{Huang}, \binits{X.}},
\oauthor{\bsnm{Vahdat}, \binits{A.}},
\oauthor{\bsnm{Song}, \binits{J.}},
\oauthor{\bsnm{Kreis}, \binits{K.}},
\oauthor{\bsnm{Aittala}, \binits{M.}},
\oauthor{\bsnm{Aila}, \binits{T.}},
\oauthor{\bsnm{Laine}, \binits{S.}},
\oauthor{\bsnm{Catanzaro}, \binits{B.}}, et al.:
ediffi: Text-to-image diffusion models with an ensemble of expert denoisers.
arXiv preprint arXiv:2211.01324
(2022)
\end{botherref}
\endbibitem

\bibitem[\protect\citeauthoryear{Gafni et~al.}{2022}]{Make-A-Scene}
\begin{bchapter}
\bauthor{\bsnm{Gafni}, \binits{O.}},
\bauthor{\bsnm{Polyak}, \binits{A.}},
\bauthor{\bsnm{Ashual}, \binits{O.}},
\bauthor{\bsnm{Sheynin}, \binits{S.}},
\bauthor{\bsnm{Parikh}, \binits{D.}},
\bauthor{\bsnm{Taigman}, \binits{Y.}}:
\bctitle{Make-a-scene: Scene-based text-to-image generation with human priors}.
In: \bbtitle{ECCV}
(\byear{2022})
\end{bchapter}
\endbibitem

\bibitem[\protect\citeauthoryear{Zhang and Agrawala}{2023}]{zhang2023adding}
\begin{bchapter}
\bauthor{\bsnm{Zhang}, \binits{L.}},
\bauthor{\bsnm{Agrawala}, \binits{M.}}:
\bctitle{Adding conditional control to text-to-image diffusion models}.
In: \bbtitle{ICCV}
(\byear{2023})
\end{bchapter}
\endbibitem

\bibitem[\protect\citeauthoryear{Brooks et~al.}{2023}]{brooks2022instructpix2pix}
\begin{bchapter}
\bauthor{\bsnm{Brooks}, \binits{T.}},
\bauthor{\bsnm{Holynski}, \binits{A.}},
\bauthor{\bsnm{Efros}, \binits{A.A.}}:
\bctitle{Instructpix2pix: Learning to follow image editing instructions}.
In: \bbtitle{CVPR}
(\byear{2023})
\end{bchapter}
\endbibitem

\bibitem[\protect\citeauthoryear{Hertz et~al.}{2023}]{hertz2022prompt}
\begin{bchapter}
\bauthor{\bsnm{Hertz}, \binits{A.}},
\bauthor{\bsnm{Mokady}, \binits{R.}},
\bauthor{\bsnm{Tenenbaum}, \binits{J.}},
\bauthor{\bsnm{Aberman}, \binits{K.}},
\bauthor{\bsnm{Pritch}, \binits{Y.}},
\bauthor{\bsnm{Cohen-or}, \binits{D.}}:
\bctitle{Prompt-to-prompt image editing with cross-attention control}.
In: \bbtitle{ICLR}
(\byear{2023})
\end{bchapter}
\endbibitem

\bibitem[\protect\citeauthoryear{Colorado State~University}{2012}]{rtf}
\begin{botherref}
\oauthor{\bsnm{Colorado State~University}, \binits{T.A.P.}}:
tutorial: Rich Text Format (RTF) from Microsoft Word - The ACCESS Project
(2012).
\url{http://accessproject.colostate.edu/udl/modules/word/tut_rtf.cfm}
\end{botherref}
\endbibitem

\bibitem[\protect\citeauthoryear{Witten et~al.}{2009}]{witten2009build}
\begin{bbook}
\bauthor{\bsnm{Witten}, \binits{I.H.}},
\bauthor{\bsnm{Bainbridge}, \binits{D.}},
\bauthor{\bsnm{Nichols}, \binits{D.M.}}:
\bbtitle{How to Build a Digital Library},
(\byear{2009})
\end{bbook}
\endbibitem

\bibitem[\protect\citeauthoryear{Chefer et~al.}{2023}]{chefer2023attend}
\begin{barticle}
\bauthor{\bsnm{Chefer}, \binits{H.}},
\bauthor{\bsnm{Alaluf}, \binits{Y.}},
\bauthor{\bsnm{Vinker}, \binits{Y.}},
\bauthor{\bsnm{Wolf}, \binits{L.}},
\bauthor{\bsnm{Cohen-Or}, \binits{D.}}:
\batitle{Attend-and-excite: Attention-based semantic guidance for text-to-image diffusion models}.
\bjtitle{ACM Transactions on Graphics (TOG)}
\bvolume{42}(\bissue{4}),
\bfpage{1}--\blpage{10}
(\byear{2023})
\end{barticle}
\endbibitem

\bibitem[\protect\citeauthoryear{Ge et~al.}{2023}]{ge2023expressive}
\begin{bchapter}
\bauthor{\bsnm{Ge}, \binits{S.}},
\bauthor{\bsnm{Park}, \binits{T.}},
\bauthor{\bsnm{Zhu}, \binits{J.-Y.}},
\bauthor{\bsnm{Huang}, \binits{J.-B.}}:
\bctitle{Expressive text-to-image generation with rich text}.
In: \bbtitle{ICCV}
(\byear{2023})
\end{bchapter}
\endbibitem

\bibitem[\protect\citeauthoryear{Zhu et~al.}{2007}]{zhu2007text}
\begin{bchapter}
\bauthor{\bsnm{Zhu}, \binits{X.}},
\bauthor{\bsnm{Goldberg}, \binits{A.B.}},
\bauthor{\bsnm{Eldawy}, \binits{M.}},
\bauthor{\bsnm{Dyer}, \binits{C.R.}},
\bauthor{\bsnm{Strock}, \binits{B.}}:
\bctitle{A text-to-picture synthesis system for augmenting communication}.
In: \bbtitle{AAAI}
(\byear{2007})
\end{bchapter}
\endbibitem

\bibitem[\protect\citeauthoryear{Mansimov et~al.}{2016}]{mansimov2015generating}
\begin{bchapter}
\bauthor{\bsnm{Mansimov}, \binits{E.}},
\bauthor{\bsnm{Parisotto}, \binits{E.}},
\bauthor{\bsnm{Ba}, \binits{J.L.}},
\bauthor{\bsnm{Salakhutdinov}, \binits{R.}}:
\bctitle{Generating images from captions with attention}.
In: \bbtitle{ICLR}
(\byear{2016})
\end{bchapter}
\endbibitem

\bibitem[\protect\citeauthoryear{Schuhmann et~al.}{2022}]{schuhmann2022laion}
\begin{bchapter}
\bauthor{\bsnm{Schuhmann}, \binits{C.}},
\bauthor{\bsnm{Beaumont}, \binits{R.}},
\bauthor{\bsnm{Vencu}, \binits{R.}},
\bauthor{\bsnm{Gordon}, \binits{C.}},
\bauthor{\bsnm{Wightman}, \binits{R.}},
\bauthor{\bsnm{Cherti}, \binits{M.}},
\bauthor{\bsnm{Coombes}, \binits{T.}},
\bauthor{\bsnm{Katta}, \binits{A.}},
\bauthor{\bsnm{Mullis}, \binits{C.}},
\bauthor{\bsnm{Wortsman}, \binits{M.}}, \betal:
\bctitle{Laion-5b: An open large-scale dataset for training next generation image-text models}.
In: \bbtitle{NeurIPS}
(\byear{2022})
\end{bchapter}
\endbibitem

\bibitem[\protect\citeauthoryear{Byeon et~al.}{2022}]{kakaobrain2022coyo-700m}
\begin{botherref}
\oauthor{\bsnm{Byeon}, \binits{M.}},
\oauthor{\bsnm{Park}, \binits{B.}},
\oauthor{\bsnm{Kim}, \binits{H.}},
\oauthor{\bsnm{Lee}, \binits{S.}},
\oauthor{\bsnm{Baek}, \binits{W.}},
\oauthor{\bsnm{Kim}, \binits{S.}}:
COYO-700M: Image-Text Pair Dataset.
\url{https://github.com/kakaobrain/coyo-dataset}
(2022)
\end{botherref}
\endbibitem

\bibitem[\protect\citeauthoryear{Ho et~al.}{2020}]{ho2020denoising}
\begin{botherref}
\oauthor{\bsnm{Ho}, \binits{J.}},
\oauthor{\bsnm{Jain}, \binits{A.}},
\oauthor{\bsnm{Abbeel}, \binits{P.}}:
Denoising diffusion probabilistic models.
NeurIPS
(2020)
\end{botherref}
\endbibitem

\bibitem[\protect\citeauthoryear{Song et~al.}{2021}]{song2021denoising}
\begin{bchapter}
\bauthor{\bsnm{Song}, \binits{J.}},
\bauthor{\bsnm{Meng}, \binits{C.}},
\bauthor{\bsnm{Ermon}, \binits{S.}}:
\bctitle{Denoising diffusion implicit models}.
In: \bbtitle{ICLR}
(\byear{2021})
\end{bchapter}
\endbibitem

\bibitem[\protect\citeauthoryear{Ho et~al.}{2022}]{ho2022cascaded}
\begin{barticle}
\bauthor{\bsnm{Ho}, \binits{J.}},
\bauthor{\bsnm{Saharia}, \binits{C.}},
\bauthor{\bsnm{Chan}, \binits{W.}},
\bauthor{\bsnm{Fleet}, \binits{D.J.}},
\bauthor{\bsnm{Norouzi}, \binits{M.}},
\bauthor{\bsnm{Salimans}, \binits{T.}}:
\batitle{Cascaded diffusion models for high fidelity image generation}.
\bjtitle{Journal of Machine Learning Research}
\bvolume{23}(\bissue{47}),
\bfpage{1}--\blpage{33}
(\byear{2022})
\end{barticle}
\endbibitem

\bibitem[\protect\citeauthoryear{Ho and Salimans}{2021}]{Classifier}
\begin{bchapter}
\bauthor{\bsnm{Ho}, \binits{J.}},
\bauthor{\bsnm{Salimans}, \binits{T.}}:
\bctitle{Classifier-free diffusion guidance}.
In: \bbtitle{NeurIPS 2021 Workshop on Deep Generative Models and Downstream Applications}
(\byear{2021})
\end{bchapter}
\endbibitem

\bibitem[\protect\citeauthoryear{Ramesh et~al.}{2022}]{DALLE2}
\begin{botherref}
\oauthor{\bsnm{Ramesh}, \binits{A.}},
\oauthor{\bsnm{Dhariwal}, \binits{P.}},
\oauthor{\bsnm{Nichol}, \binits{A.}},
\oauthor{\bsnm{Chu}, \binits{C.}},
\oauthor{\bsnm{Chen}, \binits{M.}}:
Hierarchical text-conditional image generation with clip latents.
arXiv preprint arXiv:2204.06125
(2022)
\end{botherref}
\endbibitem

\bibitem[\protect\citeauthoryear{Nichol et~al.}{2022}]{GLIDE}
\begin{bchapter}
\bauthor{\bsnm{Nichol}, \binits{A.Q.}},
\bauthor{\bsnm{Dhariwal}, \binits{P.}},
\bauthor{\bsnm{Ramesh}, \binits{A.}},
\bauthor{\bsnm{Shyam}, \binits{P.}},
\bauthor{\bsnm{Mishkin}, \binits{P.}},
\bauthor{\bsnm{Mcgrew}, \binits{B.}},
\bauthor{\bsnm{Sutskever}, \binits{I.}},
\bauthor{\bsnm{Chen}, \binits{M.}}:
\bctitle{Glide: Towards photorealistic image generation and editing with text-guided diffusion models}.
In: \bbtitle{ICML}
(\byear{2022})
\end{bchapter}
\endbibitem

\bibitem[\protect\citeauthoryear{Yu et~al.}{2022}]{Parti}
\begin{botherref}
\oauthor{\bsnm{Yu}, \binits{J.}},
\oauthor{\bsnm{Xu}, \binits{Y.}},
\oauthor{\bsnm{Koh}, \binits{J.Y.}},
\oauthor{\bsnm{Luong}, \binits{T.}},
\oauthor{\bsnm{Baid}, \binits{G.}},
\oauthor{\bsnm{Wang}, \binits{Z.}},
\oauthor{\bsnm{Vasudevan}, \binits{V.}},
\oauthor{\bsnm{Ku}, \binits{A.}},
\oauthor{\bsnm{Yang}, \binits{Y.}},
\oauthor{\bsnm{Ayan}, \binits{B.K.}}, et al.:
Scaling autoregressive models for content-rich text-to-image generation.
Transactions on Machine Learning Research
(2022)
\end{botherref}
\endbibitem

\bibitem[\protect\citeauthoryear{Chang et~al.}{2023}]{chang2023muse}
\begin{botherref}
\oauthor{\bsnm{Chang}, \binits{H.}},
\oauthor{\bsnm{Zhang}, \binits{H.}},
\oauthor{\bsnm{Barber}, \binits{J.}},
\oauthor{\bsnm{Maschinot}, \binits{A.}},
\oauthor{\bsnm{Lezama}, \binits{J.}},
\oauthor{\bsnm{Jiang}, \binits{L.}},
\oauthor{\bsnm{Yang}, \binits{M.-H.}},
\oauthor{\bsnm{Murphy}, \binits{K.}},
\oauthor{\bsnm{Freeman}, \binits{W.T.}},
\oauthor{\bsnm{Rubinstein}, \binits{M.}}, et al.:
Muse: Text-to-image generation via masked generative transformers.
arXiv preprint arXiv:2301.00704
(2023)
\end{botherref}
\endbibitem

\bibitem[\protect\citeauthoryear{Ding et~al.}{2022}]{ding2022cogview2}
\begin{botherref}
\oauthor{\bsnm{Ding}, \binits{M.}},
\oauthor{\bsnm{Zheng}, \binits{W.}},
\oauthor{\bsnm{Hong}, \binits{W.}},
\oauthor{\bsnm{Tang}, \binits{J.}}:
Cogview2: Faster and better text-to-image generation via hierarchical transformers.
arXiv preprint arXiv:2204.14217
(2022)
\end{botherref}
\endbibitem

\bibitem[\protect\citeauthoryear{Sauer et~al.}{2023}]{sauer2023stylegan}
\begin{botherref}
\oauthor{\bsnm{Sauer}, \binits{A.}},
\oauthor{\bsnm{Karras}, \binits{T.}},
\oauthor{\bsnm{Laine}, \binits{S.}},
\oauthor{\bsnm{Geiger}, \binits{A.}},
\oauthor{\bsnm{Aila}, \binits{T.}}:
Stylegan-t: Unlocking the power of gans for fast large-scale text-to-image synthesis.
arXiv preprint arXiv:2301.09515
(2023)
\end{botherref}
\endbibitem

\bibitem[\protect\citeauthoryear{Ruiz et~al.}{2023}]{ruiz2022dreambooth}
\begin{bchapter}
\bauthor{\bsnm{Ruiz}, \binits{N.}},
\bauthor{\bsnm{Li}, \binits{Y.}},
\bauthor{\bsnm{Jampani}, \binits{V.}},
\bauthor{\bsnm{Pritch}, \binits{Y.}},
\bauthor{\bsnm{Rubinstein}, \binits{M.}},
\bauthor{\bsnm{Aberman}, \binits{K.}}:
\bctitle{Dreambooth: Fine tuning text-to-image diffusion models for subject-driven generation}.
In: \bbtitle{CVPR}
(\byear{2023})
\end{bchapter}
\endbibitem

\bibitem[\protect\citeauthoryear{Kumari et~al.}{2023}]{kumari2022multi}
\begin{bchapter}
\bauthor{\bsnm{Kumari}, \binits{N.}},
\bauthor{\bsnm{Zhang}, \binits{B.}},
\bauthor{\bsnm{Zhang}, \binits{R.}},
\bauthor{\bsnm{Shechtman}, \binits{E.}},
\bauthor{\bsnm{Zhu}, \binits{J.-Y.}}:
\bctitle{Multi-concept customization of text-to-image diffusion}.
In: \bbtitle{CVPR}
(\byear{2023})
\end{bchapter}
\endbibitem

\bibitem[\protect\citeauthoryear{Avrahami et~al.}{2023}]{avrahami2022spatext}
\begin{bchapter}
\bauthor{\bsnm{Avrahami}, \binits{O.}},
\bauthor{\bsnm{Hayes}, \binits{T.}},
\bauthor{\bsnm{Gafni}, \binits{O.}},
\bauthor{\bsnm{Gupta}, \binits{S.}},
\bauthor{\bsnm{Taigman}, \binits{Y.}},
\bauthor{\bsnm{Parikh}, \binits{D.}},
\bauthor{\bsnm{Lischinski}, \binits{D.}},
\bauthor{\bsnm{Fried}, \binits{O.}},
\bauthor{\bsnm{Yin}, \binits{X.}}:
\bctitle{Spatext: Spatio-textual representation for controllable image generation}.
In: \bbtitle{CVPR}
(\byear{2023})
\end{bchapter}
\endbibitem

\bibitem[\protect\citeauthoryear{Wu et~al.}{2023}]{wu2022tune}
\begin{botherref}
\oauthor{\bsnm{Wu}, \binits{J.Z.}},
\oauthor{\bsnm{Ge}, \binits{Y.}},
\oauthor{\bsnm{Wang}, \binits{X.}},
\oauthor{\bsnm{Lei}, \binits{S.W.}},
\oauthor{\bsnm{Gu}, \binits{Y.}},
\oauthor{\bsnm{Shi}, \binits{Y.}},
\oauthor{\bsnm{Hsu}, \binits{W.}},
\oauthor{\bsnm{Shan}, \binits{Y.}},
\oauthor{\bsnm{Qie}, \binits{X.}},
\oauthor{\bsnm{Shou}, \binits{M.Z.}}:
Tune-a-video: One-shot tuning of image diffusion models for text-to-video generation,
7623--7633
(2023)
\end{botherref}
\endbibitem

\bibitem[\protect\citeauthoryear{Kawar et~al.}{2023}]{kawar2023imagic}
\begin{bchapter}
\bauthor{\bsnm{Kawar}, \binits{B.}},
\bauthor{\bsnm{Zada}, \binits{S.}},
\bauthor{\bsnm{Lang}, \binits{O.}},
\bauthor{\bsnm{Tov}, \binits{O.}},
\bauthor{\bsnm{Chang}, \binits{H.}},
\bauthor{\bsnm{Dekel}, \binits{T.}},
\bauthor{\bsnm{Mosseri}, \binits{I.}},
\bauthor{\bsnm{Irani}, \binits{M.}}:
\bctitle{Imagic: Text-based real image editing with diffusion models}.
In: \bbtitle{CVPR}
(\byear{2023})
\end{bchapter}
\endbibitem

\bibitem[\protect\citeauthoryear{Ma et~al.}{2023}]{ma2023follow}
\begin{botherref}
\oauthor{\bsnm{Ma}, \binits{Y.}},
\oauthor{\bsnm{He}, \binits{Y.}},
\oauthor{\bsnm{Cun}, \binits{X.}},
\oauthor{\bsnm{Wang}, \binits{X.}},
\oauthor{\bsnm{Shan}, \binits{Y.}},
\oauthor{\bsnm{Li}, \binits{X.}},
\oauthor{\bsnm{Chen}, \binits{Q.}}:
Follow Your Pose: Pose-Guided Text-to-Video Generation using Pose-Free Videos
(2023)
\end{botherref}
\endbibitem

\bibitem[\protect\citeauthoryear{Li et~al.}{2023}]{li2023gligen}
\begin{bchapter}
\bauthor{\bsnm{Li}, \binits{Y.}},
\bauthor{\bsnm{Liu}, \binits{H.}},
\bauthor{\bsnm{Wu}, \binits{Q.}},
\bauthor{\bsnm{Mu}, \binits{F.}},
\bauthor{\bsnm{Yang}, \binits{J.}},
\bauthor{\bsnm{Gao}, \binits{J.}},
\bauthor{\bsnm{Li}, \binits{C.}},
\bauthor{\bsnm{Lee}, \binits{Y.J.}}:
\bctitle{Gligen: Open-set grounded text-to-image generation}.
In: \bbtitle{CVPR}
(\byear{2023})
\end{bchapter}
\endbibitem

\bibitem[\protect\citeauthoryear{Meng et~al.}{2022}]{meng2021sdedit}
\begin{bchapter}
\bauthor{\bsnm{Meng}, \binits{C.}},
\bauthor{\bsnm{Song}, \binits{Y.}},
\bauthor{\bsnm{Song}, \binits{J.}},
\bauthor{\bsnm{Wu}, \binits{J.}},
\bauthor{\bsnm{Zhu}, \binits{J.-Y.}},
\bauthor{\bsnm{Ermon}, \binits{S.}}:
\bctitle{Sdedit: Image synthesis and editing with stochastic differential equations}.
In: \bbtitle{ICLR}
(\byear{2022})
\end{bchapter}
\endbibitem

\bibitem[\protect\citeauthoryear{Choi et~al.}{2021}]{choi2021ilvr}
\begin{bchapter}
\bauthor{\bsnm{Choi}, \binits{J.}},
\bauthor{\bsnm{Kim}, \binits{S.}},
\bauthor{\bsnm{Jeong}, \binits{Y.}},
\bauthor{\bsnm{Gwon}, \binits{Y.}},
\bauthor{\bsnm{Yoon}, \binits{S.}}:
\bctitle{Ilvr: Conditioning method for denoising diffusion probabilistic models}.
In: \bbtitle{ICCV}
(\byear{2021})
\end{bchapter}
\endbibitem

\bibitem[\protect\citeauthoryear{Parmar et~al.}{2023}]{parmar2023zero}
\begin{bchapter}
\bauthor{\bsnm{Parmar}, \binits{G.}},
\bauthor{\bsnm{Singh}, \binits{K.K.}},
\bauthor{\bsnm{Zhang}, \binits{R.}},
\bauthor{\bsnm{Li}, \binits{Y.}},
\bauthor{\bsnm{Lu}, \binits{J.}},
\bauthor{\bsnm{Zhu}, \binits{J.-Y.}}:
\bctitle{Zero-shot image-to-image translation}.
In: \bbtitle{SIGGRAPH}
(\byear{2023})
\end{bchapter}
\endbibitem

\bibitem[\protect\citeauthoryear{Bansal et~al.}{2023}]{bansal2023universal}
\begin{botherref}
\oauthor{\bsnm{Bansal}, \binits{A.}},
\oauthor{\bsnm{Chu}, \binits{H.-M.}},
\oauthor{\bsnm{Schwarzschild}, \binits{A.}},
\oauthor{\bsnm{Sengupta}, \binits{S.}},
\oauthor{\bsnm{Goldblum}, \binits{M.}},
\oauthor{\bsnm{Geiping}, \binits{J.}},
\oauthor{\bsnm{Goldstein}, \binits{T.}}:
Universal guidance for diffusion models.
arXiv preprint arXiv:2302.07121
(2023)
\end{botherref}
\endbibitem

\bibitem[\protect\citeauthoryear{Avrahami et~al.}{2022}]{avrahami2022blended}
\begin{botherref}
\oauthor{\bsnm{Avrahami}, \binits{O.}},
\oauthor{\bsnm{Fried}, \binits{O.}},
\oauthor{\bsnm{Lischinski}, \binits{D.}}:
Blended latent diffusion.
arXiv preprint arXiv:2206.02779
(2022)
\end{botherref}
\endbibitem

\bibitem[\protect\citeauthoryear{Jim{\'e}nez}{2023}]{jimenez2023mixture}
\begin{botherref}
\oauthor{\bsnm{Jim{\'e}nez}, \binits{{\'A}.B.}}:
Mixture of diffusers for scene composition and high resolution image generation.
arXiv preprint arXiv:2302.02412
(2023)
\end{botherref}
\endbibitem

\bibitem[\protect\citeauthoryear{Bar-Tal et~al.}{2023}]{bar2023multidiffusion}
\begin{bchapter}
\bauthor{\bsnm{Bar-Tal}, \binits{O.}},
\bauthor{\bsnm{Yariv}, \binits{L.}},
\bauthor{\bsnm{Lipman}, \binits{Y.}},
\bauthor{\bsnm{Dekel}, \binits{T.}}:
\bctitle{Multidiffusion: Fusing diffusion paths for controlled image generation}.
In: \bbtitle{ICML}
(\byear{2023})
\end{bchapter}
\endbibitem

\bibitem[\protect\citeauthoryear{Sarukkai et~al.}{2023}]{Sarukkai2023CollageD}
\begin{botherref}
\oauthor{\bsnm{Sarukkai}, \binits{V.}},
\oauthor{\bsnm{Li}, \binits{L.}},
\oauthor{\bsnm{Ma}, \binits{A.}},
\oauthor{\bsnm{R'e}, \binits{C.}},
\oauthor{\bsnm{Fatahalian}, \binits{K.}}:
Collage diffusion.
ArXiv
\textbf{abs/2303.00262}
(2023)
\end{botherref}
\endbibitem

\bibitem[\protect\citeauthoryear{Zhang et~al.}{2023}]{zhang2023diffcollage}
\begin{bchapter}
\bauthor{\bsnm{Zhang}, \binits{Q.}},
\bauthor{\bsnm{Song}, \binits{J.}},
\bauthor{\bsnm{Huang}, \binits{X.}},
\bauthor{\bsnm{Chen}, \binits{Y.}},
\bauthor{\bsnm{Liu}, \binits{M.-Y.}}:
\bctitle{Diffcollage: Parallel generation of large content with diffusion models}.
(\byear{2023})
\end{bchapter}
\endbibitem

\bibitem[\protect\citeauthoryear{Cao et~al.}{2023}]{cao2023masactrl}
\begin{bchapter}
\bauthor{\bsnm{Cao}, \binits{M.}},
\bauthor{\bsnm{Wang}, \binits{X.}},
\bauthor{\bsnm{Qi}, \binits{Z.}},
\bauthor{\bsnm{Shan}, \binits{Y.}},
\bauthor{\bsnm{Qie}, \binits{X.}},
\bauthor{\bsnm{Zheng}, \binits{Y.}}:
\bctitle{Masactrl: Tuning-free mutual self-attention control for consistent image synthesis and editing}.
In: \bbtitle{ICCV}
(\byear{2023})
\end{bchapter}
\endbibitem

\bibitem[\protect\citeauthoryear{Phung et~al.}{2024}]{phung2023grounded}
\begin{bchapter}
\bauthor{\bsnm{Phung}, \binits{Q.}},
\bauthor{\bsnm{Ge}, \binits{S.}},
\bauthor{\bsnm{Huang}, \binits{J.-B.}}:
\bctitle{Grounded text-to-image synthesis with attention refocusing}.
In: \bbtitle{CVPR}
(\byear{2024})
\end{bchapter}
\endbibitem

\bibitem[\protect\citeauthoryear{Xiao et~al.}{2023}]{xiao2023fastcomposer}
\begin{botherref}
\oauthor{\bsnm{Xiao}, \binits{G.}},
\oauthor{\bsnm{Yin}, \binits{T.}},
\oauthor{\bsnm{Freeman}, \binits{W.T.}},
\oauthor{\bsnm{Durand}, \binits{F.}},
\oauthor{\bsnm{Han}, \binits{S.}}:
Fastcomposer: Tuning-free multi-subject image generation with localized attention.
arXiv preprint arXiv:2305.10431
(2023)
\end{botherref}
\endbibitem

\bibitem[\protect\citeauthoryear{Feng et~al.}{2023}]{feng2023trainingfree}
\begin{bchapter}
\bauthor{\bsnm{Feng}, \binits{W.}},
\bauthor{\bsnm{He}, \binits{X.}},
\bauthor{\bsnm{Fu}, \binits{T.-J.}},
\bauthor{\bsnm{Jampani}, \binits{V.}},
\bauthor{\bsnm{Akula}, \binits{A.R.}},
\bauthor{\bsnm{Narayana}, \binits{P.}},
\bauthor{\bsnm{Basu}, \binits{S.}},
\bauthor{\bsnm{Wang}, \binits{X.E.}},
\bauthor{\bsnm{Wang}, \binits{W.Y.}}:
\bctitle{Training-free structured diffusion guidance for compositional text-to-image synthesis}.
In: \bbtitle{ICLR}
(\byear{2023})
\end{bchapter}
\endbibitem

\bibitem[\protect\citeauthoryear{Ho et~al.}{2022}]{ho2022video}
\begin{bchapter}
\bauthor{\bsnm{Ho}, \binits{J.}},
\bauthor{\bsnm{Salimans}, \binits{T.}},
\bauthor{\bsnm{Gritsenko}, \binits{A.}},
\bauthor{\bsnm{Chan}, \binits{W.}},
\bauthor{\bsnm{Norouzi}, \binits{M.}},
\bauthor{\bsnm{Fleet}, \binits{D.J.}}:
\bctitle{Video diffusion models}.
In: \bbtitle{NeurIPS}
(\byear{2022})
\end{bchapter}
\endbibitem

\bibitem[\protect\citeauthoryear{Park et~al.}{2019}]{park2019semantic}
\begin{bchapter}
\bauthor{\bsnm{Park}, \binits{T.}},
\bauthor{\bsnm{Liu}, \binits{M.-Y.}},
\bauthor{\bsnm{Wang}, \binits{T.-C.}},
\bauthor{\bsnm{Zhu}, \binits{J.-Y.}}:
\bctitle{Semantic image synthesis with spatially-adaptive normalization}.
In: \bbtitle{CVPR}
(\byear{2019})
\end{bchapter}
\endbibitem

\bibitem[\protect\citeauthoryear{Johnson et~al.}{2018}]{johnson2018image}
\begin{bchapter}
\bauthor{\bsnm{Johnson}, \binits{J.}},
\bauthor{\bsnm{Gupta}, \binits{A.}},
\bauthor{\bsnm{Fei-Fei}, \binits{L.}}:
\bctitle{Image generation from scene graphs}.
In: \bbtitle{CVPR}
(\byear{2018})
\end{bchapter}
\endbibitem

\bibitem[\protect\citeauthoryear{Isola et~al.}{2017}]{isola2017image}
\begin{bchapter}
\bauthor{\bsnm{Isola}, \binits{P.}},
\bauthor{\bsnm{Zhu}, \binits{J.-Y.}},
\bauthor{\bsnm{Zhou}, \binits{T.}},
\bauthor{\bsnm{Efros}, \binits{A.A.}}:
\bctitle{Image-to-image translation with conditional adversarial networks}.
In: \bbtitle{CVPR}
(\byear{2017})
\end{bchapter}
\endbibitem

\bibitem[\protect\citeauthoryear{Couairon et~al.}{2023}]{couairon2023zero}
\begin{bchapter}
\bauthor{\bsnm{Couairon}, \binits{G.}},
\bauthor{\bsnm{Careil}, \binits{M.}},
\bauthor{\bsnm{Cord}, \binits{M.}},
\bauthor{\bsnm{Lathuiliere}, \binits{S.}},
\bauthor{\bsnm{Verbeek}, \binits{J.}}:
\bctitle{Zero-shot spatial layout conditioning for text-to-image diffusion models}.
In: \bbtitle{ICCV}
(\byear{2023})
\end{bchapter}
\endbibitem

\bibitem[\protect\citeauthoryear{Park et~al.}{2023}]{park2023learning}
\begin{bchapter}
\bauthor{\bsnm{Park}, \binits{M.}},
\bauthor{\bsnm{Yun}, \binits{J.}},
\bauthor{\bsnm{Choi}, \binits{S.}},
\bauthor{\bsnm{Choo}, \binits{J.}}:
\bctitle{Learning to generate semantic layouts for higher text-image correspondence in text-to-image synthesis}.
In: \bbtitle{ICCV}
(\byear{2023})
\end{bchapter}
\endbibitem

\bibitem[\protect\citeauthoryear{Feng et~al.}{2024}]{feng2024layoutgpt}
\begin{bchapter}
\bauthor{\bsnm{Feng}, \binits{W.}},
\bauthor{\bsnm{Zhu}, \binits{W.}},
\bauthor{\bsnm{Fu}, \binits{T.-j.}},
\bauthor{\bsnm{Jampani}, \binits{V.}},
\bauthor{\bsnm{Akula}, \binits{A.}},
\bauthor{\bsnm{He}, \binits{X.}},
\bauthor{\bsnm{Basu}, \binits{S.}},
\bauthor{\bsnm{Wang}, \binits{X.E.}},
\bauthor{\bsnm{Wang}, \binits{W.Y.}}:
\bctitle{Layoutgpt: Compositional visual planning and generation with large language models}.
(\byear{2024})
\end{bchapter}
\endbibitem

\bibitem[\protect\citeauthoryear{Qu et~al.}{2023}]{qu2023layoutllm}
\begin{bchapter}
\bauthor{\bsnm{Qu}, \binits{L.}},
\bauthor{\bsnm{Wu}, \binits{S.}},
\bauthor{\bsnm{Fei}, \binits{H.}},
\bauthor{\bsnm{Nie}, \binits{L.}},
\bauthor{\bsnm{Chua}, \binits{T.-S.}}:
\bctitle{Layoutllm-t2i: Eliciting layout guidance from llm for text-to-image generation}.
In: \bbtitle{ACM MM}
(\byear{2023})
\end{bchapter}
\endbibitem

\bibitem[\protect\citeauthoryear{Liu et~al.}{2023}]{liu2023zero}
\begin{bchapter}
\bauthor{\bsnm{Liu}, \binits{R.}},
\bauthor{\bsnm{Wu}, \binits{R.}},
\bauthor{\bsnm{Van~Hoorick}, \binits{B.}},
\bauthor{\bsnm{Tokmakov}, \binits{P.}},
\bauthor{\bsnm{Zakharov}, \binits{S.}},
\bauthor{\bsnm{Vondrick}, \binits{C.}}:
\bctitle{Zero-1-to-3: Zero-shot one image to 3d object}.
In: \bbtitle{ICCV}
(\byear{2023})
\end{bchapter}
\endbibitem

\bibitem[\protect\citeauthoryear{Tseng et~al.}{2023}]{tseng2023consistent}
\begin{bchapter}
\bauthor{\bsnm{Tseng}, \binits{H.-Y.}},
\bauthor{\bsnm{Li}, \binits{Q.}},
\bauthor{\bsnm{Kim}, \binits{C.}},
\bauthor{\bsnm{Alsisan}, \binits{S.}},
\bauthor{\bsnm{Huang}, \binits{J.-B.}},
\bauthor{\bsnm{Kopf}, \binits{J.}}:
\bctitle{Consistent view synthesis with pose-guided diffusion models}.
In: \bbtitle{CVPR}
(\byear{2023})
\end{bchapter}
\endbibitem

\bibitem[\protect\citeauthoryear{Watson et~al.}{2022}]{watson2022novel}
\begin{botherref}
\oauthor{\bsnm{Watson}, \binits{D.}},
\oauthor{\bsnm{Chan}, \binits{W.}},
\oauthor{\bsnm{Martin-Brualla}, \binits{R.}},
\oauthor{\bsnm{Ho}, \binits{J.}},
\oauthor{\bsnm{Tagliasacchi}, \binits{A.}},
\oauthor{\bsnm{Norouzi}, \binits{M.}}:
Novel View Synthesis with Diffusion Models
(2022)
\end{botherref}
\endbibitem

\bibitem[\protect\citeauthoryear{Patashnik et~al.}{2023}]{patashnik2023localizing}
\begin{bchapter}
\bauthor{\bsnm{Patashnik}, \binits{O.}},
\bauthor{\bsnm{Garibi}, \binits{D.}},
\bauthor{\bsnm{Azuri}, \binits{I.}},
\bauthor{\bsnm{Averbuch-Elor}, \binits{H.}},
\bauthor{\bsnm{Cohen-Or}, \binits{D.}}:
\bctitle{Localizing object-level shape variations with text-to-image diffusion models}.
In: \bbtitle{ICCV}
(\byear{2023})
\end{bchapter}
\endbibitem

\bibitem[\protect\citeauthoryear{Liu et~al.}{2023}]{liu2023videop2p}
\begin{botherref}
\oauthor{\bsnm{Liu}, \binits{S.}},
\oauthor{\bsnm{Zhang}, \binits{Y.}},
\oauthor{\bsnm{Li}, \binits{W.}},
\oauthor{\bsnm{Lin}, \binits{Z.}},
\oauthor{\bsnm{Jia}, \binits{J.}}:
Video-p2p: Video editing with cross-attention control.
arXiv:2303.04761
(2023)
\end{botherref}
\endbibitem

\bibitem[\protect\citeauthoryear{QI et~al.}{2023}]{qi2023fatezero}
\begin{bchapter}
\bauthor{\bsnm{QI}, \binits{C.}},
\bauthor{\bsnm{Cun}, \binits{X.}},
\bauthor{\bsnm{Zhang}, \binits{Y.}},
\bauthor{\bsnm{Lei}, \binits{C.}},
\bauthor{\bsnm{Wang}, \binits{X.}},
\bauthor{\bsnm{Shan}, \binits{Y.}},
\bauthor{\bsnm{Chen}, \binits{Q.}}:
\bctitle{Fatezero: Fusing attentions for zero-shot text-based video editing}.
In: \bbtitle{ICCV}
(\byear{2023})
\end{bchapter}
\endbibitem

\bibitem[\protect\citeauthoryear{Ceylan et~al.}{2023}]{ceylan2023pix2video}
\begin{botherref}
\oauthor{\bsnm{Ceylan}, \binits{D.}},
\oauthor{\bsnm{Huang}, \binits{C.-H.}},
\oauthor{\bsnm{Mitra}, \binits{N.J.}}:
Pix2video: Video editing using image diffusion.
arXiv:2303.12688
(2023)
\end{botherref}
\endbibitem

\bibitem[\protect\citeauthoryear{Ma et~al.}{2023}]{ma2023directed}
\begin{botherref}
\oauthor{\bsnm{Ma}, \binits{W.-D.K.}},
\oauthor{\bsnm{Lewis}, \binits{J.}},
\oauthor{\bsnm{Kleijn}, \binits{W.B.}},
\oauthor{\bsnm{Leung}, \binits{T.}}:
Directed diffusion: Direct control of object placement through attention guidance.
arXiv preprint arXiv:2302.13153
(2023)
\end{botherref}
\endbibitem

\bibitem[\protect\citeauthoryear{Meng et~al.}{2019}]{meng2019glyce}
\begin{botherref}
\oauthor{\bsnm{Meng}, \binits{Y.}},
\oauthor{\bsnm{Wu}, \binits{W.}},
\oauthor{\bsnm{Wang}, \binits{F.}},
\oauthor{\bsnm{Li}, \binits{X.}},
\oauthor{\bsnm{Nie}, \binits{P.}},
\oauthor{\bsnm{Yin}, \binits{F.}},
\oauthor{\bsnm{Li}, \binits{M.}},
\oauthor{\bsnm{Han}, \binits{Q.}},
\oauthor{\bsnm{Sun}, \binits{X.}},
\oauthor{\bsnm{Li}, \binits{J.}}:
Glyce: Glyph-vectors for chinese character representations.
NeurIPS
\textbf{32}
(2019)
\end{botherref}
\endbibitem

\bibitem[\protect\citeauthoryear{Sun et~al.}{2021}]{sun2021chinesebert}
\begin{botherref}
\oauthor{\bsnm{Sun}, \binits{Z.}},
\oauthor{\bsnm{Li}, \binits{X.}},
\oauthor{\bsnm{Sun}, \binits{X.}},
\oauthor{\bsnm{Meng}, \binits{Y.}},
\oauthor{\bsnm{Ao}, \binits{X.}},
\oauthor{\bsnm{He}, \binits{Q.}},
\oauthor{\bsnm{Wu}, \binits{F.}},
\oauthor{\bsnm{Li}, \binits{J.}}:
Chinesebert: Chinese pretraining enhanced by glyph and pinyin information.
ACL
(2021)
\end{botherref}
\endbibitem

\bibitem[\protect\citeauthoryear{Xu et~al.}{2020}]{xu2020layoutlm}
\begin{bchapter}
\bauthor{\bsnm{Xu}, \binits{Y.}},
\bauthor{\bsnm{Li}, \binits{M.}},
\bauthor{\bsnm{Cui}, \binits{L.}},
\bauthor{\bsnm{Huang}, \binits{S.}},
\bauthor{\bsnm{Wei}, \binits{F.}},
\bauthor{\bsnm{Zhou}, \binits{M.}}:
\bctitle{Layoutlm: Pre-training of text and layout for document image understanding}.
In: \bbtitle{SIGKDD}
(\byear{2020})
\end{bchapter}
\endbibitem

\bibitem[\protect\citeauthoryear{Li et~al.}{2022}]{li2022dit}
\begin{bchapter}
\bauthor{\bsnm{Li}, \binits{J.}},
\bauthor{\bsnm{Xu}, \binits{Y.}},
\bauthor{\bsnm{Lv}, \binits{T.}},
\bauthor{\bsnm{Cui}, \binits{L.}},
\bauthor{\bsnm{Zhang}, \binits{C.}},
\bauthor{\bsnm{Wei}, \binits{F.}}:
\bctitle{Dit: Self-supervised pre-training for document image transformer}.
In: \bbtitle{MM}
(\byear{2022})
\end{bchapter}
\endbibitem

\bibitem[\protect\citeauthoryear{Gatys et~al.}{2016}]{gatys2016image}
\begin{bchapter}
\bauthor{\bsnm{Gatys}, \binits{L.A.}},
\bauthor{\bsnm{Ecker}, \binits{A.S.}},
\bauthor{\bsnm{Bethge}, \binits{M.}}:
\bctitle{Image style transfer using convolutional neural networks}.
In: \bbtitle{CVPR}
(\byear{2016})
\end{bchapter}
\endbibitem

\bibitem[\protect\citeauthoryear{Zhu et~al.}{2017}]{Zhu_2017_ICCV}
\begin{bchapter}
\bauthor{\bsnm{Zhu}, \binits{J.-Y.}},
\bauthor{\bsnm{Park}, \binits{T.}},
\bauthor{\bsnm{Isola}, \binits{P.}},
\bauthor{\bsnm{Efros}, \binits{A.A.}}:
\bctitle{Unpaired image-to-image translation using cycle-consistent adversarial networks}.
In: \bbtitle{ICCV}
(\byear{2017})
\end{bchapter}
\endbibitem

\bibitem[\protect\citeauthoryear{Luan et~al.}{2017}]{Luan2017DeepPS}
\begin{bchapter}
\bauthor{\bsnm{Luan}, \binits{F.}},
\bauthor{\bsnm{Paris}, \binits{S.}},
\bauthor{\bsnm{Shechtman}, \binits{E.}},
\bauthor{\bsnm{Bala}, \binits{K.}}:
\bctitle{Deep photo style transfer}.
In: \bbtitle{CVPR}
(\byear{2017})
\end{bchapter}
\endbibitem

\bibitem[\protect\citeauthoryear{Reinhard et~al.}{2001}]{946629}
\begin{barticle}
\bauthor{\bsnm{Reinhard}, \binits{E.}},
\bauthor{\bsnm{Adhikhmin}, \binits{M.}},
\bauthor{\bsnm{Gooch}, \binits{B.}},
\bauthor{\bsnm{Shirley}, \binits{P.}}:
\batitle{Color transfer between images}.
\bjtitle{IEEE Computer Graphics and Applications}
\bvolume{21}(\bissue{5}),
\bfpage{34}--\blpage{41}
(\byear{2001})
\doiurl{10.1109/38.946629}
\end{barticle}
\endbibitem

\bibitem[\protect\citeauthoryear{Tai et~al.}{2005}]{1467343}
\begin{bchapter}
\bauthor{\bsnm{Tai}, \binits{Y.-W.}},
\bauthor{\bsnm{Jia}, \binits{J.}},
\bauthor{\bsnm{Tang}, \binits{C.-K.}}:
\bctitle{Local color transfer via probabilistic segmentation by expectation-maximization}.
In: \bbtitle{CVPR}
(\byear{2005})
\end{bchapter}
\endbibitem

\bibitem[\protect\citeauthoryear{Xu et~al.}{2013}]{Xu2013ASC}
\begin{barticle}
\bauthor{\bsnm{Xu}, \binits{L.}},
\bauthor{\bsnm{Yan}, \binits{Q.}},
\bauthor{\bsnm{Jia}, \binits{J.}}:
\batitle{A sparse control model for image and video editing}.
\bjtitle{ACM Transactions on Graphics (TOG)}
\bvolume{32},
\bfpage{1}--\blpage{10}
(\byear{2013})
\end{barticle}
\endbibitem

\bibitem[\protect\citeauthoryear{Levin et~al.}{2004}]{Levin2004ColorizationUO}
\begin{botherref}
\oauthor{\bsnm{Levin}, \binits{A.}},
\oauthor{\bsnm{Lischinski}, \binits{D.}},
\oauthor{\bsnm{Weiss}, \binits{Y.}}:
Colorization using optimization.
SIG
(2004)
\end{botherref}
\endbibitem

\bibitem[\protect\citeauthoryear{Zhang et~al.}{2016}]{zhang2016colorful}
\begin{bchapter}
\bauthor{\bsnm{Zhang}, \binits{R.}},
\bauthor{\bsnm{Isola}, \binits{P.}},
\bauthor{\bsnm{Efros}, \binits{A.A.}}:
\bctitle{Colorful image colorization}.
In: \bbtitle{ECCV}
(\byear{2016})
\end{bchapter}
\endbibitem

\bibitem[\protect\citeauthoryear{Zhang et~al.}{2017}]{zhang2017real}
\begin{botherref}
\oauthor{\bsnm{Zhang}, \binits{R.}},
\oauthor{\bsnm{Zhu}, \binits{J.-Y.}},
\oauthor{\bsnm{Isola}, \binits{P.}},
\oauthor{\bsnm{Geng}, \binits{X.}},
\oauthor{\bsnm{Lin}, \binits{A.S.}},
\oauthor{\bsnm{Yu}, \binits{T.}},
\oauthor{\bsnm{Efros}, \binits{A.A.}}:
Real-time user-guided image colorization with learned deep priors.
ACM Transactions on Graphics (TOG)
\textbf{9}(4)
(2017)
\end{botherref}
\endbibitem

\bibitem[\protect\citeauthoryear{Betker et~al.}{2023}]{betker2023improving}
\begin{barticle}
\bauthor{\bsnm{Betker}, \binits{J.}},
\bauthor{\bsnm{Goh}, \binits{G.}},
\bauthor{\bsnm{Jing}, \binits{L.}},
\bauthor{\bsnm{Brooks}, \binits{T.}},
\bauthor{\bsnm{Wang}, \binits{J.}},
\bauthor{\bsnm{Li}, \binits{L.}},
\bauthor{\bsnm{Ouyang}, \binits{L.}},
\bauthor{\bsnm{Zhuang}, \binits{J.}},
\bauthor{\bsnm{Lee}, \binits{J.}},
\bauthor{\bsnm{Guo}, \binits{Y.}}, \betal:
\batitle{Improving image generation with better captions}.
\bjtitle{Computer Science. https://cdn. openai. com/papers/dall-e-3. pdf}
\bvolume{2}(\bissue{3}),
\bfpage{8}
(\byear{2023})
\end{barticle}
\endbibitem

\bibitem[\protect\citeauthoryear{Wang et~al.}{2024}]{wang2024instancediffusion}
\begin{botherref}
\oauthor{\bsnm{Wang}, \binits{X.}},
\oauthor{\bsnm{Darrell}, \binits{T.}},
\oauthor{\bsnm{Rambhatla}, \binits{S.S.}},
\oauthor{\bsnm{Girdhar}, \binits{R.}},
\oauthor{\bsnm{Misra}, \binits{I.}}:
Instancediffusion: Instance-level control for image generation.
arXiv preprint arXiv:2402.03290
(2024)
\end{botherref}
\endbibitem

\bibitem[\protect\citeauthoryear{Wu et~al.}{2023}]{wu2023paragraph}
\begin{botherref}
\oauthor{\bsnm{Wu}, \binits{W.}},
\oauthor{\bsnm{Li}, \binits{Z.}},
\oauthor{\bsnm{He}, \binits{Y.}},
\oauthor{\bsnm{Shou}, \binits{M.Z.}},
\oauthor{\bsnm{Shen}, \binits{C.}},
\oauthor{\bsnm{Cheng}, \binits{L.}},
\oauthor{\bsnm{Li}, \binits{Y.}},
\oauthor{\bsnm{Gao}, \binits{T.}},
\oauthor{\bsnm{Zhang}, \binits{D.}},
\oauthor{\bsnm{Wang}, \binits{Z.}}:
Paragraph-to-image generation with information-enriched diffusion model.
arXiv preprint arXiv:2311.14284
(2023)
\end{botherref}
\endbibitem

\bibitem[\protect\citeauthoryear{Bakr et~al.}{2023}]{bakr2023hrs}
\begin{bchapter}
\bauthor{\bsnm{Bakr}, \binits{E.M.}},
\bauthor{\bsnm{Sun}, \binits{P.}},
\bauthor{\bsnm{Shen}, \binits{X.}},
\bauthor{\bsnm{Khan}, \binits{F.F.}},
\bauthor{\bsnm{Li}, \binits{L.E.}},
\bauthor{\bsnm{Elhoseiny}, \binits{M.}}:
\bctitle{Hrs-bench: Holistic, reliable and scalable benchmark for text-to-image models}.
In: \bbtitle{ICCV}
(\byear{2023})
\end{bchapter}
\endbibitem

\bibitem[\protect\citeauthoryear{Hu et~al.}{2023}]{hu2023tifa}
\begin{bchapter}
\bauthor{\bsnm{Hu}, \binits{Y.}},
\bauthor{\bsnm{Liu}, \binits{B.}},
\bauthor{\bsnm{Kasai}, \binits{J.}},
\bauthor{\bsnm{Wang}, \binits{Y.}},
\bauthor{\bsnm{Ostendorf}, \binits{M.}},
\bauthor{\bsnm{Krishna}, \binits{R.}},
\bauthor{\bsnm{Smith}, \binits{N.A.}}:
\bctitle{Tifa: Accurate and interpretable text-to-image faithfulness evaluation with question answering}.
In: \bbtitle{ICCV}
(\byear{2023})
\end{bchapter}
\endbibitem

\bibitem[\protect\citeauthoryear{Huang et~al.}{2024}]{huang2024t2i}
\begin{bchapter}
\bauthor{\bsnm{Huang}, \binits{K.}},
\bauthor{\bsnm{Sun}, \binits{K.}},
\bauthor{\bsnm{Xie}, \binits{E.}},
\bauthor{\bsnm{Li}, \binits{Z.}},
\bauthor{\bsnm{Liu}, \binits{X.}}:
\bctitle{T2i-compbench: A comprehensive benchmark for open-world compositional text-to-image generation}.
In: \bbtitle{NeurIPS}
(\byear{2024})
\end{bchapter}
\endbibitem

\bibitem[\protect\citeauthoryear{Patel et~al.}{2024}]{patel2024conceptbed}
\begin{bchapter}
\bauthor{\bsnm{Patel}, \binits{M.}},
\bauthor{\bsnm{Gokhale}, \binits{T.}},
\bauthor{\bsnm{Baral}, \binits{C.}},
\bauthor{\bsnm{Yang}, \binits{Y.}}:
\bctitle{Conceptbed: Evaluating concept learning abilities of text-to-image diffusion models}.
In: \bbtitle{AAAI}
(\byear{2024})
\end{bchapter}
\endbibitem

\bibitem[\protect\citeauthoryear{Zhao et~al.}{2024}]{zhao2024flasheval}
\begin{botherref}
\oauthor{\bsnm{Zhao}, \binits{L.}},
\oauthor{\bsnm{Zhao}, \binits{T.}},
\oauthor{\bsnm{Lin}, \binits{Z.}},
\oauthor{\bsnm{Ning}, \binits{X.}},
\oauthor{\bsnm{Dai}, \binits{G.}},
\oauthor{\bsnm{Yang}, \binits{H.}},
\oauthor{\bsnm{Wang}, \binits{Y.}}:
Flasheval: Towards fast and accurate evaluation of text-to-image diffusion generative models.
arXiv preprint arXiv:2403.16379
(2024)
\end{botherref}
\endbibitem

\bibitem[\protect\citeauthoryear{Sahuguet and Azavant}{1999}]{sahuguet1999wysiwyg}
\begin{botherref}
\oauthor{\bsnm{Sahuguet}, \binits{A.}},
\oauthor{\bsnm{Azavant}, \binits{F.}}:
Wysiwyg web wrapper factory (w4f)
(1999)
\end{botherref}
\endbibitem

\bibitem[\protect\citeauthoryear{Litt et~al.}{2022}]{litt2022peritext}
\begin{botherref}
\oauthor{\bsnm{Litt}, \binits{G.}},
\oauthor{\bsnm{Lim}, \binits{S.}},
\oauthor{\bsnm{Kleppmann}, \binits{M.}},
\oauthor{\bsnm{Hardenberg}, \binits{P.}}:
Peritext: A crdt for collaborative rich text editing.
Proceedings of the ACM on Human-Computer Interaction (PACMHCI)
(2022)
\end{botherref}
\endbibitem

\bibitem[\protect\citeauthoryear{Ignat et~al.}{2021}]{ignat2021enhancing}
\begin{barticle}
\bauthor{\bsnm{Ignat}, \binits{C.-L.}},
\bauthor{\bsnm{Andr{\'e}}, \binits{L.}},
\bauthor{\bsnm{Oster}, \binits{G.}}:
\batitle{Enhancing rich content wikis with real-time collaboration}.
\bjtitle{Concurrency and Computation: Practice and Experience}
\bvolume{33}(\bissue{8}),
\bfpage{4110}
(\byear{2021})
\end{barticle}
\endbibitem

\bibitem[\protect\citeauthoryear{Agrawala}{2023}]{magrawala2023unpredictable}
\begin{botherref}
\oauthor{\bsnm{Agrawala}, \binits{M.}}:
Unpredictable Black Boxes are Terrible Interfaces.
https://magrawala.substack.com/p/unpredictable-black-boxes-are-terrible
(2023)
\end{botherref}
\endbibitem

\bibitem[\protect\citeauthoryear{Karpathy and Fei-Fei}{2015}]{karpathy2015deep}
\begin{bchapter}
\bauthor{\bsnm{Karpathy}, \binits{A.}},
\bauthor{\bsnm{Fei-Fei}, \binits{L.}}:
\bctitle{Deep visual-semantic alignments for generating image descriptions}.
In: \bbtitle{CVPR}
(\byear{2015})
\end{bchapter}
\endbibitem

\bibitem[\protect\citeauthoryear{Johnson et~al.}{2016}]{johnson2016densecap}
\begin{bchapter}
\bauthor{\bsnm{Johnson}, \binits{J.}},
\bauthor{\bsnm{Karpathy}, \binits{A.}},
\bauthor{\bsnm{Fei-Fei}, \binits{L.}}:
\bctitle{Densecap: Fully convolutional localization networks for dense captioning}.
In: \bbtitle{CVPR}
(\byear{2016})
\end{bchapter}
\endbibitem

\bibitem[\protect\citeauthoryear{Radford et~al.}{2021}]{radford2021learning}
\begin{bchapter}
\bauthor{\bsnm{Radford}, \binits{A.}},
\bauthor{\bsnm{Kim}, \binits{J.W.}},
\bauthor{\bsnm{Hallacy}, \binits{C.}},
\bauthor{\bsnm{Ramesh}, \binits{A.}},
\bauthor{\bsnm{Goh}, \binits{G.}},
\bauthor{\bsnm{Agarwal}, \binits{S.}},
\bauthor{\bsnm{Sastry}, \binits{G.}},
\bauthor{\bsnm{Askell}, \binits{A.}},
\bauthor{\bsnm{Mishkin}, \binits{P.}},
\bauthor{\bsnm{Clark}, \binits{J.}}, \betal:
\bctitle{Learning transferable visual models from natural language supervision}.
In: \bbtitle{ICML},
pp. \bfpage{8748}--\blpage{8763}
(\byear{2021}).
\bcomment{PMLR}
\end{bchapter}
\endbibitem

\bibitem[\protect\citeauthoryear{Gal et~al.}{2023}]{gal2023an}
\begin{bchapter}
\bauthor{\bsnm{Gal}, \binits{R.}},
\bauthor{\bsnm{Alaluf}, \binits{Y.}},
\bauthor{\bsnm{Atzmon}, \binits{Y.}},
\bauthor{\bsnm{Patashnik}, \binits{O.}},
\bauthor{\bsnm{Bermano}, \binits{A.H.}},
\bauthor{\bsnm{Chechik}, \binits{G.}},
\bauthor{\bsnm{Cohen-or}, \binits{D.}}:
\bctitle{An image is worth one word: Personalizing text-to-image generation using textual inversion}.
In: \bbtitle{ICLR}
(\byear{2023}).
\burl{https://openreview.net/forum?id=NAQvF08TcyG}
\end{bchapter}
\endbibitem

\bibitem[\protect\citeauthoryear{Chen et~al.}{2023}]{chen2023anydoor}
\begin{botherref}
\oauthor{\bsnm{Chen}, \binits{X.}},
\oauthor{\bsnm{Huang}, \binits{L.}},
\oauthor{\bsnm{Liu}, \binits{Y.}},
\oauthor{\bsnm{Shen}, \binits{Y.}},
\oauthor{\bsnm{Zhao}, \binits{D.}},
\oauthor{\bsnm{Zhao}, \binits{H.}}:
Anydoor: Zero-shot object-level image customization.
arXiv preprint arXiv:2307.09481
(2023)
\end{botherref}
\endbibitem

\bibitem[\protect\citeauthoryear{Li et~al.}{2023}]{li2023blip}
\begin{botherref}
\oauthor{\bsnm{Li}, \binits{D.}},
\oauthor{\bsnm{Li}, \binits{J.}},
\oauthor{\bsnm{Hoi}, \binits{S.C.}}:
Blip-diffusion: Pre-trained subject representation for controllable text-to-image generation and editing.
arXiv preprint arXiv:2305.14720
(2023)
\end{botherref}
\endbibitem

\bibitem[\protect\citeauthoryear{Jia et~al.}{2023}]{jia2023taming}
\begin{botherref}
\oauthor{\bsnm{Jia}, \binits{X.}},
\oauthor{\bsnm{Zhao}, \binits{Y.}},
\oauthor{\bsnm{Chan}, \binits{K.C.}},
\oauthor{\bsnm{Li}, \binits{Y.}},
\oauthor{\bsnm{Zhang}, \binits{H.}},
\oauthor{\bsnm{Gong}, \binits{B.}},
\oauthor{\bsnm{Hou}, \binits{T.}},
\oauthor{\bsnm{Wang}, \binits{H.}},
\oauthor{\bsnm{Su}, \binits{Y.-C.}}:
Taming encoder for zero fine-tuning image customization with text-to-image diffusion models.
arXiv preprint arXiv:2304.02642
(2023)
\end{botherref}
\endbibitem

\bibitem[\protect\citeauthoryear{Chen et~al.}{2023}]{chen2023subject}
\begin{botherref}
\oauthor{\bsnm{Chen}, \binits{W.}},
\oauthor{\bsnm{Hu}, \binits{H.}},
\oauthor{\bsnm{Li}, \binits{Y.}},
\oauthor{\bsnm{Rui}, \binits{N.}},
\oauthor{\bsnm{Jia}, \binits{X.}},
\oauthor{\bsnm{Chang}, \binits{M.-W.}},
\oauthor{\bsnm{Cohen}, \binits{W.W.}}:
Subject-driven text-to-image generation via apprenticeship learning.
arXiv preprint arXiv:2304.00186
(2023)
\end{botherref}
\endbibitem

\bibitem[\protect\citeauthoryear{Gal et~al.}{2023}]{gal2023encoder}
\begin{barticle}
\bauthor{\bsnm{Gal}, \binits{R.}},
\bauthor{\bsnm{Arar}, \binits{M.}},
\bauthor{\bsnm{Atzmon}, \binits{Y.}},
\bauthor{\bsnm{Bermano}, \binits{A.H.}},
\bauthor{\bsnm{Chechik}, \binits{G.}},
\bauthor{\bsnm{Cohen-Or}, \binits{D.}}:
\batitle{Encoder-based domain tuning for fast personalization of text-to-image models}.
\bjtitle{ACM Transactions on Graphics (TOG)}
\bvolume{42}(\bissue{4}),
\bfpage{1}--\blpage{13}
(\byear{2023})
\end{barticle}
\endbibitem

\bibitem[\protect\citeauthoryear{Shi et~al.}{2023}]{shi2023instantbooth}
\begin{botherref}
\oauthor{\bsnm{Shi}, \binits{J.}},
\oauthor{\bsnm{Xiong}, \binits{W.}},
\oauthor{\bsnm{Lin}, \binits{Z.}},
\oauthor{\bsnm{Jung}, \binits{H.J.}}:
Instantbooth: Personalized text-to-image generation without test-time finetuning.
arXiv preprint arXiv:2304.03411
(2023)
\end{botherref}
\endbibitem

\bibitem[\protect\citeauthoryear{Tang et~al.}{2022}]{Tang2022WhatTD}
\begin{botherref}
\oauthor{\bsnm{Tang}, \binits{R.}},
\oauthor{\bsnm{Pandey}, \binits{A.}},
\oauthor{\bsnm{Jiang}, \binits{Z.}},
\oauthor{\bsnm{Yang}, \binits{G.}},
\oauthor{\bsnm{Kumar}, \binits{K.V.S.M.}},
\oauthor{\bsnm{Lin}, \binits{J.}},
\oauthor{\bsnm{Ture}, \binits{F.}}:
What the daam: Interpreting stable diffusion using cross attention.
ArXiv
\textbf{abs/2210.04885}
(2022)
\end{botherref}
\endbibitem

\bibitem[\protect\citeauthoryear{Tumanyan et~al.}{2023}]{tumanyan2022plug}
\begin{bchapter}
\bauthor{\bsnm{Tumanyan}, \binits{N.}},
\bauthor{\bsnm{Geyer}, \binits{M.}},
\bauthor{\bsnm{Bagon}, \binits{S.}},
\bauthor{\bsnm{Dekel}, \binits{T.}}:
\bctitle{Plug-and-play diffusion features for text-driven image-to-image translation}.
In: \bbtitle{CVPR}
(\byear{2023})
\end{bchapter}
\endbibitem

\bibitem[\protect\citeauthoryear{Shi and Malik}{2000}]{868688}
\begin{barticle}
\bauthor{\bsnm{Shi}, \binits{J.}},
\bauthor{\bsnm{Malik}, \binits{J.}}:
\batitle{Normalized cuts and image segmentation}.
\bjtitle{IEEE Transactions on Pattern Analysis and Machine Intelligence}
\bvolume{22}(\bissue{8}),
\bfpage{888}--\blpage{905}
(\byear{2000})
\doiurl{10.1109/34.868688}
\end{barticle}
\endbibitem

\bibitem[\protect\citeauthoryear{Von~Luxburg}{2007}]{von2007tutorial}
\begin{barticle}
\bauthor{\bsnm{Von~Luxburg}, \binits{U.}}:
\batitle{A tutorial on spectral clustering}.
\bjtitle{Statistics and computing}
\bvolume{17},
\bfpage{395}--\blpage{416}
(\byear{2007})
\end{barticle}
\endbibitem

\bibitem[\protect\citeauthoryear{Wen et~al.}{2023}]{wen2023hard}
\begin{bchapter}
\bauthor{\bsnm{Wen}, \binits{Y.}},
\bauthor{\bsnm{Jain}, \binits{N.}},
\bauthor{\bsnm{Kirchenbauer}, \binits{J.}},
\bauthor{\bsnm{Goldblum}, \binits{M.}},
\bauthor{\bsnm{Geiping}, \binits{J.}},
\bauthor{\bsnm{Goldstein}, \binits{T.}}:
\bctitle{Hard prompts made easy: Gradient-based discrete optimization for prompt tuning and discovery}.
In: \bbtitle{NeurIPS}
(\byear{2023})
\end{bchapter}
\endbibitem

\bibitem[\protect\citeauthoryear{Ho and Salimans}{2022}]{ho2022classifier}
\begin{botherref}
\oauthor{\bsnm{Ho}, \binits{J.}},
\oauthor{\bsnm{Salimans}, \binits{T.}}:
Classifier-free diffusion guidance.
arXiv preprint arXiv:2207.12598
(2022)
\end{botherref}
\endbibitem

\bibitem[\protect\citeauthoryear{Dhariwal and Nichol}{}]{NEURIPS2021_49ad23d1}
\begin{botherref}
\oauthor{\bsnm{Dhariwal}, \binits{P.}},
\oauthor{\bsnm{Nichol}, \binits{A.}}:
Diffusion models beat gans on image synthesis.
In: NeurIPS
\end{botherref}
\endbibitem

\bibitem[\protect\citeauthoryear{Johnson et~al.}{2016}]{johnson2016perceptual}
\begin{bchapter}
\bauthor{\bsnm{Johnson}, \binits{J.}},
\bauthor{\bsnm{Alahi}, \binits{A.}},
\bauthor{\bsnm{Fei-Fei}, \binits{L.}}:
\bctitle{Perceptual losses for real-time style transfer and super-resolution}.
In: \bbtitle{ECCV}
(\byear{2016})
\end{bchapter}
\endbibitem

\bibitem[\protect\citeauthoryear{Podell et~al.}{2024}]{podell2024sdxl}
\begin{bchapter}
\bauthor{\bsnm{Podell}, \binits{D.}},
\bauthor{\bsnm{English}, \binits{Z.}},
\bauthor{\bsnm{Lacey}, \binits{K.}},
\bauthor{\bsnm{Blattmann}, \binits{A.}},
\bauthor{\bsnm{Dockhorn}, \binits{T.}},
\bauthor{\bsnm{M{\"u}ller}, \binits{J.}},
\bauthor{\bsnm{Penna}, \binits{J.}},
\bauthor{\bsnm{Rombach}, \binits{R.}}:
\bctitle{{SDXL}: Improving latent diffusion models for high-resolution image synthesis}.
In: \bbtitle{ICLR}
(\byear{2024})
\end{bchapter}
\endbibitem

\bibitem[\protect\citeauthoryear{OpenAI}{2023}]{openai2023gpt4}
\begin{botherref}
\oauthor{\bsnm{OpenAI}}:
GPT-4 Technical Report
(2023)
\end{botherref}
\endbibitem

\bibitem[\protect\citeauthoryear{Li et~al.}{2022}]{li2022mplug}
\begin{bchapter}
\bauthor{\bsnm{Li}, \binits{C.}},
\bauthor{\bsnm{Xu}, \binits{H.}},
\bauthor{\bsnm{Tian}, \binits{J.}},
\bauthor{\bsnm{Wang}, \binits{W.}},
\bauthor{\bsnm{Yan}, \binits{M.}},
\bauthor{\bsnm{Bi}, \binits{B.}},
\bauthor{\bsnm{Ye}, \binits{J.}},
\bauthor{\bsnm{Chen}, \binits{H.}},
\bauthor{\bsnm{Xu}, \binits{G.}},
\bauthor{\bsnm{Cao}, \binits{Z.}}, \betal:
\bctitle{mplug: Effective and efficient vision-language learning by cross-modal skip-connections}.
In: \bbtitle{EMNLP}
(\byear{2022})
\end{bchapter}
\endbibitem

\bibitem[\protect\citeauthoryear{Huberman-Spiegelglas et~al.}{2023}]{HubermanSpiegelglas2023}
\begin{botherref}
\oauthor{\bsnm{Huberman-Spiegelglas}, \binits{I.}},
\oauthor{\bsnm{Kulikov}, \binits{V.}},
\oauthor{\bsnm{Michaeli}, \binits{T.}}:
An edit friendly ddpm noise space: Inversion and manipulations.
arXiv preprint arXiv:2304.06140
(2023)
\end{botherref}
\endbibitem

\bibitem[\protect\citeauthoryear{Wu and De~la Torre}{2022}]{wu2022unifying}
\begin{botherref}
\oauthor{\bsnm{Wu}, \binits{C.H.}},
\oauthor{\bsnm{Torre}, \binits{F.}}:
Unifying diffusion models' latent space, with applications to cyclediffusion and guidance.
arXiv preprint arXiv:2210.05559
(2022)
\end{botherref}
\endbibitem

\bibitem[\protect\citeauthoryear{Ju et~al.}{2024}]{ju2024pnp}
\begin{bchapter}
\bauthor{\bsnm{Ju}, \binits{X.}},
\bauthor{\bsnm{Zeng}, \binits{A.}},
\bauthor{\bsnm{Bian}, \binits{Y.}},
\bauthor{\bsnm{Liu}, \binits{S.}},
\bauthor{\bsnm{Xu}, \binits{Q.}}:
\bctitle{Pnp inversion: Boosting diffusion-based editing with 3 lines of code}.
In: \bbtitle{ICLR}
(\byear{2024})
\end{bchapter}
\endbibitem

\bibitem[\protect\citeauthoryear{Kirillov et~al.}{2023}]{kirillov2023segany}
\begin{bchapter}
\bauthor{\bsnm{Kirillov}, \binits{A.}},
\bauthor{\bsnm{Mintun}, \binits{E.}},
\bauthor{\bsnm{Ravi}, \binits{N.}},
\bauthor{\bsnm{Mao}, \binits{H.}},
\bauthor{\bsnm{Rolland}, \binits{C.}},
\bauthor{\bsnm{Gustafson}, \binits{L.}},
\bauthor{\bsnm{Xiao}, \binits{T.}},
\bauthor{\bsnm{Whitehead}, \binits{S.}},
\bauthor{\bsnm{Berg}, \binits{A.C.}},
\bauthor{\bsnm{Lo}, \binits{W.-Y.}},
\bauthor{\bsnm{Dollar}, \binits{P.}},
\bauthor{\bsnm{Girshick}, \binits{R.}}:
\bctitle{Segment anything}.
In: \bbtitle{ICCV}
(\byear{2023})
\end{bchapter}
\endbibitem

\bibitem[\protect\citeauthoryear{Liu et~al.}{2023}]{liu2023grounding}
\begin{botherref}
\oauthor{\bsnm{Liu}, \binits{S.}},
\oauthor{\bsnm{Zeng}, \binits{Z.}},
\oauthor{\bsnm{Ren}, \binits{T.}},
\oauthor{\bsnm{Li}, \binits{F.}},
\oauthor{\bsnm{Zhang}, \binits{H.}},
\oauthor{\bsnm{Yang}, \binits{J.}},
\oauthor{\bsnm{Li}, \binits{C.}},
\oauthor{\bsnm{Yang}, \binits{J.}},
\oauthor{\bsnm{Su}, \binits{H.}},
\oauthor{\bsnm{Zhu}, \binits{J.}}, et al.:
Grounding dino: Marrying dino with grounded pre-training for open-set object detection.
arXiv preprint arXiv:2303.05499
(2023)
\end{botherref}
\endbibitem

\bibitem[\protect\citeauthoryear{Ren et~al.}{2024}]{ren2024grounded}
\begin{botherref}
\oauthor{\bsnm{Ren}, \binits{T.}},
\oauthor{\bsnm{Liu}, \binits{S.}},
\oauthor{\bsnm{Zeng}, \binits{A.}},
\oauthor{\bsnm{Lin}, \binits{J.}},
\oauthor{\bsnm{Li}, \binits{K.}},
\oauthor{\bsnm{Cao}, \binits{H.}},
\oauthor{\bsnm{Chen}, \binits{J.}},
\oauthor{\bsnm{Huang}, \binits{X.}},
\oauthor{\bsnm{Chen}, \binits{Y.}},
\oauthor{\bsnm{Yan}, \binits{F.}},
\oauthor{\bsnm{Zeng}, \binits{Z.}},
\oauthor{\bsnm{Zhang}, \binits{H.}},
\oauthor{\bsnm{Li}, \binits{F.}},
\oauthor{\bsnm{Yang}, \binits{J.}},
\oauthor{\bsnm{Li}, \binits{H.}},
\oauthor{\bsnm{Jiang}, \binits{Q.}},
\oauthor{\bsnm{Zhang}, \binits{L.}}:
Grounded SAM: Assembling Open-World Models for Diverse Visual Tasks
(2024)
\end{botherref}
\endbibitem

\end{thebibliography}

\backmatter


\onecolumn

\appendix
In this appendix, we provide additional experimental results and details. 
In section~\ref{sec:addtional_results}, we show the images generated by our model, Attend-and-Excite~\citep{chefer2023attend}, Prompt-to-Prompt~\citep{hertz2022prompt}, and InstructPix2Pix~\citep{brooks2022instructpix2pix} with various RGB colors, local styles, and detailed descriptions via footnotes. 
In section~\ref{sec:addtional_details}, we provide additional details on the implementation and evaluation.
\section{Additional Results}
\label{sec:addtional_results}
In this section, we first show additional results of rich-text-to-image generation on complex scene synthesis (Figures 15, 16, and 17), precise color rendering (Figures 18, 19, and 20), local style control (Figures 21 and 22), and explicit token re-weighting (Figure 23, 24, and 25). We also show an ablation study of the averaging and maximizing operations across tokens to obtain token maps in Figure 26. We present additional results compared with a composition-based baseline in Figure 27. Last, we show an ablation of the hyperparameters of our baseline method InstructPix2Pix~\citep{brooks2022instructpix2pix} on the local style generation application in Figure 28.

\begin{figure}[ht]
    \centering
    \includegraphics[width=\linewidth, trim=0 0 0 0, clip]{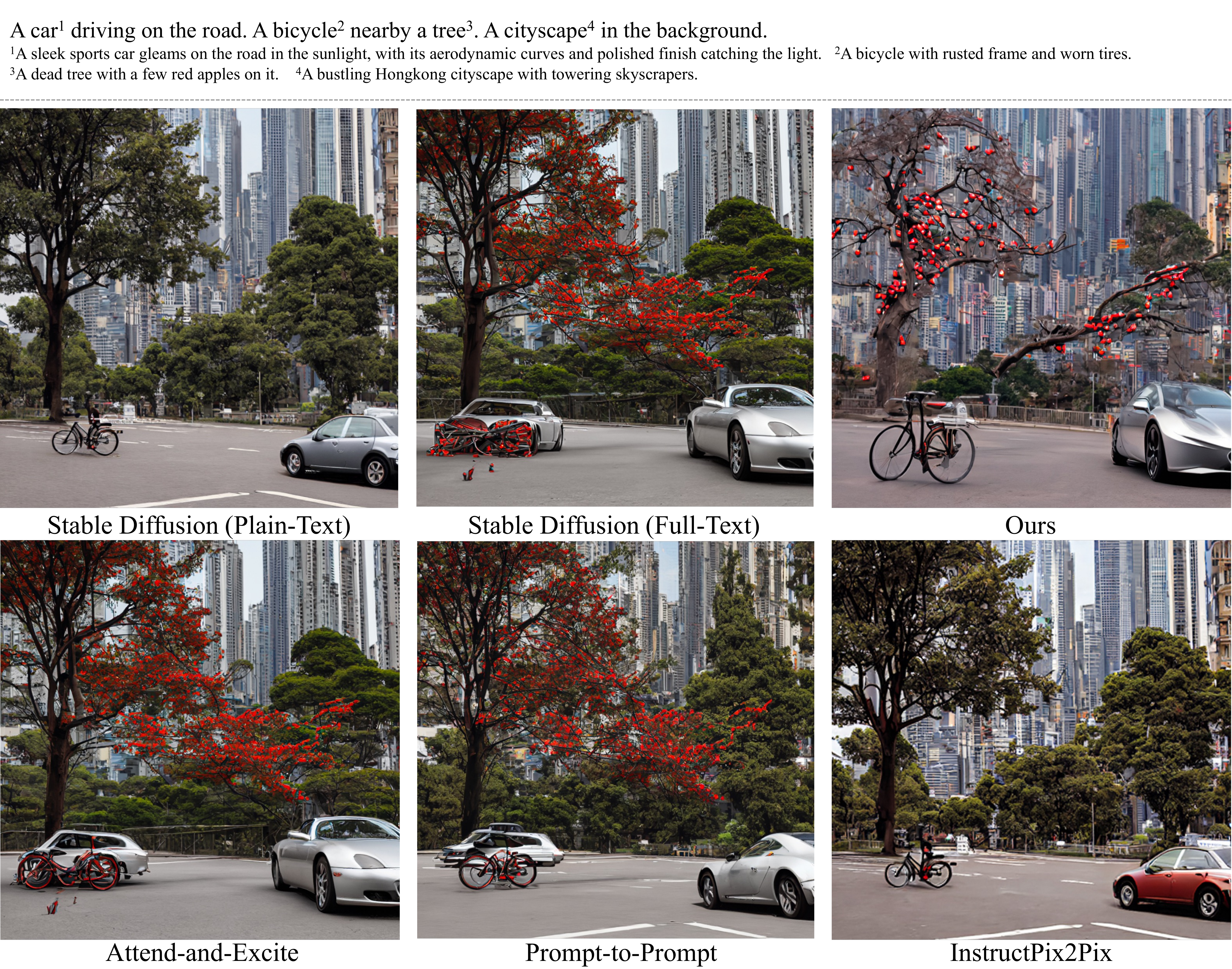}
    \caption{\textbf{Additional results of the footnote.} We show the generation from a complex description of a garden. Note that all the methods except for ours fail to generate accurate details of the mansion and fountain as described.}
    \label{fig:appendix_footnote4}
\end{figure}

\begin{figure}[ht]
    \centering
    \includegraphics[width=\linewidth, trim=0 0 0 0, clip]{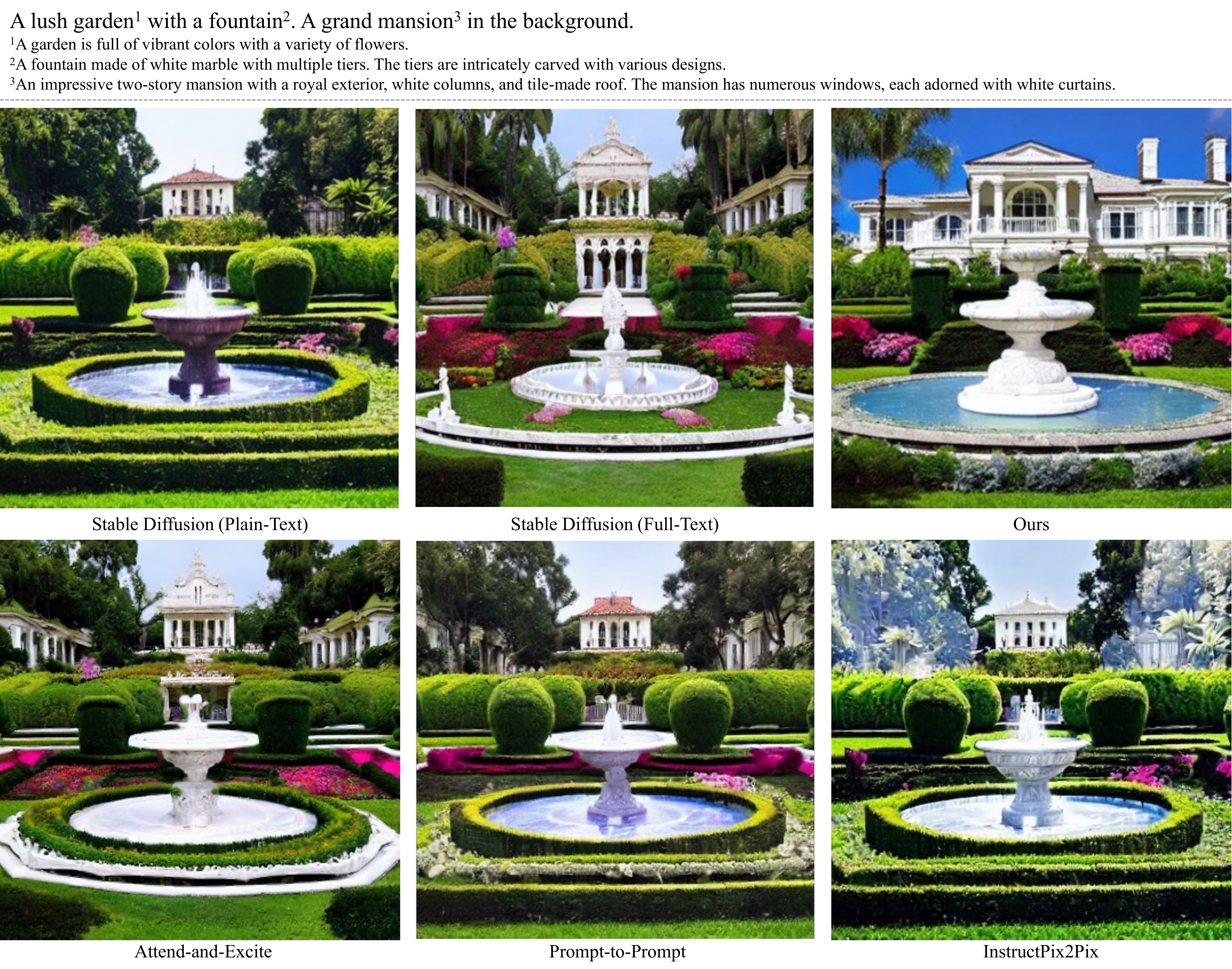}
    \caption{\textbf{Additional results of the footnote.} We show the generation from a complex description of a garden. Note that all the methods except for ours fail to generate accurate details of the mansion and fountain as described.}
    \label{fig:appendix_footnote2}
\end{figure}

\begin{figure}[ht]
    \centering
    \includegraphics[width=\linewidth, trim=0 0 0 0, clip]{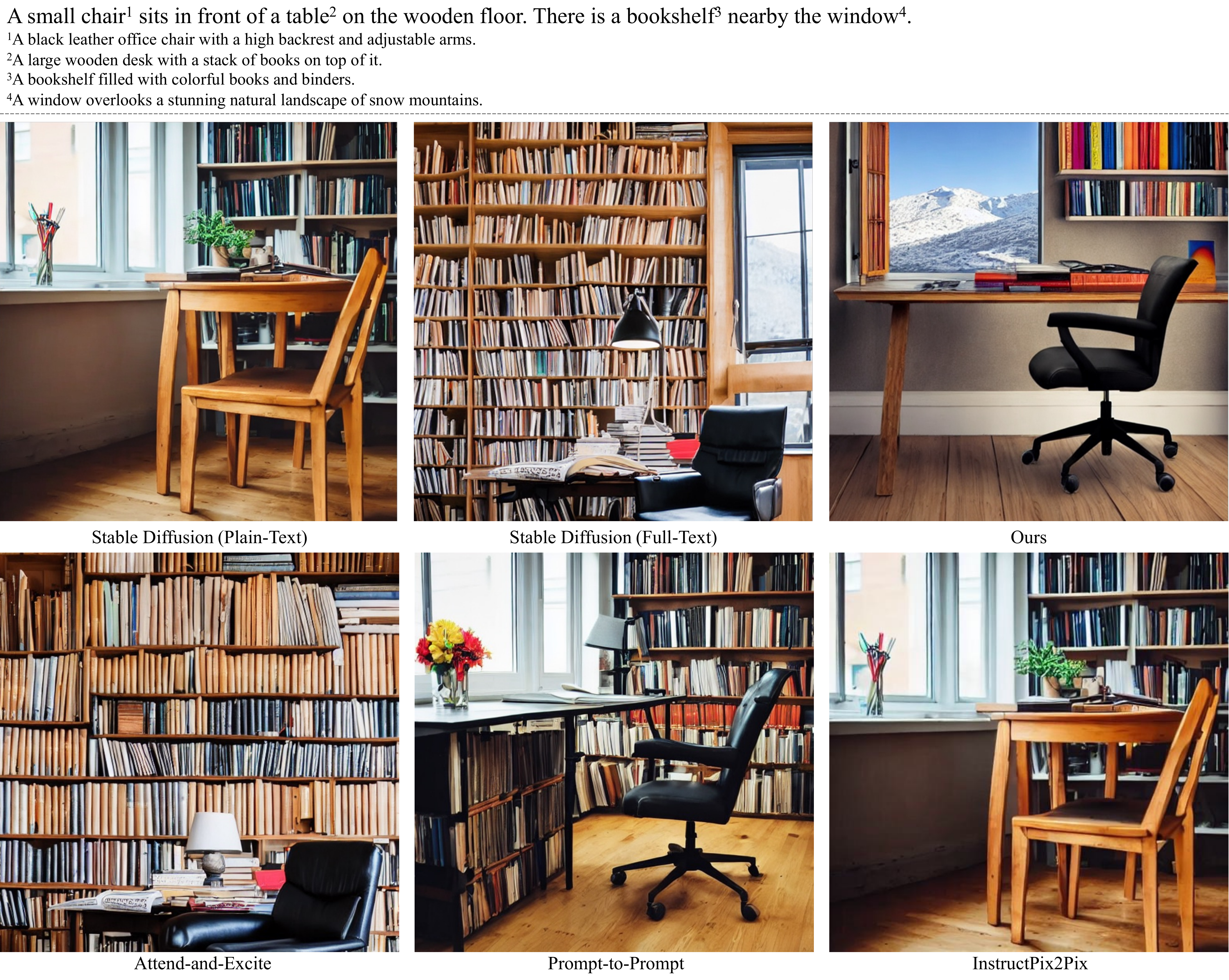}
    \caption{\textbf{Additional results of the footnote.} We show the generation from a complex description of an office. Note that all the methods except ours fail to generate accurate window overlooks and colorful binders as described.}
    \label{fig:appendix_footnote1}
\end{figure}

\begin{figure}[ht]
    \centering
    \includegraphics[width=\linewidth, trim=0 0 0 0, clip]{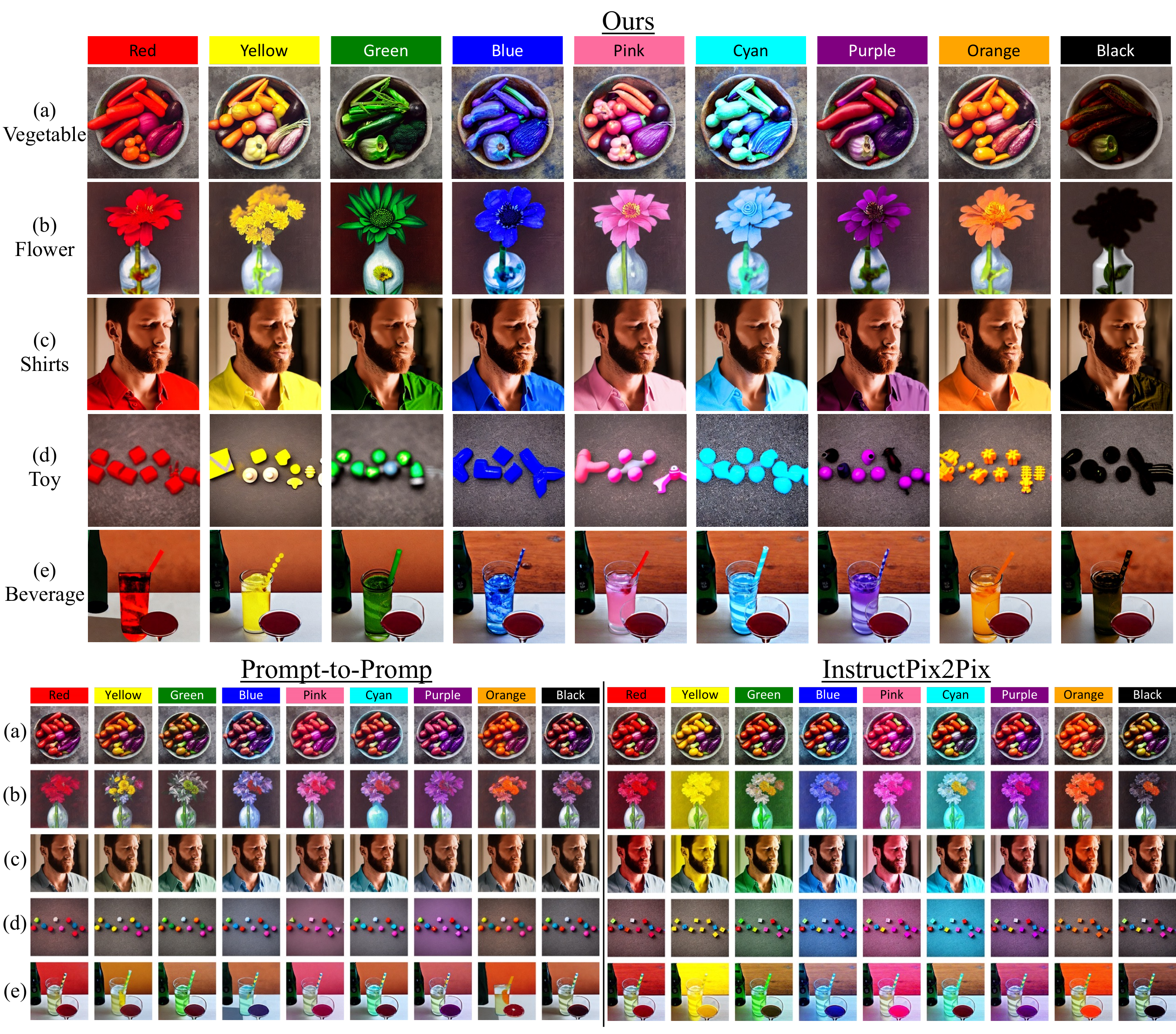}
    \caption{\textbf{Additional results of the font color.} We show the generation of different objects with colors from the \emph{Common category}. Prompt-to-Prompt has a large failure rate of respecting the given color name, while InstructPix2Pix tends to color the background and irrelevant objects.}
    \label{fig:appendix_color1}
\end{figure}

\begin{figure}[ht]
    \centering
    \includegraphics[width=\linewidth, trim=0 0 0 0, clip]{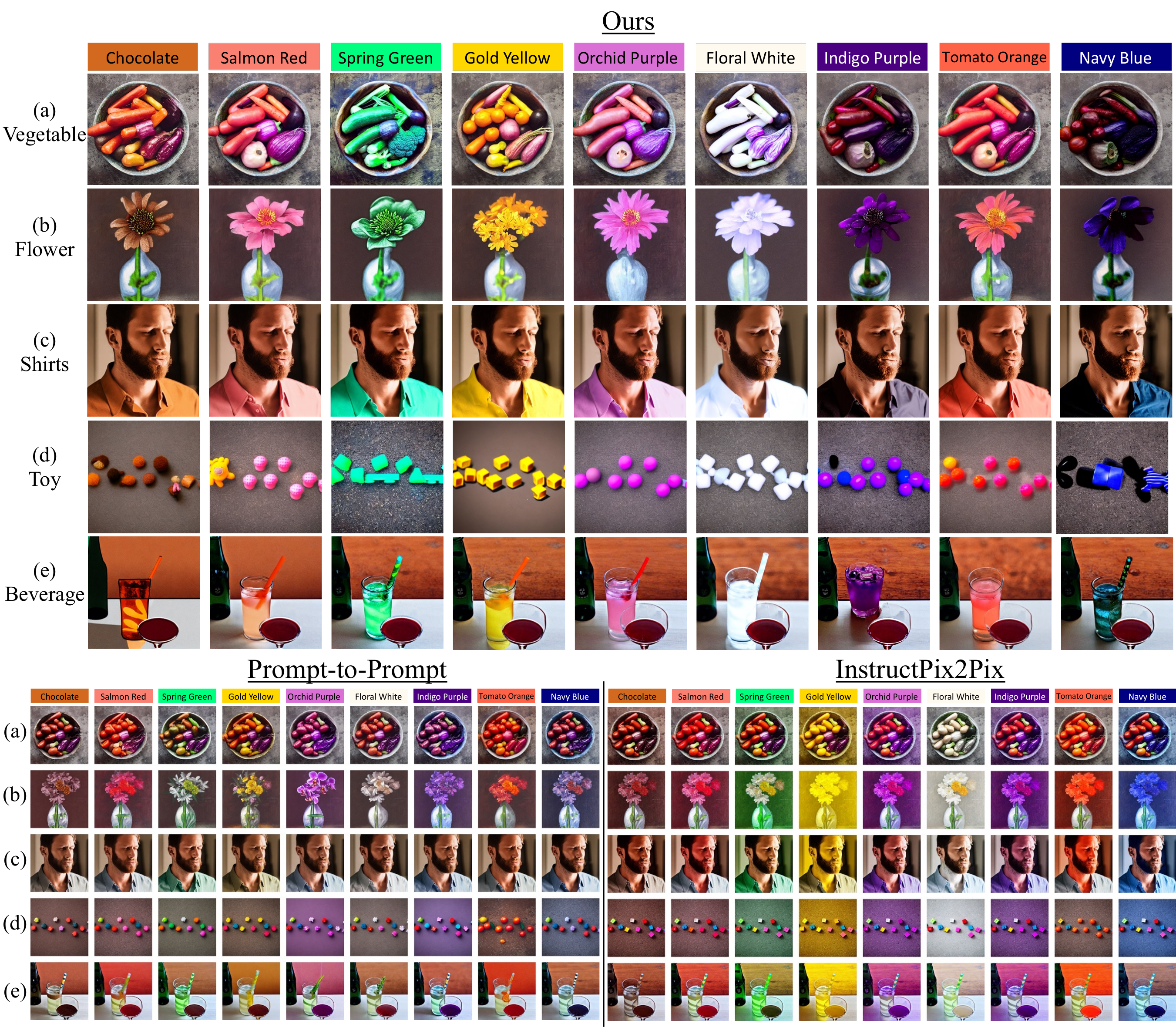}
    \caption{\textbf{Additional results of the font color.} We show the generation of different objects with colors from the \emph{HTML category}. Both methods fail to generate the precise color, and InstructPix2Pix tends to color the background and irrelevant objects.}
    \label{fig:appendix_color2}
\end{figure}

\begin{figure}[ht]
    \centering
    \includegraphics[width=\linewidth, trim=0 0 0 0, clip]{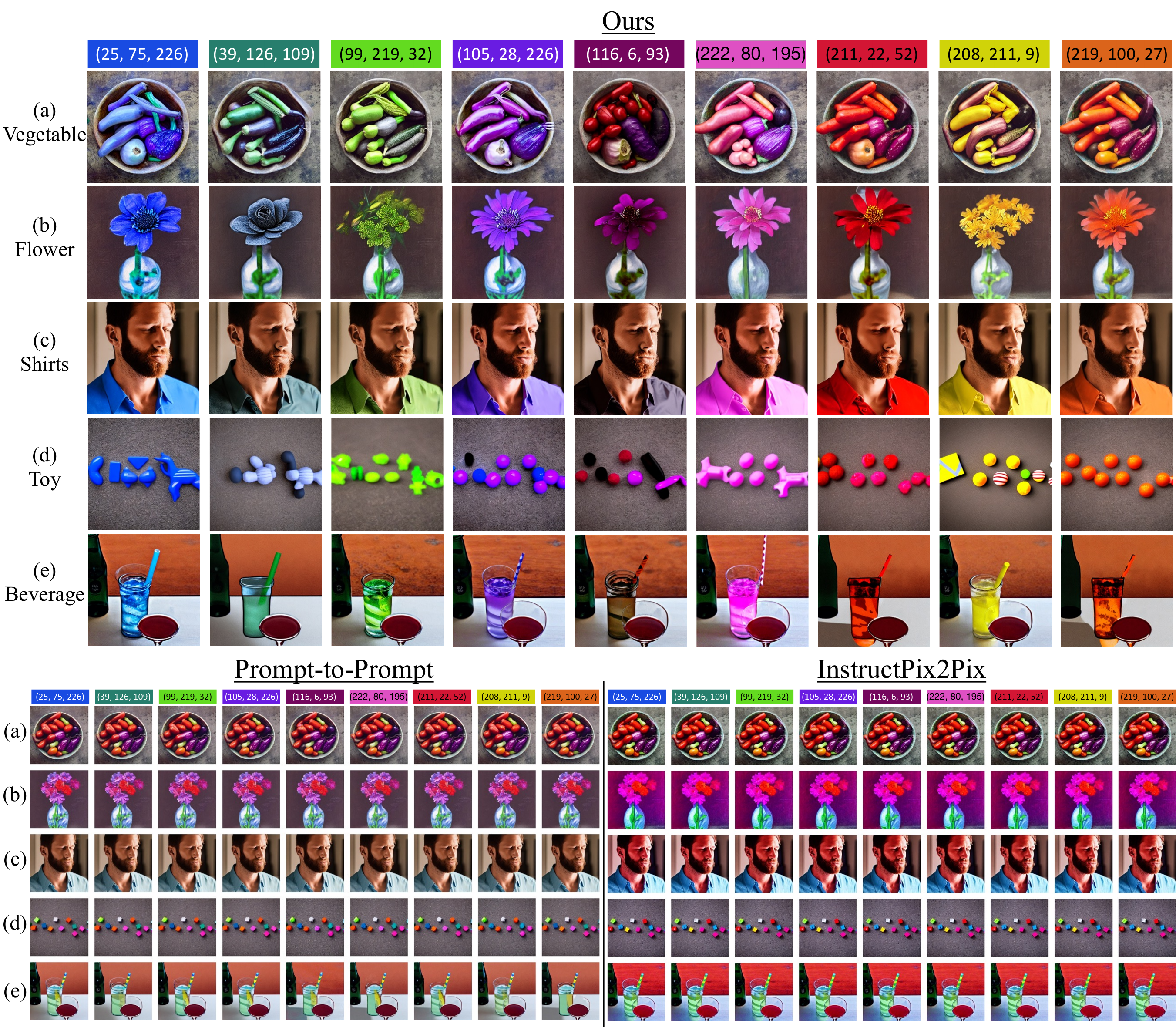}
    \caption{\textbf{Additional results of the font color.} We show the generation of different objects with colors from the \emph{RGB category}. Both baseline methods cannot interpret the RGB values correctly.}
    \label{fig:appendix_color3}
\end{figure}

\begin{figure}[ht]
    \centering
    \includegraphics[width=0.9\linewidth, trim=0 0 0 0, clip]{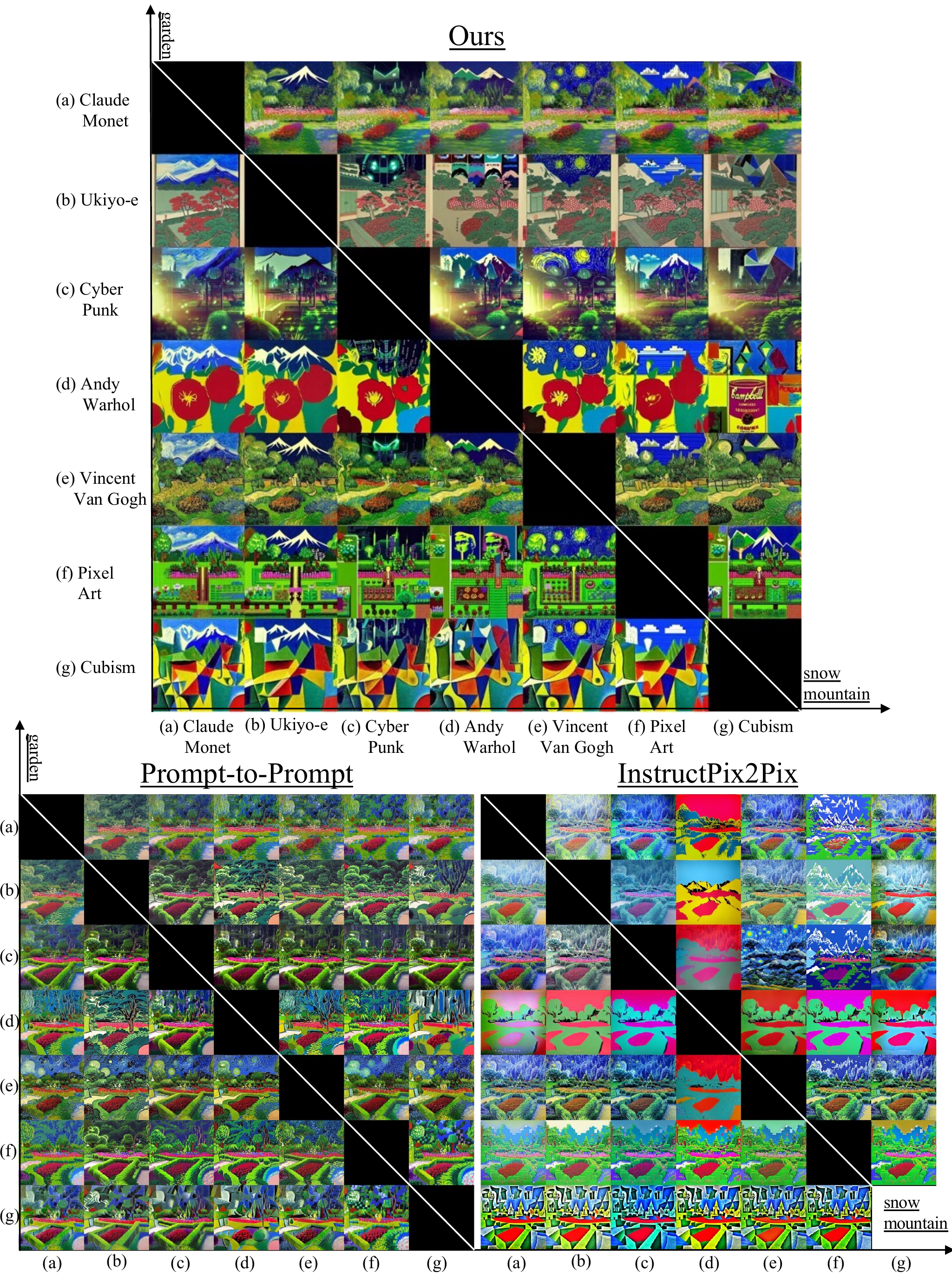}
    \caption{\textbf{Additional results of the font style.} We show images generated with different style combinations and prompt ``a beautiful garden in front of a snow mountain''. Each row contains ``snow mountain'' in 7 styles, and each column contains ``garden'' in 7 styles. Only our method can generate distinct styles for both objects.}
    \label{fig:appendix_style1}
\end{figure}

\begin{figure}[ht]
    \centering
    \includegraphics[width=0.9\linewidth, trim=0 0 0 0, clip]{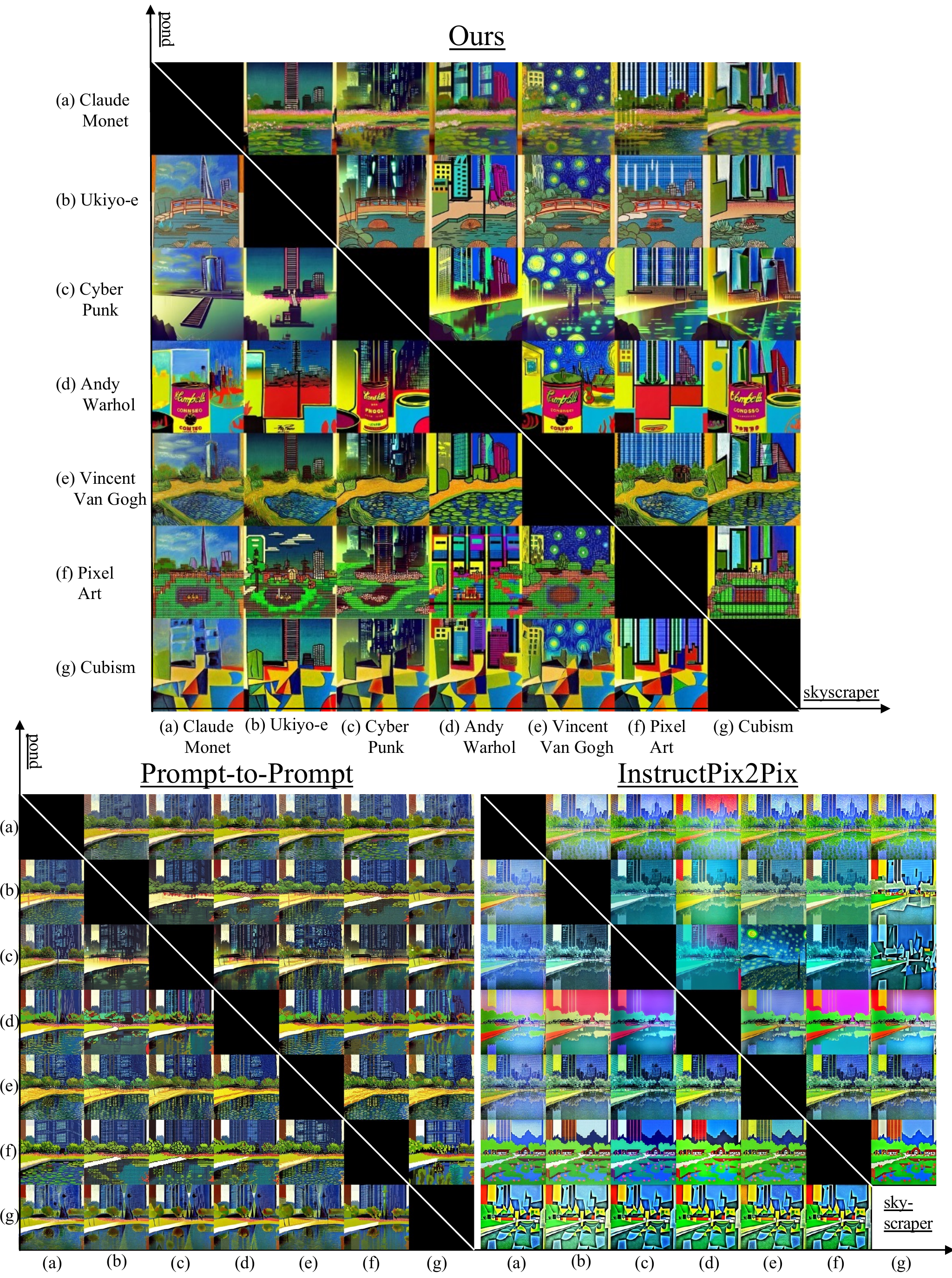}
    \caption{\textbf{Additional results of the font style.} We show images generated with different style combinations and prompt ``a small pond surrounded by skyscraper''. Each row contains ``skyscraper'' in 7 styles, and each column contains ``pond'' in 7 styles. Only our method can generate distinct styles for both objects.}
    \label{fig:appendix_style2}
\end{figure}

\begin{figure}[t]
    \centering
    \includegraphics[width=0.9\linewidth, trim=0 0 0 0, clip]{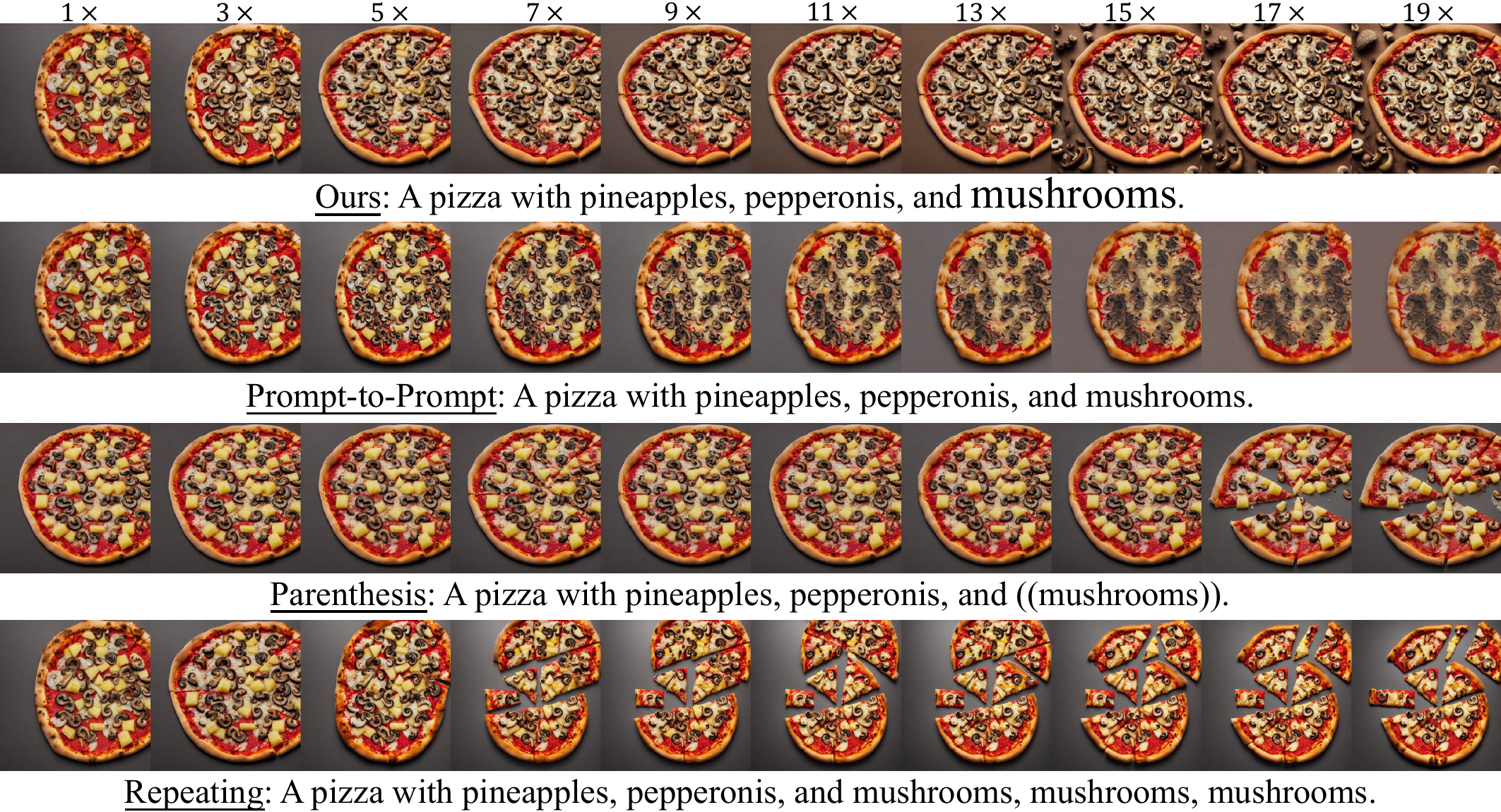}
    \caption{\textbf{Additional results of font sizes.} We use a token weight evenly sampled from $1$ to $20$ for the word `mushrooms' with our method and Prompt-to-Prompt. For parenthesis and repeating, we show results by repeating the word `mushrooms' and adding parentheses to the word `mushrooms' for $1$ to $10$ times. Prompt-to-Prompt suffers from generating artifacts. Heuristic methods are not effective.}
    \label{fig:fontsize1}
\end{figure}

\begin{figure}[t]
    \centering
    \includegraphics[width=0.9\linewidth, trim=0 0 0 0, clip]{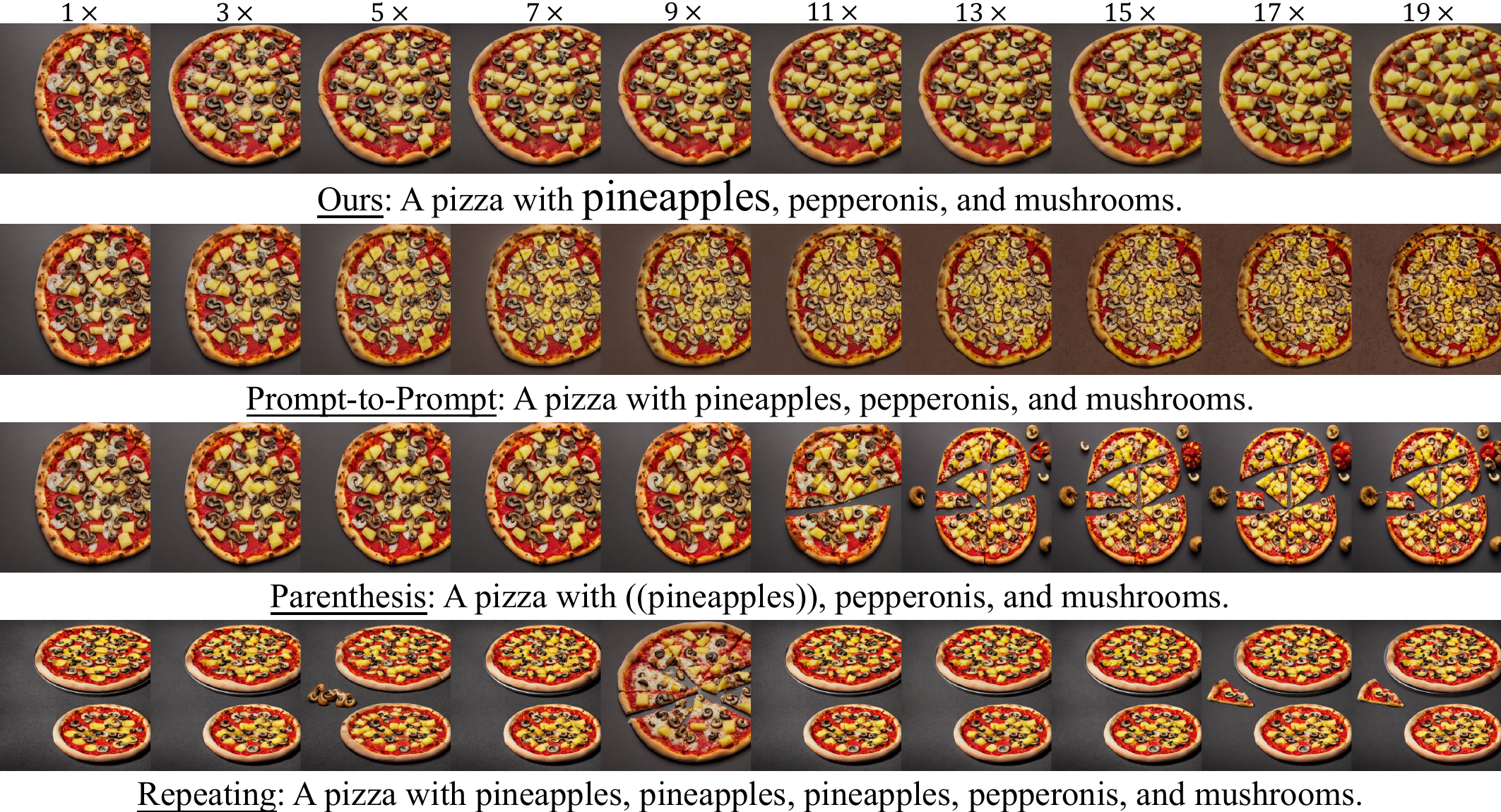}
    \caption{\textbf{Additional results of font sizes.} We use a token weight evenly sampled from $1$ to $20$ for the word `pineapples' with our method and Prompt-to-Prompt. For parenthesis and repeating, we show results by repeating the word `pineapples' and adding parentheses to the word `pineapples' for $1$ to $10$ times. Prompt-to-Prompt suffers from generating artifacts. Heuristic methods are not effective.}
    \label{fig:fontsize2}
\end{figure}

\begin{figure}[t]
    \centering
    \includegraphics[width=0.9\linewidth, trim=0 0 0 0, clip]{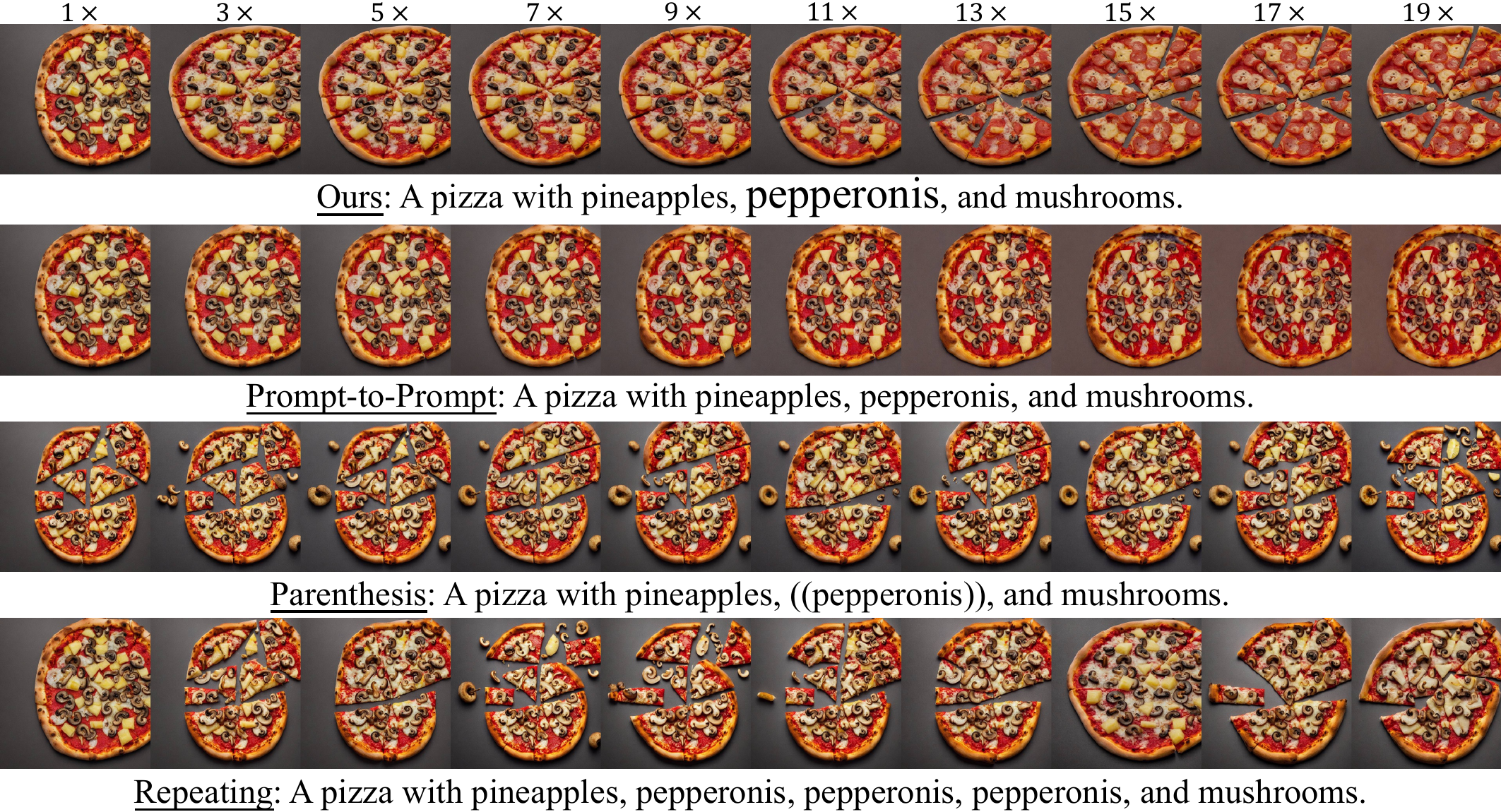}
    \caption{\textbf{Additional results of font sizes.} We use a token weight evenly sampled from $1$ to $20$ for the word `pepperonis' with our method and Prompt-to-Prompt. For parenthesis and repeating, we show results by repeating the word `pepperonis' and adding parentheses to the word `pepperonis' for $1$ to $10$ times. Prompt-to-Prompt suffers from generating artifacts. Heuristic methods are not effective.}
    \label{fig:fontsize3}
\end{figure}


\begin{figure}[ht]
    \vspace{-10pt}
    \centering
    \includegraphics[width=0.8\linewidth, trim=0 0 0 0, clip]{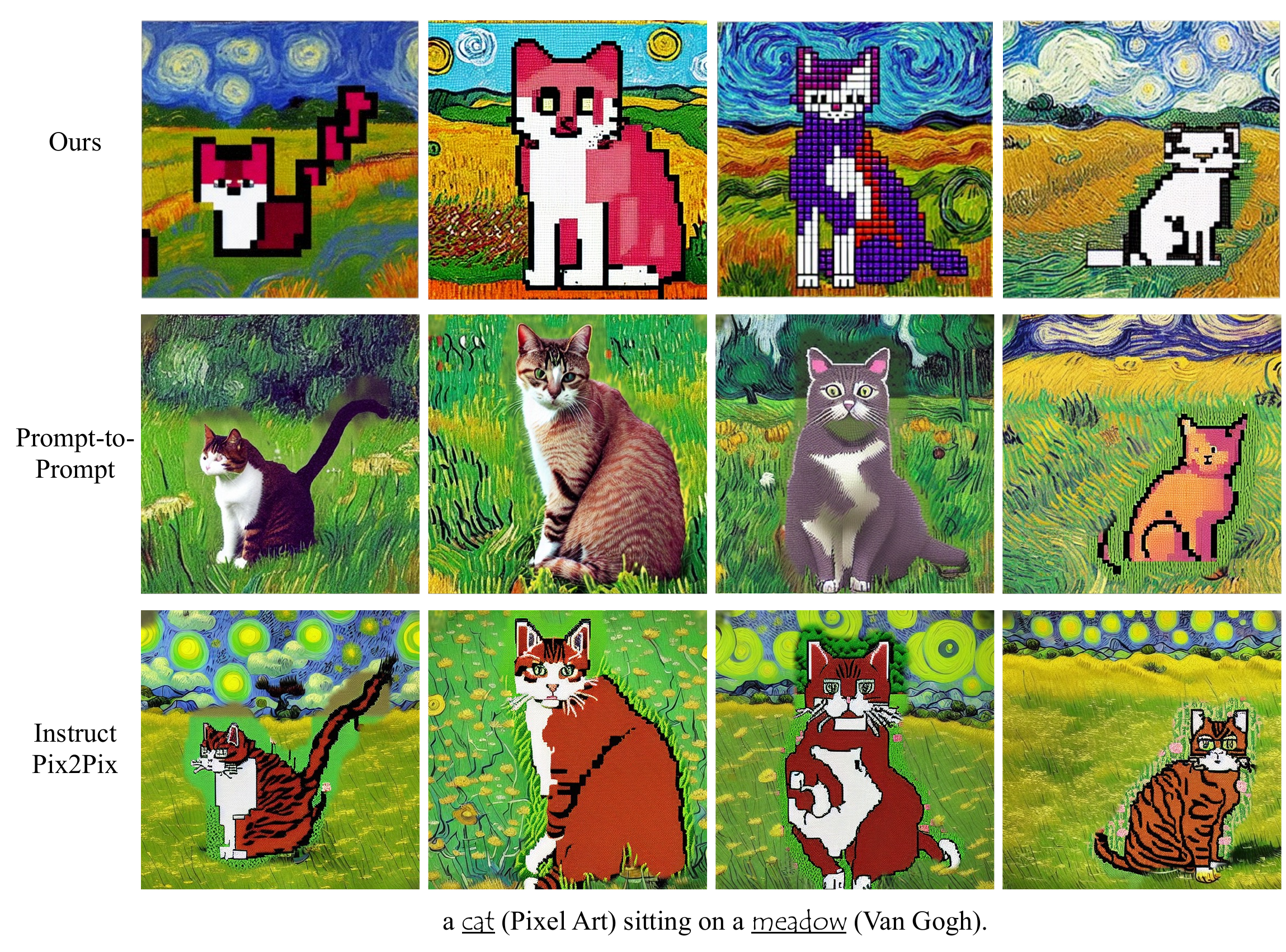}
    \includegraphics[width=0.8\linewidth, trim=0 5 0 10, clip]{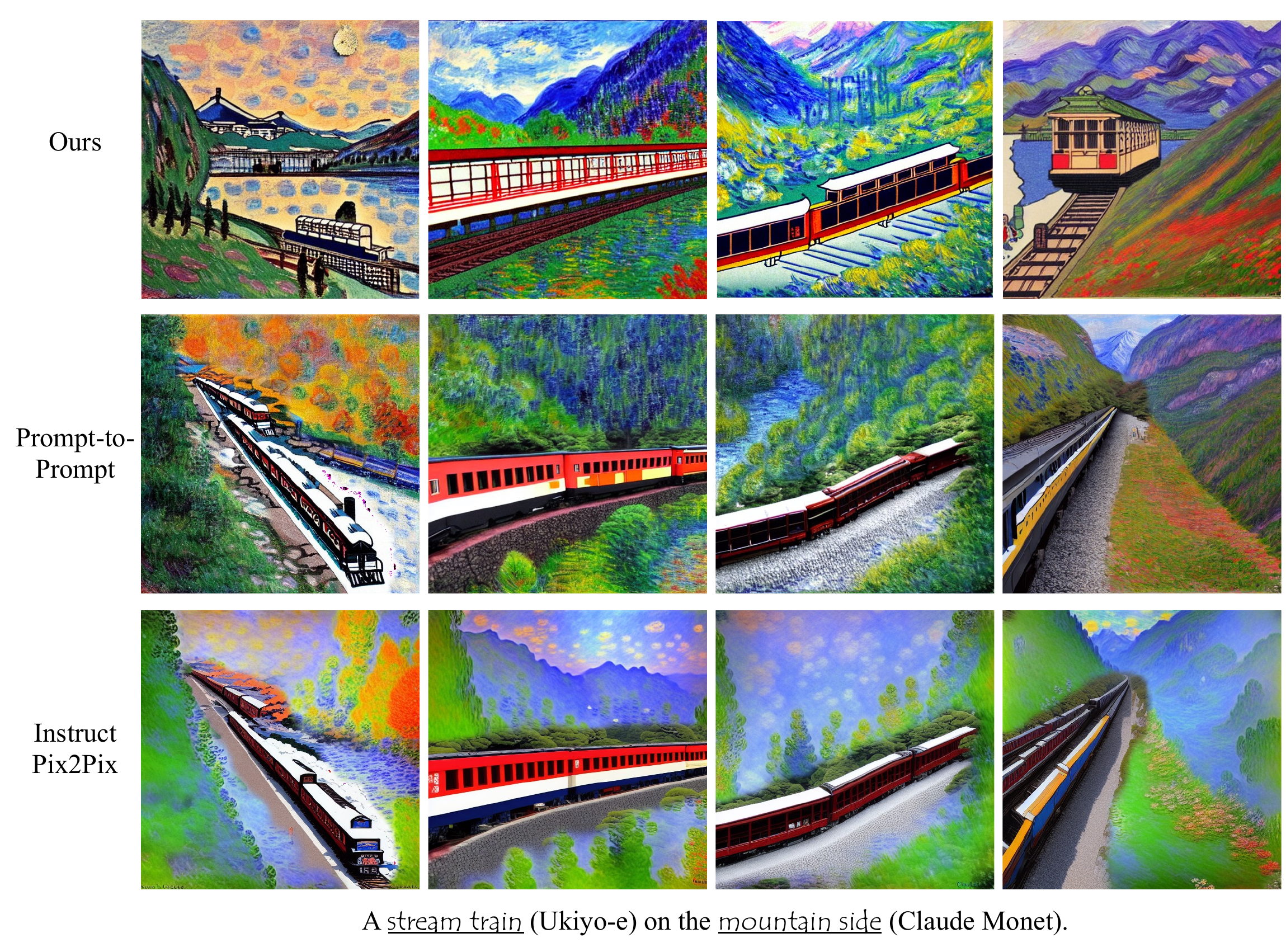}
    \caption{\textbf{Comparison with a simple composed-based method using different random seeds.} Since the methods like Prompt-to-Prompt~\citep{hertz2022prompt} cannot generate multiple styles on a single image, one simple idea to fix this is to apply the methods on two regions separately and compose them using the token maps. However, we show that this leads to sharp changes and artifacts at the boundary areas.}
    \label{fig:composed_comparison}
\end{figure}

\begin{figure}[ht]
    \vspace{-10pt}
    \centering
    \includegraphics[width=0.85\linewidth, trim=0 0 0 0, clip]{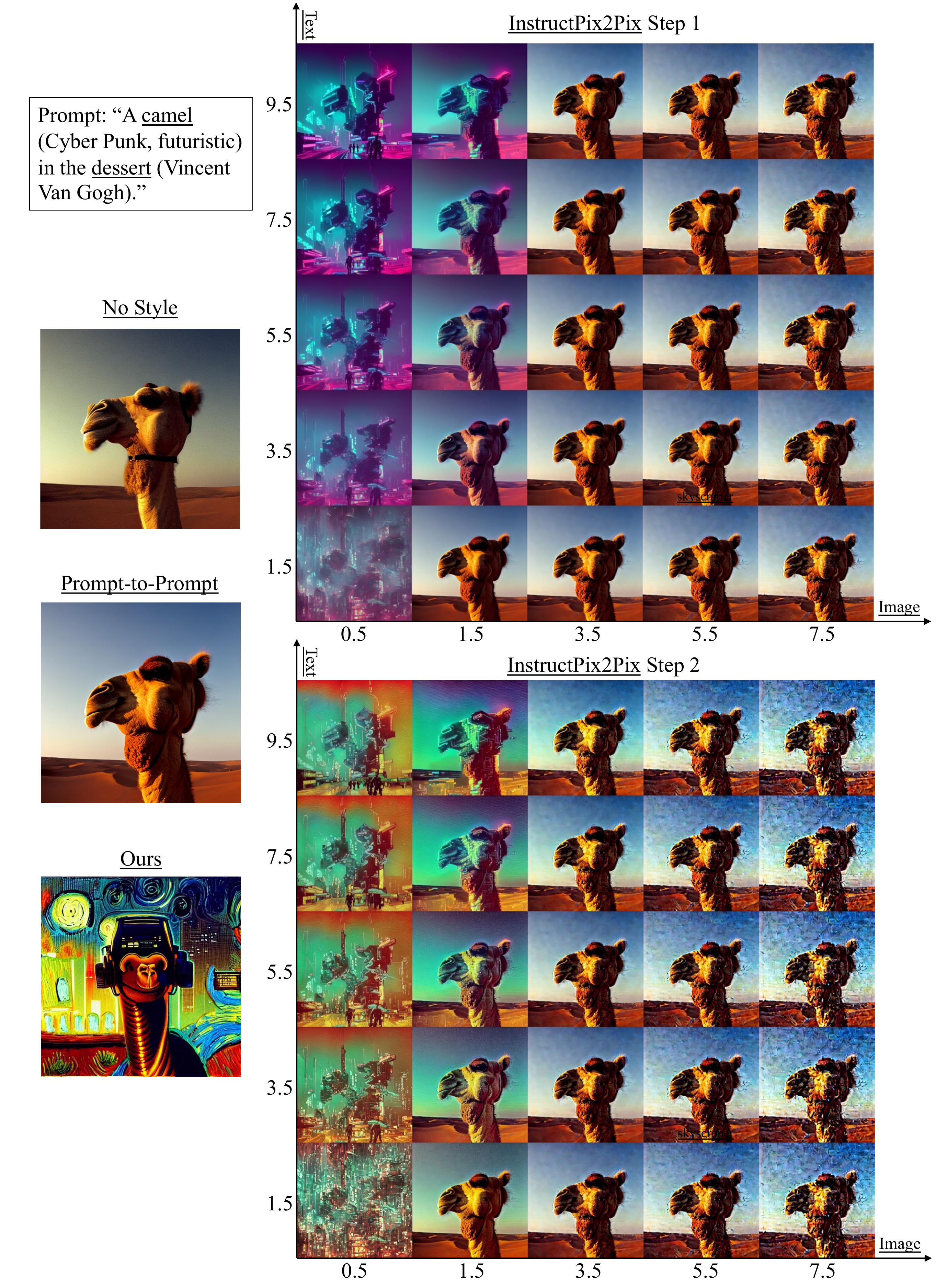}
    \caption{\textbf{Ablation of the classifier free guidance of InstructPix2Pix.} We show that InstruxtPix2Pix fails to generate both styles with different image and text classifier-free guidance (cfg) weights. When image-cfg is low, the desert is lost after the first editing. We use image-cfg$=1.5$ and text-cfg$=7.5$ in our experiment.}
    \label{fig:sweep_ip2p}
\end{figure}

\clearpage


\begin{figure}
\scriptsize
\centering
\begin{subfigure}{0.5\textwidth}
\centering
\begin{tikzpicture}
\begin{axis}[
    title={Minimal Distance},
    xlabel={CLIP Similarity},
    xlabel near ticks,
    ylabel={Distance to Target Color ($\downarrow$)},
    ymin=0,
    ymax=0.15,
    ytick={0,0.1},
    ylabel near ticks,
    ymajorgrids=true,
    xmin=0.25,
    xmax=0.28,
    xtick={0.26,0.27},
    xticklabels={0.26,0.27},
    major x tick style = transparent,
    width=92mm,
    height=55mm,
    ]
    
    \addplot[color=turquoise,mark=*] coordinates {
        (0.276, 0.124)
        (0.274, 0.046)
        (0.272, 0.034)
        (0.271, 0.036)
        (0.269, 0.040)
        (0.267, 0.040)
        (0.259, 0.046)
        (0.253, 0.055)
    };
\end{axis}
\end{tikzpicture}
\end{subfigure}%
\begin{subfigure}{0.5\textwidth}
\centering
\begin{tikzpicture}
\begin{axis}[
    title={Mean Distance},
    xlabel={CLIP Similarity},
    xlabel near ticks,
    ymin=0.35,
    ymax=0.7,
    ytick={0.4,0.5,0.6},
    xmin=0.25,
    xmax=0.28,
    xtick={0.25,0.26,0.27,0.28},
    ylabel near ticks,
    ymajorgrids=true,
    xmin=0.25,
    xmax=0.28,
    xtick={0.26,0.27},
    xticklabels={0.26,0.27},
    major x tick style = transparent,
    width=92mm,
    height=55mm,
    ]

\addplot[color=orange,mark=*] coordinates {
    (0.276, 0.639)
    (0.274, 0.456)
    (0.272, 0.434)
    (0.271, 0.424)
    (0.269, 0.417)
    (0.267, 0.421)
    (0.259, 0.433)
    (0.253, 0.445)
};
\end{axis}
\end{tikzpicture}
\end{subfigure}

\caption{\textbf{Ablation on the hyperparameter $\lambda$ in Equation (7)}. We report the trade-off of CLIP similarity and color distance achieved by sweeping the strength of color optimization $\lambda$.}
\label{fig:ablation1}
\end{figure}








\begin{figure*}[t]
    \centering
    \includegraphics[width=\linewidth, trim=0 0 0 0, clip]{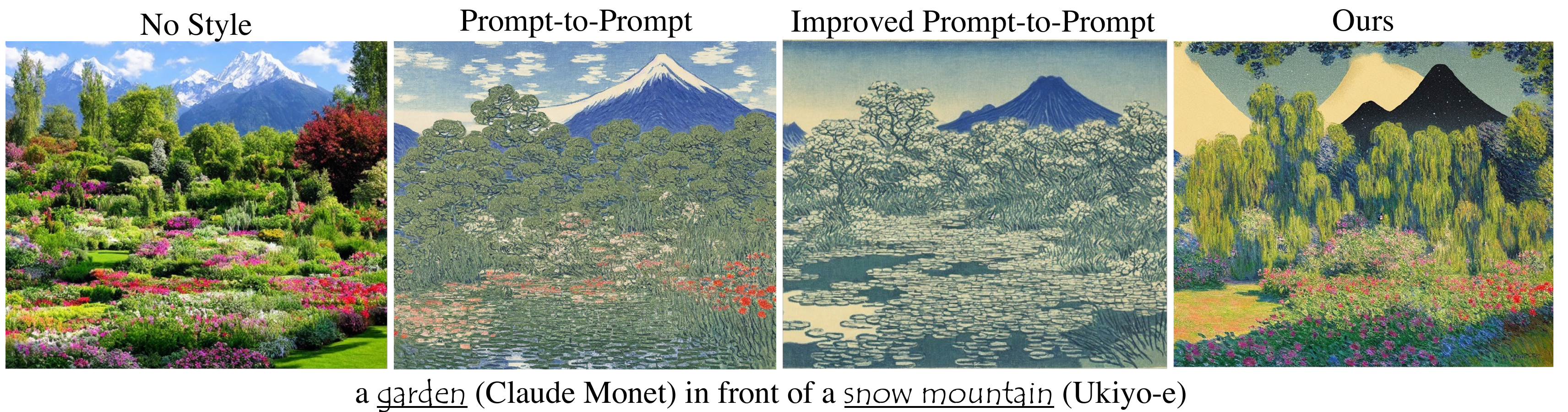}
    \caption{\textbf{Improved Prompt-to-Prompt.} Further constraining the attention maps for styles does not resolve the mixed style issue.}
    \label{fig:ablation3}
\end{figure*}


\topic{Ablation of the color guidance weight.} 
Changing the guidance strength $\lambda$ allows us to control the trade-off between \emph{fidelity} and \emph{color precision}. 
To evaluate the fidelity of the image, we compute the CLIP score between the generation and the plain text prompt. 
We plot the CLIP similarity vs. color distance in Figure~\ref{fig:ablation1} by sweeping $\lambda$ from $0$ to $20$. 
Increasing the strength always reduces the CLIP similarity as details are removed to satisfy the color objective. 
We find that larger $\lambda$ first reduces and then increases the distances due to the optimization divergence. 

\topic{Constrained Prompt-to-Prompt.} 
The original Attention Refinement proposed in Prompt-to-Prompt~\citep{hertz2022prompt} does not apply any constraint to newly added tokens' attention maps, which may be the reason that it fails with generating distinct styles. 
Therefore, we attempt to improve Prompt-to-Prompt by injecting the cross-attention maps for the newly added style tokens. 
For example, in Figure~\ref{fig:ablation3}, we use the cross attention map of ``garden'' for the style ``Claude Monet''. 
However, the method still produces a uniform style.

\topic{Human Evaluation}
We conduct a user study on crowdsourcing platforms.  
We show human annotators a pair of generated images and ask them which image more accurately expresses the reference color, artistic styles, or supplementary descriptions. 
To compare ours with each baseline, we show $135$ font color pairs, $167$ font style pairs, and $21$ footnote pairs to three individuals and receive $1938$ responses.
As shown in the table below, our method is chosen more than 80\% of the time over both baselines for producing more precise color and content given the long prompt and more than 65\% of the time for rendering more accurate artistic styles.
We will include a similar study at a larger scale in our revision.
\begin{table}[h]
\centering
\caption{Human evaluation results.}
\begin{tabular}{lccc}
\toprule
                          & Color & Style & Footnote \\
\midrule
Ours vs. Prompt-to-Prompt &  88.2\%  & 65.2\%  & 84.1\%  \\
Ours vs. InstructPix2Pix~  & 80.7\% &  69.8\%  & 87.3\%  \\
\bottomrule
\end{tabular}
\end{table}

\section{Additional Details}
\label{sec:addtional_details}

\subsection{Rich-text Benchmark}
\label{sec:benchmark_details}

This section details our construction process of the rich-text benchmark.

\topic{Footnote evaluation.} Here we provide our full prompt template used for GPT-4 to create complex scene prompts with two stages, which typically include three primary parts components:

\begin{itemize}
    \item \textbf{Instruction} describe the goal of task. In the first step, the goal is to generate both global and local description of the scene. In the second step, the goal is to compose the descriptions in an organic manner.

    \item \textbf{In-context exemplars} help enhance the model’s capability for these specific tasks and define the output format by providing manually created examples.

    \item \textbf{User prompt} is added in the end to initiate the task and let the model continue the conversation and generate the complex scene descriptions.
\end{itemize}

\begin{table*}
\centering
    \caption{The complete prompt to generate scene and object descriptions used in the first step in creating complex image prompt.}
    \label{tab:chatgpt}
    \tabulinesep=4pt
    \begin{tabu} to 1.\textwidth {@{}X[1.9,l]X[9,l]@{}}
    \toprule
    \textbf{Role} & \textbf{Content} \\
    \midrule
      \textbf{Instruction}& System:{ "In this task, you act like visual and captioning experts, write a caption of a complex scene image. The description contains an overall description of the scene and several descriptions of the object details that appear in the scene. Please only describe visual information in the caption. Here are some examples:"}\\
    \midrule
     \multirow{2}{*}{\textbf{In-context examples}} 
     & Assistant: {``Scene: A cozy cabin nestled in a snowy forest, and a snowman stands in the yard. Cabin: A charming wooden cabin with Christmas decoration, warm light coming out from the windows. Snowy Forest: Towering evergreen trees covered in a thick layer of pristine snow. Snowman: A cute snowman wearing a carrot nose, coal eyes, and a colorful scarf, welcoming visitors with a cheerful vibe."} \\
    & Assistant: {``Scene: A coffee table sits in front of a sofa on a cozy carpet. A painting on the wall. Coffee table: A rustic wooden coffee table adorned with scented candles and many books. Sofa: A plush sofa with a soft blanket and colorful pillows on it. Painting: A painting of wheat field with a cottage in the distance, close up shot, trending on artstation, HD, calm, complimentary color, realistic lighting, by Albert Bierstadt, Frederic Church."} \\
    & Assistant: {``Scene: A close-up of a cat riding a scooter. Tropical trees in the background. Cat:  A cat wearing sunglasses and has a bandana around its neck."} \\
    & Assistant: {``Scene: A car driving on the road. A bicycle nearby a tree. A cityscape in the background. Car: A sleek sports car gleams on the road in the sunlight, with its aerodynamic curves and polished finish catching the light. Bicycle: A bicycle with rusted frame and worn tires. Tree: A dead tree with a few red apples on it. Cityscape: A bustling Hongkong cityscape with towering skyscrapers."} \\
    & Assistant: {``Scene: A beach bonfire party with people sitting around the fire, surfboards leaning on a dune, and a cooler is nearby. People: People wearing casual beachwear, some wrapped in blankets, some holding cups of beer. Fire: Crackling fire with driftwood logs and glowing embers. Surfboards: Colorful and waxed surfboards are lined up, leaning against a sand dune. Cooler: A white cooler with blue cover filled with an assortment of beverages."} \\
    & Assistant: {``Scene: A small chair sits in front of a table on the wooden floor. There is a bookshelf nearby the window. Chair: A black leather office chair with a high backrest and adjustable arms. Table: A large wooden desk with a stack of books on top of it. Bookshelf: A bookshelf filled with colorful books and binders. Window: A window overlooks a stunning natural landscape of snow mountains."} \\
    & Assistant: {``Scene: A lush garden with a fountain. A grand mansion in the background. Garden: A garden is full of vibrant colors with a variety of flowers. Fountain: A fountain made of white marble with multiple tiers. The tiers are intricately carved with various designs. Mansion: An impressive two-story mansion with a royal exterior, white columns, and tile-made roof. The mansion has numerous windows, each adorned with white curtains."} \\
    & Assistant: {``Scene: A serene beach with crystal-clear turquoise water and palm trees. A couple walking along the shoreline. Beach: The beach has pink sand with seashells, crabs, and starfish on it. Palm Trees: Tall, swaying palm trees with lush green fronds and green coconuts. Couple: A young couple in matching, red beachwear, smiling as they enjoy their romantic stroll."} \\
    & Assistant: {``Scene: A cozy cafe with chalkboard menus, a barista behind the counter, and a pastry display. Chalkboard Menus: Chalkboard menus with handwritten lists of coffees and teas, specials highlighted. Barista: A barista wearing black uniform and a tie is steaming milk for a latte. Pastry Display: A glass pastry display filled with croissants, muffins, and scones."} \\
     \midrule
    \textbf{User prompt} & User : "Can you provide more examples like those above?" \\
    \bottomrule
    \end{tabu}
\end{table*}

\begin{table*}
\centering
    \caption{The complete prompt to generate full descriptions used in the second step in creating complex image prompt.}
    \label{tab:chatgpt}
    \tabulinesep=4pt
    \begin{tabu} to 1.\textwidth {@{}X[1.9,l]X[9,l]@{}}
    \toprule
    \textbf{Role} & \textbf{Content} \\
    \midrule
      \textbf{Instruction}& System:{ "In this task you are acting like a text editing expert. You are given an overall description of a scene and several descriptions of the object details that appear in the scene. Your job is to compose the descriptions into a paragraph that describe the scene. Make sure that the grammar is correct and the flow is natural. Here are a few examples:"}\\
    \midrule
     \multirow{2}{*}{\textbf{In-context examples}} 
     & User: ``Scene: A serene beach with crystal-clear turquoise water and palm trees. A couple walking along the shoreline. Beach: The beach has pink sand with seashells, crabs, and starfish on it. Palm Trees: Tall, swaying palm trees with lush green fronds and green coconuts. Couple: A young couple in matching, red beachwear, smiling as they enjoy their romantic stroll.'' \\
    & Assistant: ``A serene beach with crystal-clear turquoise water and tall, swaying palm trees with lush green fronds and green coconuts. The beach has pink sand with seashells, crabs, and starfish on it. A couple in matching, red beachwear walking along the shoreline,smiling as they enjoy their romantic stroll.'' \\
    
    & User: ``Scene: A cozy cabin nestled in a snowy forest, and a snowman stands in the yard. Cabin: A charming wooden cabin with Christmas decoration, warm light coming out from the windows. Snowy Forest: Towering evergreen trees covered in a thick layer of pristine snow. Snowman: A cute snowman wearing a carrot nose, coal eyes, and a colorful scarf, welcoming visitors with a cheerful vibe.'' \\
    & Assistant: ``A cozy, charming wooden cabin with Christmas decoration and warm light coming out from the windows nestled in a snowy forest with towering evergreen trees covered in a thick layer of pristine snow. A cute snowman a wearing a carrot nose, coal eyes, and acolorful scarf, welcoming visitors with a cheerful vibe, stands in the yard.'' \\
    
    & User: ``Scene: A street vendor surrounded by the shoppers. Buildings in the background. Street Vendor: A friendly street vendor stands behind a cart filled with an assortment of fresh fruits and vegetables. Shoppers: Diverse shoppers wearing Indian traditional clothes, carrying eco-friendly bags. Buildings: Indian-style old buildings in various colors.'' \\
    & Assistant: ``A friendly street vendor stands behind a cart filled with an assortment of fresh fruits and vegetables surrounded by the diverse shoppers wearing Indian traditional clothes, carrying eco-friendly bags. Indian-style old buildings in various colors in thebackground.'' \\
    
    & User: ``Scene: A vineyard in front of a winery on a sunny day. Winery: An elegant winery building with a rustic charm, set against a backdrop of rolling hills. Oak Wine barrels stacked outside. Vineyard: vineyard with rows of Lush green grapevines heavy with ripe, plump grapes, waiting to be harvested.'' \\
    & Assistant: ``A vineyard with rows of Lush green grapevines heavy with ripe, plump grapes, waiting to be harvested, in front of a elegant winery building with a rustic charm, set against a backdrop of rolling hills. Oak Wine barrels stacked outside, a sunny day'' \\
     \midrule
    \textbf{User prompt} & User : ``Can you write the descriptions given the new user input below. Do not add descriptions that are not part of the user input. + [user prompt]\\
    \bottomrule
    \end{tabu}
\end{table*}

\topic{Font style evaluation.} 
To compute the local CLIP scores at each local region to evaluate the stylization quality, we need to create test prompts with multiple objects and styles. 
We use seven popular styles that people use to describe the artistic styles of the generation, as listed below.
Note that for each style, to achieve the best quality, we also include complementary information like the name of famous artists in addition to the style.
\begin{verbatim}
styles = [
    'Claud Monet, impressionism, oil on canvas',
    'Ukiyoe',
    'Cyber Punk, futuristic',
    'Pop Art, masterpiece, Andy Warhol',
    'Vincent Van Gogh',
    'Pixel Art, 8 bits, 16 bits',
    'Abstract Cubism, Pablo Picasso'
]
\end{verbatim}
We also manually create a set of prompts, where each contains a combination of two objects, for stylization, resulting in $420$ prompts in total.
We generally confirm that Stable Diffusion~\citep{rombach2022high} can generate the correct combination, as our goal is not to evaluate the compositionality of the generation as in DrawBench~\citep{Imagen}.
The prompts and the object tokens used for our method are listed below.
\begin{verbatim}
candidate_prompts = [
    'A garden with a mountain in the distance.': ['garden', 'mountain'],
    'A fountain in front of an castle.': ['fountain', 'castle'],
    'A cat sitting on a meadow.': ['cat', 'meadow'],
    'A lighthouse among the turbulent waves in the night.': ['lighthouse', 'turbulent waves'],
    'A stream train on the mountain side.': ['stream train', 'mountain side'],
    'A cactus standing in the desert.': ['cactus', 'desert'],
    'A dog sitting on a beach.': ['dog', 'beach'],
    'A solitary rowboat tethered on a serene pond.': ['rowboat', 'pond'],
    'A house on a rocky mountain.': ['house', 'mountain'],
    'A rustic windmill on a grassy hill.': ['rustic', 'hill'],
]
\end{verbatim}

\topic{Font color evaluation.} 
To evaluate precise color generation capacity, we create a set of prompts with colored objects.
We divide the potential colors into three levels according to the difficulty of text-to-image generation models to depend on. 
The \emph{easy-level} color set contains 17 basic color names that these models generally understand. The complete set is as below. 
\begin{verbatim}
COLORS_easy = {
    'brown': [165, 42, 42],
    'red': [255, 0, 0],
    'pink': [253, 108, 158],
    'orange': [255, 165, 0],
    'yellow': [255, 255, 0],
    'purple': [128, 0, 128],
    'green': [0, 128, 0],
    'blue': [0, 0, 255],
    'white': [255, 255, 255],
    'gray': [128, 128, 128],
    'black': [0, 0, 0],
    'crimson': [220, 20, 60],
    'maroon': [128, 0, 0],
    'cyan': [0, 255, 255],
    'azure': [240, 255, 255],
    'turquoise': [64, 224, 208],
    'magenta': [255, 0, 255],
}
\end{verbatim}

The \emph{medium-level} set contain color names that are selected from the HTML color names~\footnote{\url{https://simple.wikipedia.org/wiki/Web_color}}. These colors are also standard to use for website design. 
However, their names are less often occurring in the image captions, making interpretation by a text-to-image model challenging.
To address this issue, we also append the coarse color category when possible, e.g., ``Chocolate'' to ``Chocolate brown''. The complete list is below.

\begin{verbatim}
COLORS_medium = {
    'Fire Brick red': [178, 34, 34],
    'Salmon red': [250, 128, 114],
    'Coral orange': [255, 127, 80],
    'Tomato orange': [255, 99, 71],
    'Peach Puff orange': [255, 218, 185],
    'Moccasin orange': [255, 228, 181],
    'Goldenrod yellow': [218, 165, 32],
    'Olive yellow': [128, 128, 0],
    'Gold yellow': [255, 215, 0],
    'Lavender purple': [230, 230, 250],
    'Indigo purple': [75, 0, 130],
    'Thistle purple': [216, 191, 216],
    'Plum purple': [221, 160, 221],
    'Violet purple': [238, 130, 238],
    'Orchid purple': [218, 112, 214],
    'Chartreuse green': [127, 255, 0],
    'Lawn green': [124, 252, 0],
    'Lime green': [50, 205, 50],
    'Forest green': [34, 139, 34],
    'Spring green': [0, 255, 127],
    'Sea green': [46, 139, 87],
    'Sky blue': [135, 206, 235],
    'Dodger blue': [30, 144, 255],
    'Steel blue': [70, 130, 180],
    'Navy blue': [0, 0, 128],
    'Slate blue': [106, 90, 205],
    'Wheat brown': [245, 222, 179],
    'Tan brown': [210, 180, 140],
    'Peru brown': [205, 133, 63],
    'Chocolate brown': [210, 105, 30],
    'Sienna brown': [160, 82, 4],
    'Floral White': [255, 250, 240],
    'Honeydew White': [240, 255, 240],
}
\end{verbatim}

The \emph{hard-level} set contains 50 randomly sampled RGB triplets as we aim to generate objects with arbitrary colors indicated in rich texts. For example, the color can be selected by an RGB slider.
\begin{verbatim}
COLORS_hard = {
    'color of RGB values [68, 17, 237]': [68, 17, 237],
    'color of RGB values [173, 99, 227]': [173, 99, 227],
    'color of RGB values [48, 131, 172]': [48, 131, 172],
    'color of RGB values [198, 234, 45]': [198, 234, 45],
    'color of RGB values [182, 53, 74]': [182, 53, 74],
    'color of RGB values [29, 139, 118]': [29, 139, 118],
    'color of RGB values [105, 96, 172]': [105, 96, 172],
    'color of RGB values [216, 118, 105]': [216, 118, 105],
    'color of RGB values [88, 119, 37]': [88, 119, 37],
    'color of RGB values [189, 132, 98]': [189, 132, 98],
    'color of RGB values [78, 174, 11]': [78, 174, 11],
    'color of RGB values [39, 126, 109]': [39, 126, 109],
    'color of RGB values [236, 81, 34]': [236, 81, 34],
    'color of RGB values [157, 69, 64]': [157, 69, 64],
    'color of RGB values [67, 192, 60]': [67, 192, 60],
    'color of RGB values [181, 57, 181]': [181, 57, 181],
    'color of RGB values [71, 240, 139]': [71, 240, 139],
    'color of RGB values [34, 153, 226]': [34, 153, 226],
    'color of RGB values [47, 221, 120]': [47, 221, 120],
    'color of RGB values [219, 100, 27]': [219, 100, 27],
    'color of RGB values [228, 168, 120]': [228, 168, 120],
    'color of RGB values [195, 31, 8]': [195, 31, 8],
    'color of RGB values [84, 142, 64]': [84, 142, 64],
    'color of RGB values [104, 120, 31]': [104, 120, 31],
    'color of RGB values [240, 209, 78]': [240, 209, 78],
    'color of RGB values [38, 175, 96]': [38, 175, 96],
    'color of RGB values [116, 233, 180]': [116, 233, 180],
    'color of RGB values [205, 196, 126]': [205, 196, 126],
    'color of RGB values [56, 107, 26]': [56, 107, 26],
    'color of RGB values [200, 55, 100]': [200, 55, 100],
    'color of RGB values [35, 21, 185]': [35, 21, 185],
    'color of RGB values [77, 26, 73]': [77, 26, 73],
    'color of RGB values [216, 185, 14]': [216, 185, 14],
    'color of RGB values [53, 21, 50]': [53, 21, 50],
    'color of RGB values [222, 80, 195]': [222, 80, 195],
    'color of RGB values [103, 168, 84]': [103, 168, 84],
    'color of RGB values [57, 51, 218]': [57, 51, 218],
    'color of RGB values [143, 77, 162]': [143, 77, 162],
    'color of RGB values [25, 75, 226]': [25, 75, 226],
    'color of RGB values [99, 219, 32]': [99, 219, 32],
    'color of RGB values [211, 22, 52]': [211, 22, 52],
    'color of RGB values [162, 239, 198]': [162, 239, 198],
    'color of RGB values [40, 226, 144]': [40, 226, 144],
    'color of RGB values [208, 211, 9]': [208, 211, 9],
    'color of RGB values [231, 121, 82]': [231, 121, 82],
    'color of RGB values [108, 105, 52]': [108, 105, 52],
    'color of RGB values [105, 28, 226]': [105, 28, 226],
    'color of RGB values [31, 94, 190]': [31, 94, 190],
    'color of RGB values [116, 6, 93]': [116, 6, 93],
    'color of RGB values [61, 82, 239]': [61, 82, 239],
}
\end{verbatim}

To write a complete prompt, we create a list of 12 objects and simple prompts containing them as below. The objects would naturally exhibit different colors in practice, such as ``flower'', ``gem'', and ``house''.

\begin{verbatim}
candidate_prompts = [
    'a man wearing a shirt': 'shirt',
    'a woman wearing pants': 'pants',
    'a car in the street': 'car',
    'a basket of fruit': 'fruit',
    'a bowl of vegetable': 'vegetable',
    'a flower in a vase': 'flower',
    'a bottle of beverage on the table': 'bottle beverage',
    'a plant in the garden': 'plant',
    'a candy on the table': 'candy',
    'a toy on the floor': 'toy',
    'a gem on the ground': 'gem',
    'a church with beautiful landscape in the background': 'church',
]
\end{verbatim}

\subsection{Experiment Details}

\topic{Baseline.}
We compare our method quantitatively with two strong baselines, Prompt-to-Prompt~\citep{hertz2022prompt} and InstructPix2Pix~\citep{brooks2022instructpix2pix}. 
The prompt refinement application of Prompt-to-Prompt allows adding new tokens to the prompt. 
We use plain text as the base prompt and add color or style to create the modified prompt. 
InstructPix2Pix~\citep{brooks2022instructpix2pix} allows using instructions to edit the image. 
We use the image generated by the plain text as the input image and create the instructions using templates ``turn the \textit{[object]} into the style of \textit{[style]},'' or ``make the color of \textit{[object]} to be \textit{[color]}''.
For the stylization experiment,  we apply two instructions in both parallel (InstructPix2Pix-para) and sequence (InstructPix2Pix-seq). We tune both methods on a separate set of manually created prompts to find the best hyperparameters. 
In contrast, it is worth noting that our method \emph{does not} require hyperparameter tuning.

\topic{Running time.} The inference time of our models depends on the number of attributes added to the rich text since we implement each attribute with an independent diffusion process. In practice, we always use a batch size of 1 to make the code compatible with low-resource devices. In our experiments on an NVIDIA RTX A6000 GPU, each sampling based on the plain text takes around $5.06$ seconds, while sampling an image with two styles takes around $8.07$ seconds, and sampling an image with our color optimization takes around $13.14$ seconds.

\end{document}